\pdfoutput=1

\documentclass[11pt]{article}

\usepackage{acl}

\usepackage{times}
\usepackage{latexsym}

\usepackage[T1]{fontenc}

\usepackage[utf8]{inputenc}

\usepackage{microtype}

\usepackage{graphicx}
\usepackage{bm,amsmath,amssymb,mathrsfs,amsfonts}
\usepackage{caption}
\usepackage{scalefnt}
\usepackage{multirow}
\usepackage{here}
\usepackage{booktabs}
\usepackage{enumitem}
\usepackage{type1cm}
\usepackage{cite}
\usepackage{url}
\usepackage{algorithm}
\usepackage{algpseudocode}
\usepackage{arydshln}
\usepackage{color}
\usepackage{graphics}
\usepackage{tikz}
\usepackage{pgfplots}
\pgfplotsset{compat=1.17}
\usepackage{svg}
\usepackage{cleveref}

\usepackage[subrefformat=parens]{subcaption}
\crefformat{section}{\S#2#1#3} %
\crefformat{subsection}{\S#2#1#3}
\crefformat{subsubsection}{\S#2#1#3}

\newcommand{\ie}{\textit{i.e.}}

\newcommand{\specv}{S2SpecT}
\newcommand{\specvvplus}{S2SpecT2}
\newcommand{\translatotronv}{Translatotron}
\newcommand{\translatotronvv}{Translatotron2}
\newcommand{\unitv}{S2UT}
\newcommand{\unity}{UnitY}

\newcommand{\todo}{\textcolor{black}}

\newcommand{\srcspeech}{X}
\newcommand{\srcspeechone}{X_{\rm 1}}
\newcommand{\srcspeechtwo}{X_{\rm 2}}
\newcommand{\srcspeechi}{X_{i}}
\newcommand{\srctext}{Y_{\rm src}}
\newcommand{\tgttext}{Y}
\newcommand{\tgttexti}{y_{i}}
\newcommand{\tgttextbeforei}{Y_{<i}}
\newcommand{\tgtspeech}{S}
\newcommand{\tgtunit}{U}
\newcommand{\tgtuniti}{u_{i}}
\newcommand{\tgtunitbeforei}{U_{<i}}
\newcommand{\hyptgttext}{\hat{Y}}
\newcommand{\hyptgtunit}{\hat{U}}

\newcommand{\hyptgtwave}{\hat{W}}

\newcommand{\textlen}{M}
\newcommand{\unitlen}{L}
\newcommand{\srcencout}{H}
\newcommand{\textdecout}{D^{\rm text}}
\newcommand{\textdecouti}{D^{\rm text}_{i}}
\newcommand{\ttuencout}{Z}
\newcommand{\unitdecout}{D^{\rm unit}}
\newcommand{\unitdecouti}{D^{\rm unit}_{i}}
\newcommand{\dmodel}{d_{\rm model}}
\newcommand{\dff}{d_{\rm ff}}
\newcommand{\nheads}{N_{\rm head}}
\newcommand{\nlayerfirstdec}{N_{\rm 1st}}
\newcommand{\nlayerseconddec}{N_{\rm 2nd}}
\newcommand{\nlayerttuenc}{N_{\rm t2u}}
\newcommand{\nlayerttsenc}{N_{\rm t2s}}
\newcommand{\textdec}{\texttt{TDec}}
\newcommand{\ttuenc}{\texttt{T2UEnc}}
\newcommand{\unitdec}{\texttt{UDec}}

\newcommand{\probstt}{P_{\rm s2t}}
\newcommand{\probstu}{P_{\rm s2u}}
\newcommand{\lossasr}{{\cal L}_{\rm asr}}
\newcommand{\lossstt}{{\cal L}_{\rm s2t}}
\newcommand{\lossstu}{{\cal L}_{\rm s2u}}
\newcommand{\losssts}{{\cal L}_{\rm s2s}}
\newcommand{\lossrdropasr}{{\cal L}^{\rm asr}_{\rm kl}}
\newcommand{\lossrdropstt}{{\cal L}^{\rm s2t}_{\rm kl}}
\newcommand{\lossrdropstu}{{\cal L}^{\rm s2u}_{\rm kl}}

\newcommand{\lossrdrop}{{\cal L}_{\rm kl}}
\newcommand{\losslone}{{\cal L}_{\rm 1}}
\newcommand{\lossltwo}{{\cal L}_{\rm 2}}
\newcommand{\losseos}{{\cal L}_{\rm eos}}
\newcommand{\lossctc}{{\cal L}_{\rm ctc}}
\newcommand{\losstotal}{{\cal L}_{\rm total}}
\newcommand{\kldiv}{{\mathcal D}_{\rm kl}}
\newcommand{\beamsizetext}{B_{\rm 1st}}
\newcommand{\beamsizeunit}{B_{\rm 2nd}}
\newcommand{\omegafirst}{\Omega_{\rm 1st}}
\newcommand{\omegasecond}{\Omega_{\rm 2nd}}
\newcommand{\weightasr}{w_{\rm asr}}
\newcommand{\weightstt}{w_{\rm s2t}}
\newcommand{\weightrdropstu}{\alpha}
\newcommand{\weightrdropstt}{\beta}
\newcommand{\weightrdropasr}{\gamma}
\newcommand{\weightctc}{w_{\rm ctc}}

\newcommand{\asrbleuimpoverstutfisher}{4.2}  %
\newcommand{\asrbleuimpoverstutcvss}{3.7}  %
\newcommand{\asrbleuimpoverstutmultidomain}{2.5}  %
\newcommand{\speedupoverspec}{2.51}
\newcommand{\speedupoverstut}{2.83}
\newcommand{\flopsreductionoverspec}{1.65}
\newcommand{\flopsreductionoverstut}{3.19}

\makeatletter
\def\adl@drawiv#1#2#3{%
        \hskip.5\tabcolsep
        \xleaders#3{#2.5\@tempdimb #1{1}#2.5\@tempdimb}%
                #2\z@ plus1fil minus1fil\relax
        \hskip.5\tabcolsep}
\newcommand{\cdashlinelr}[1]{%
  \noalign{\vskip\aboverulesep
           \global\let\@dashdrawstore\adl@draw
           \global\let\adl@draw\adl@drawiv}
  \cdashline{#1}
  \noalign{\global\let\adl@draw\@dashdrawstore
           \vskip\belowrulesep}}
\makeatother

\title{UnitY: Two-pass Direct Speech-to-speech Translation with Discrete Units}

\author{Hirofumi Inaguma${}^{\heartsuit}$, Sravya Popuri${}^{\heartsuit}$, Ilia Kulikov${}^{\heartsuit}$, Peng-Jen Chen${}^{\heartsuit}$, \\
{\bf Changhan Wang${}^{\heartsuit}$},
{\bf Yu-An Chung${}^{\heartsuit}$},
{\bf Yun Tang${}^{\heartsuit}$}, \\
{\bf Ann Lee${}^{\heartsuit}$},
{\bf Shinji Watanabe${}^{\clubsuit}$},
{\bf Juan Pino${}^{\heartsuit}$} \\
FAIR, Meta AI${}^{\heartsuit}$,
Carnegie Mellon University${}^{\clubsuit}$ \\ \texttt{\{hirofumii,juancarabina\}@meta.com} \\}

\begin{document}
\maketitle
\begin{abstract}
Direct speech-to-speech translation (S2ST), in which all components can be optimized jointly, is advantageous over cascaded approaches to achieve fast inference with a simplified pipeline.
We present a novel two-pass direct S2ST architecture, {\textit UnitY}, which first generates textual representations and predicts discrete acoustic units subsequently.
We enhance the model performance by subword prediction in the first-pass decoder, advanced two-pass decoder architecture design and search strategy, and better training regularization.
To leverage large amounts of unlabeled text data, we pre-train the first-pass text decoder based on the self-supervised denoising auto-encoding task.
Experimental evaluations on benchmark datasets at various data scales demonstrate that UnitY outperforms a single-pass speech-to-unit translation model by {\asrbleuimpoverstutmultidomain}-{\asrbleuimpoverstutfisher} ASR-BLEU with {\speedupoverstut}$\times$ decoding speed-up.
We show that the proposed methods boost the performance even when predicting spectrogram in the second pass.
However, predicting discrete units achieves {\speedupoverspec}$\times$ decoding speed-up compared to that case.
\end{abstract}

\section{Introduction}\label{sec:intro}
Automatic speech translation to another language is an indispensable technology for international communications, with the spread of social media and virtual communications nowadays.
A traditional approach of speech-to-speech translation (S2ST) is to cascade automatic speech recognition (ASR), machine translation (MT), and text-to-speech (TTS) components, each of which is optimized separately on different data~\citep{lavie1997janus,nakamura2006atr,wahlster2013verbmobil}.
With the emergence of sequence-to-sequence models~\citep{sutskever2014sequence,cho2014learning,bahdanau2014neural}, however, it is getting prevailing to adopt a direct approach\footnote{\citet{lee2021direct} defines a direct S2ST model as a model that does not use intermediate text representations while \citet{jia2022translatotron} defines it as a model that directly predicts the target spectrogram. In this paper, we use a more general definition that the entire architecture is optimized jointly and the translation is conducted in a more direct way. We do not include a vocoder in the training pipeline of all direct models.}.
This approach consists in translating input speech into the other language based on a single architecture with fewer components than the cascaded systems~\citep{jia2019direct,tjandra2019speech,zhang2021uwspeech}.
The direct approach is attractive for building a low-latency system with a simplified pipeline, thus reducing developing costs.
However, direct S2ST models suffer from poor performance due to data scarcity, similar to direct speech-to-text translation (S2TT) models~\citep{listen_and_translate}.
In the field of S2TT, data shortage has been addressed by leveraging pre-training~\citep{berard2018end,wang21r_interspeech,tang-etal-2022-unified}, multi-task learning~\citep{weiss2017sequence,tang-etal-2021-improving}, pseudo labeling~\citep{jia2019leveraging,pino2020self}, knowledge distillation~\citep{liu2019end,inaguma-etal-2021-source}.
Consequently, the translation quality of direct S2TT models is approaching that of cascaded S2TT models~\citep{ansari-etal-2020-findings,anastasopoulos-etal-2021-findings}.
These techniques have also shown the effectiveness for direct S2ST models and led to a decent performance~\citep{kano2021transformer,dong2022leveraging,jia2022leveraging,popuri2022enhanced}.

Recent works~\citep{lee2021direct,lee2021textless} propose to model discrete acoustic units, extracted from HuBERT~\citep{hsu2021hubert}, instead of a continuous speech signal that enables usage of a standard cross-entropy loss during training.
This speech-to-unit translation (S2UT) model significantly shortens the target sequence length and thus makes training and inference more efficient.
The discrete units are directly converted to the waveform with a unit-based neural vocoder~\citep{polyak21_interspeech} bypassing spectrogram representation.
On the other hand, {\translatotronvv}~\citep{jia2022translatotron} decomposes the target representations into linguistic and acoustic counterparts explicitly.
The former predicts a phoneme sequence first, and the latter synthesizes the target spectrogram conditioned on the continuous representation of the linguistic part.

This paper presents a novel two-pass direct S2ST architecture, dubbed {\textit \unity}, which takes the best of both worlds of the S2UT model and {\translatotronvv}.
Unlike {\translatotronvv}, {\unity} models linguistic sequences using subwords ({\it first pass}) instead of phonemes, and it models speech as a discrete sequence of acoustic units ({\it second pass}).
To achieve better translation quality and decoding efficiency, {\unity} consists of a deep text decoder and a shallow unit decoder and enables better generalization to the first-pass decoder.
We further introduce a text-to-unit (T2U) encoder between the two decoders to bridge the gap between textual and acoustic representations.
Following the success of large-scale pre-training, we leverage unlabeled text effectively to pre-train the first pass text decoder with multilingual BART (mBART)~\citep{liu2020multilingual} at the subword level.

Extensive experiments show the superiority of the {\unity} S2ST system measured by both translation quality and runtime efficiency. First, {\unity} achieves {\asrbleuimpoverstutfisher}, {\asrbleuimpoverstutcvss}, and {\asrbleuimpoverstutmultidomain} ASR-BLEU improvements over the S2UT model on the Fisher Es$\to$En~\citep{fisher_callhome}, CVSS-C~\citep{jia-etal-2022-cvss}, and multi-domain En$\leftrightarrow$Es~\citep{popuri2022enhanced} corpora, respectively.
The improvement holds even with high-resource data and pre-training.
In addition, our proposed design improves {\translatotronvv} as well, indicating its versatility for two-pass direct S2ST architectures regardless of the choice of the target.
Second, {\unity} achieves {\speedupoverstut}$\times$ and {\speedupoverspec}$\times$ decoding speed-ups over the {\unitv} and improved {\translatotronvv} models, respectively.
A combination of the aforementioned improvements suggests the {\unity} design as a starting point for further improvements in direct S2ST.
\footnote{Code will be available upon the paper acceptance.}

\section{\unity}\label{sec:unity}
\vspace{-1mm}  %
In this section, we propose {\textit \unity}, a two-pass direct S2ST model that generates subwords and discrete acoustic units subsequently.
Hereafter, we refer to discrete acoustic units as discrete units for brevity.
Let $\srcspeech$ denote a source speech input, and $\tgttext=(y_{1},\dots,y_{\textlen})$ and $\tgtunit=(u_{1},\dots,u_{\unitlen})$ be the corresponding reference text translation and discrete unit sequences in the target language, respectively.
Note that there is no duration information for each discrete unit in $\tgtunit$, because consecutive units are collapsed~\citep{lee2021direct}.

\begin{figure}[t]
  \centering
  \includegraphics[width=0.99\linewidth]{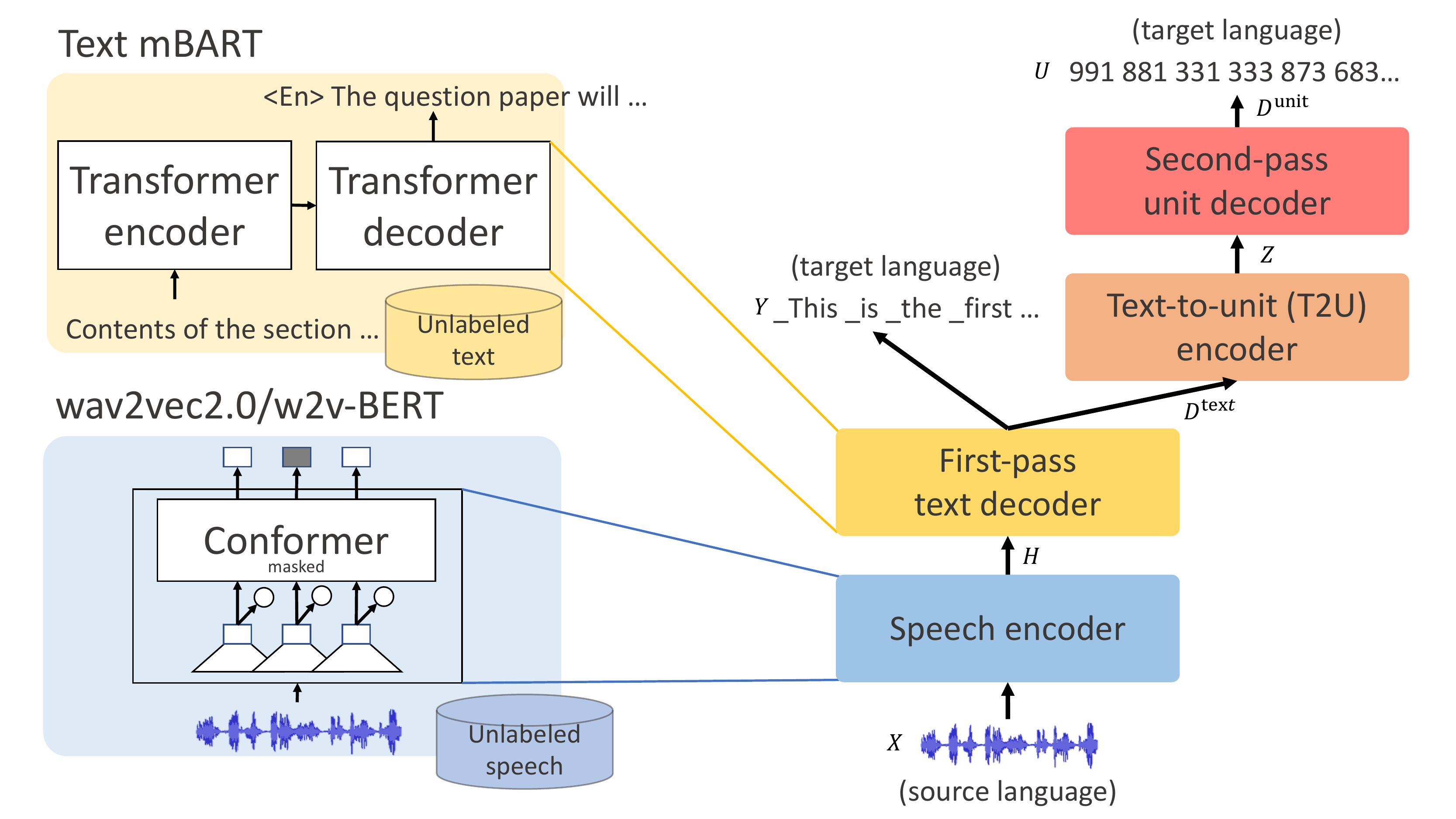}
  \vspace{-6mm}
  \caption{Model architecture of {\unity}}
  \label{fig:architecture}
  \vspace{-3mm}
\end{figure}

\subsection{Architecture}\label{ssec:architecture}
The overall architecture of {\unity} is shown in Figure~\ref{fig:architecture}.
{\unity} consists of four modules: speech encoder, first-pass text decoder, text-to-unit (T2U) encoder, and second-pass unit decoder.
We build the speech encoder based on Conformer~\citep{gulati2020}, which augments Transformer~\citep{vaswani2017attention} with a convolution module, while implementing the rest three modules based on Transformer.
{\unity} has five major architecture modifications from {\translatotronvv}~\citep{jia2022translatotron}, (1) generating subwords instead of phonemes in the first pass, (2) generating discrete units instead of spectrograms in the second pass to bypass duration modeling, (3) replacing Long Short-Term Memory (LSTM)~\citep{lstm} layers with Transformer layers in both decoders, (4) introducing a T2U encoder between the two decoders, and (5) assigning more model capacities to the first pass.

\paragraph{First-pass text decoder}
The first-pass text decoder {\textdec} generates a sequence of subwords $\tgttext$ autoregressively by attending the speech encoder output $\srcencout$.
The training objective of the first pass is to minimize the direct S2TT loss $\lossstt$ as:
\begin{eqnarray*}
\lossstt(\tgttext|\srcspeech) &=& - \frac{1}{\textlen} \sum_{i=1}^{\textlen} \log \probstt(\tgttexti|\srcspeech,\tgttextbeforei) \\
&=& - \frac{1}{\textlen} \sum_{i=1}^{\textlen} \log \probstt(\tgttexti|\textdecouti) \\
\textdecouti &=& \textdec(\srcencout,\tgttextbeforei),
\end{eqnarray*}
where $\textdecouti \in \mathbb{R}^{\dmodel}$ is the $i$-th continuous decoder state right before projecting it to the logit.
We consider that $\textdecout$ contains rich acoustic information in addition to contextual information thanks to multiple multi-head cross-attention over $\srcencout$.

There are five advantages of generating subwords instead of phonemes.
First, the sequence length is considerably reduced, leading to better training and inference efficiencies~\citep{cherry-etal-2018-revisiting}.
Second, using large vocabularies improves the translation quality of the first pass~\citep{gowda-may-2020-finding}.
Third, the text output helps the audience understand the translation content while listening to the audio.
Fourth, our approach can easily scale to more target languages, as it is unnecessary to prepare separate grapheme-to-phoneme (G2P) models for each target language.
Last, readable text can be generated without any complicated post-processing such as WFST~\citep{mohri2002weighted,bahdanau2016end}.

\vspace{-1mm}  %
\paragraph{T2U encoder}
A bidirectional T2U encoder {\ttuenc} transforms the continuous states of the first-pass decoder $\textdecout \in \mathbb{R}^{\textlen \times \dmodel}$ into $\ttuencout \in \mathbb{R}^{\textlen \times \dmodel}$ as $\ttuencout = \ttuenc(\textdecout)$.
The T2U encoder bridges the gap in representations between text and unit decoders without changing the sequence length.

\vspace{-1mm}  %
\paragraph{Second-pass unit decoder}
The second-pass unit decoder {\unitdec} generates a sequence of discrete units $\tgtunit$ autoregressively by attending to only the T2U encoder output $\ttuencout$.
The training objective of the second pass is to minimize $\lossstu$ similar to the S2UT task while being conditioned on $\tgttext$ as:
\begin{eqnarray*}
\lossstu(\tgtunit|\srcspeech,\tgttext) = - \frac{1}{\unitlen} \sum_{i=1}^{\unitlen} \log \probstu(\tgtuniti|\srcspeech,\tgttext,\tgtunitbeforei) \\
= - \frac{1}{\unitlen} \sum_{i=1}^{\unitlen} \log \probstu(\tgtuniti|\unitdecouti) \\
\unitdecouti = \unitdec(\ttuencout,\tgtunitbeforei)
= \unitdec(\srcencout,\tgttext,\tgtunitbeforei),
\end{eqnarray*}
where $\unitdecouti \in \mathbb{R}^{\dmodel}$ is the $i$-th continuous decoder state right before projecting it to the logit.
The unit decoder does not attend to $\srcencout$ to synchronize the text and unit outputs, similar to the motivation in~\citep{jia2022translatotron}.
In other words, we do not expect that the second-pass decoder corrects translation errors from the first-pass decoder.\footnote{We also investigate attending to the speech encoder output with an additional cross-attention, but it does not lead to an improvement in ASR-BLEU. We discuss this in \cref{ssec:result_ablation_study}}
Once the unit generation finishes, a separate unit-based vocoder~\citep{polyak21_interspeech} converts the discrete units to the waveform with duration prediction of each discrete unit~\citep{lee2021direct}.
The total training objective of {\unity}, $\losstotal$, is formulated as follows:
\begin{multline}\label{eq:total_loss_unitvv}
\losstotal = \lossstu(\tgtunit|\srcspeech,\tgttext) + \weightstt \lossstt(\tgttext|\srcspeech),
\end{multline}
where $\weightstt$ is a weight for the S2TT loss.

\begin{figure*}[tbp]
  \begin{minipage}[b]{0.49\columnwidth}
    \centering
    \includegraphics[keepaspectratio,scale=0.25]{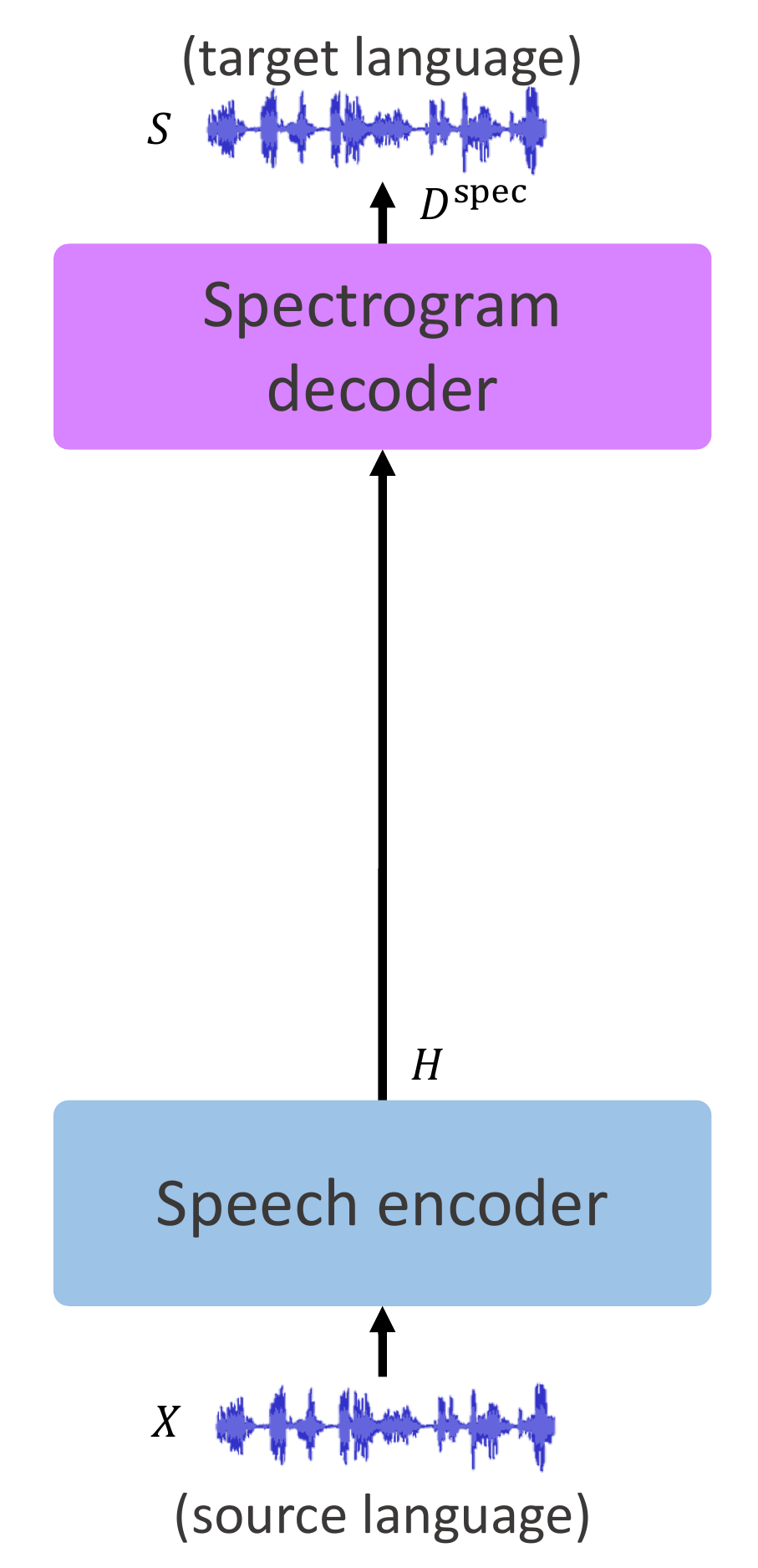}
    \subcaption{\specv}
    \label{fig:specv}
  \end{minipage}
  \begin{minipage}[b]{0.49\columnwidth}
    \centering
    \includegraphics[keepaspectratio,scale=0.25]{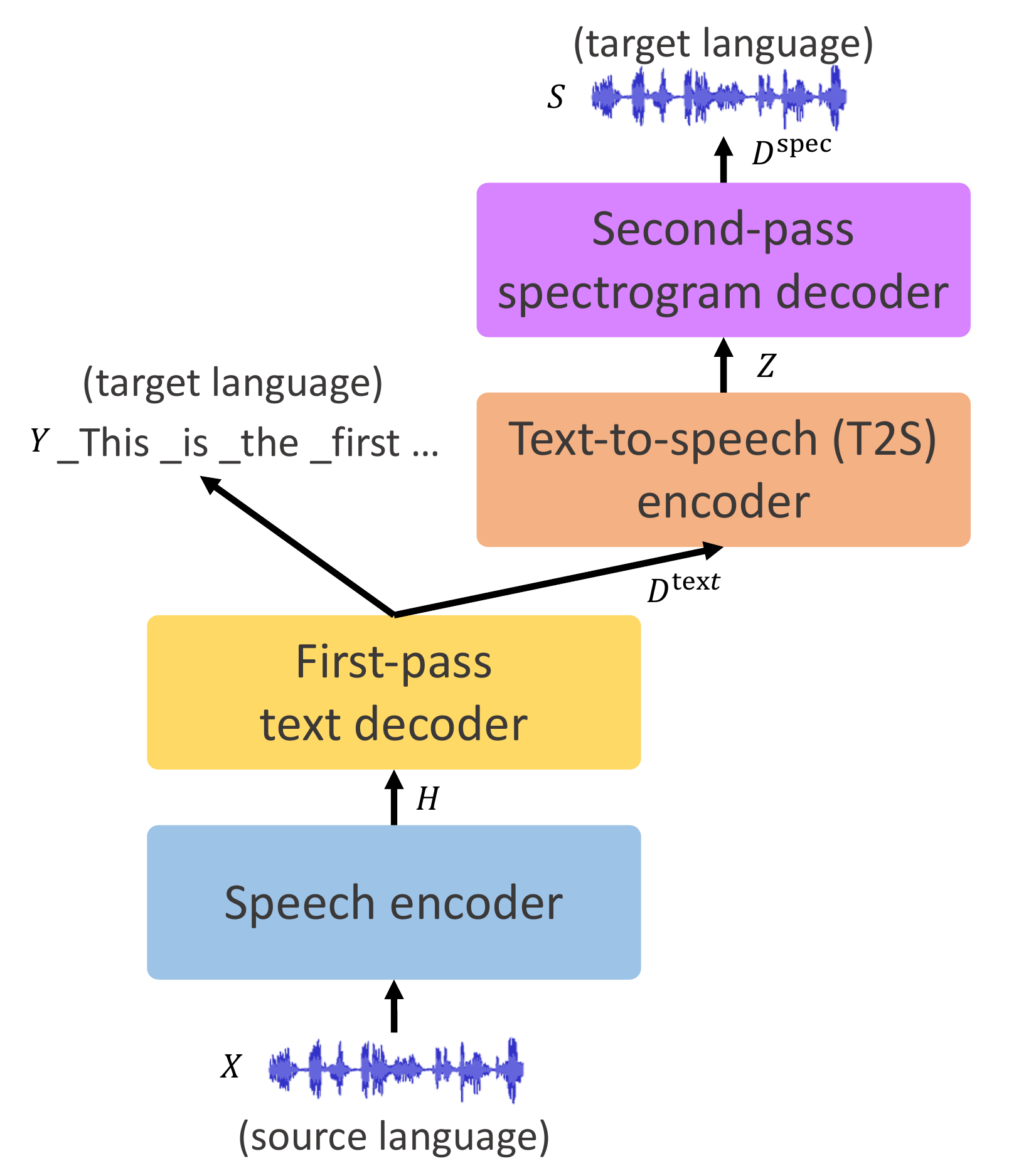}
    \subcaption{\specvvplus}
    \label{fig:specvvplus}
  \end{minipage}
  \begin{minipage}[b]{0.49\columnwidth}
    \centering
    \includegraphics[keepaspectratio,scale=0.25]{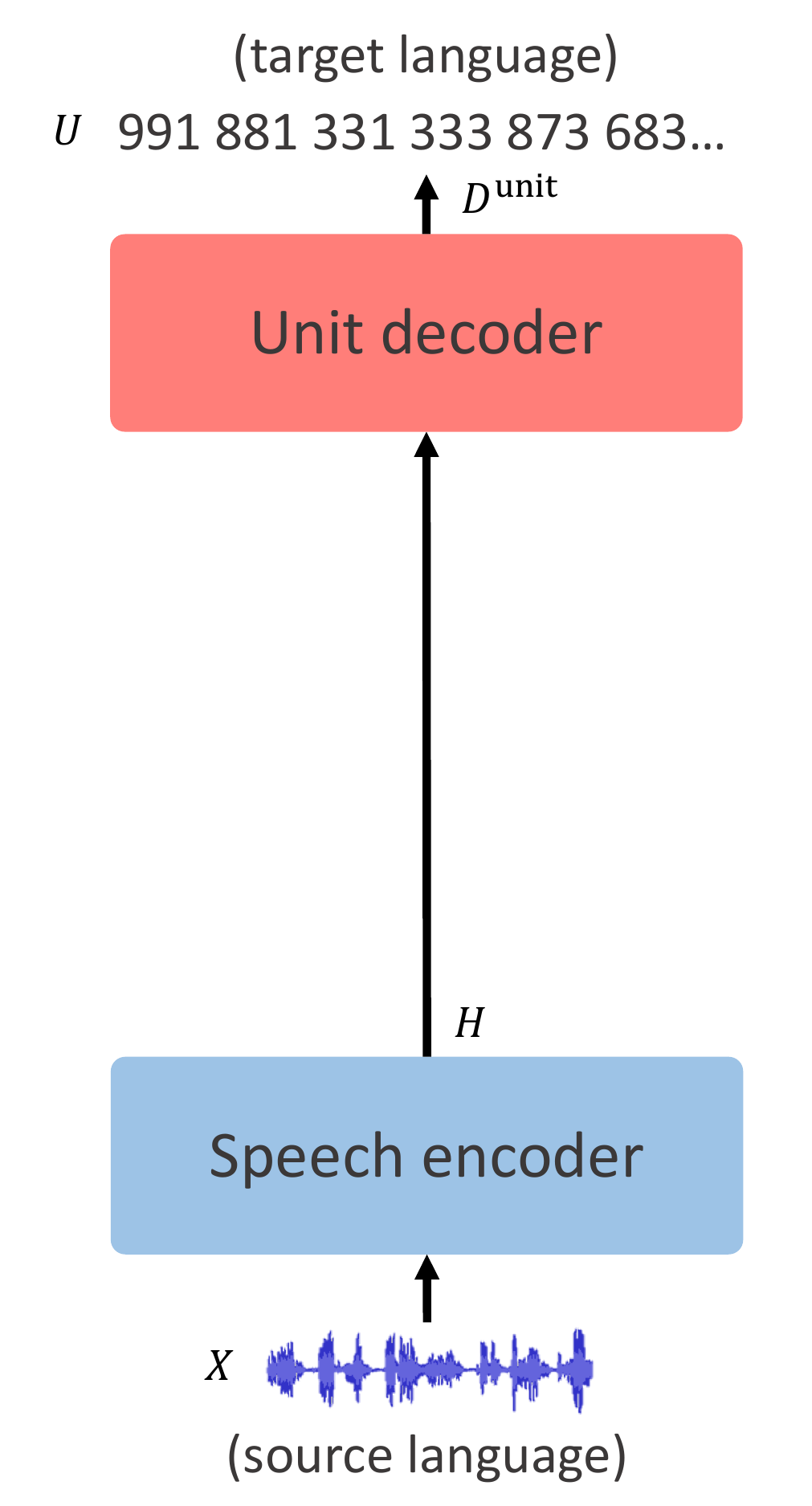}
    \subcaption{\unitv}
    \label{fig:unitv}
  \end{minipage}
  \begin{minipage}[b]{0.49\columnwidth}
    \centering
    \includegraphics[keepaspectratio,scale=0.25]{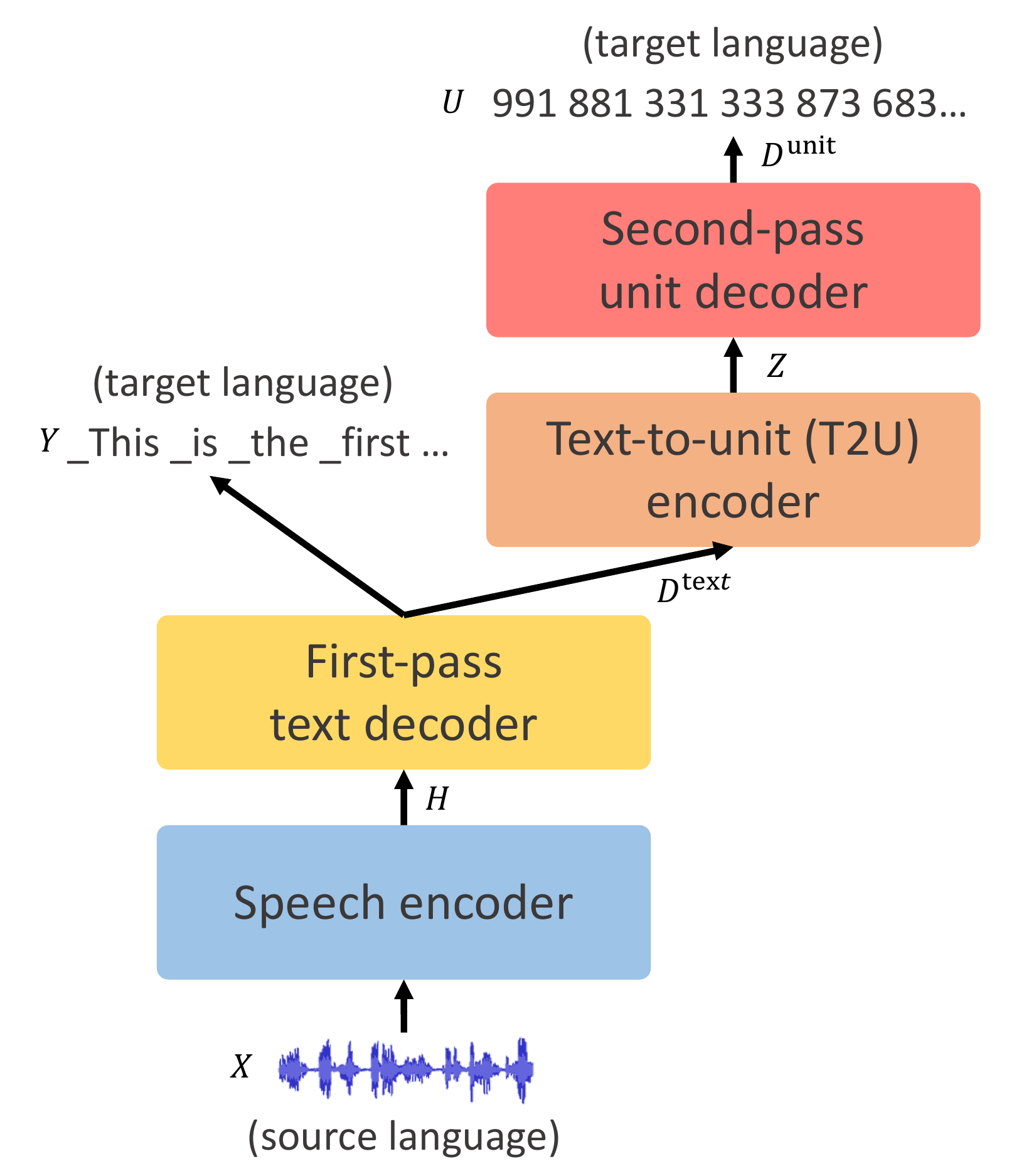}
    \subcaption{\unity}
    \label{fig:unity}
  \end{minipage}
  \caption{Direct S2ST architectures}
  \label{fig:direct_s2st_architecture}
\end{figure*}

\subsection{Text decoder pre-training}\label{ssec:text_decoder_pretraining}
Similar to ASR and S2TT studies~\citep{baevski2020wav2vec,li-etal-2021-multilingual}, S2ST models also benefit from self-supervised pre-training~\citep{jia2022leveraging,popuri2022enhanced}, especially for the speech encoder.
In addition to the speech encoder pre-training with wav2vec2.0~\citep{baevski2020wav2vec}, \citet{popuri2022enhanced} initializes the unit decoder of the single-pass {\unitv} model with a unit-based mBART (u-mBART), an encoder-decoder model pre-trained with discrete units converted from a large amount of unlabeled speech data.
However, unlabeled text data cannot be leveraged for the single-pass decoder pre-training, although it is more accessible in many written languages.

To fully leverage the unlabeled text data, we initialize the first-pass decoder of {\unity} with a text-based mBART (t-mBART) pre-trained with unlabeled text data.
Following \citet{li-etal-2021-multilingual,popuri2022enhanced}, we freeze parameters in the feed-forward network (FFN) of the text decoder during S2ST fine-tuning.
We initialize the T2U encoder and second-pass unit decoder randomly.

\subsection{Search algorithm}\label{ssec:search_algo}
During inference, we perform two-pass beam search decoding.
First, we find the most probable text hypothesis $\hyptgttext$ in the first-pass decoder using beam search with a beam size of $\beamsizetext$.
We then feed continuous decoder states $\textdecout$ corresponding to $\hyptgttext$ to the T2U encoder.
Next, we generate the most probable discrete unit sequence $\hyptgtunit$ in the second-pass decoder by another beam search with a beam size of $\beamsizeunit$.
Finally, $\hyptgtunit$ is taken as input to a separate unit-based vocoder to generate the waveform.
We find it more effective to assign a larger beam size to the first pass, {\ie}, $\beamsizetext > \beamsizeunit$, because there is more diversity among beam candidates than the second pass.
The computation time is also reduced since the sequence length of text is much shorter than that of discrete units.
Therefore, we use $\beamsizeunit=1$ unless otherwise noted.
We present the pseudo algorithm in Appendix~\ref{appendix:decoding_algo}.

\subsection{Deep-shallow two-pass decoders}\label{ssec:deep_shallow}
By increasing the number of layers, we assign more model capacities to the first-pass decoder than the second-pass decoder.
We refer to this as \textit{deep-shallow two-pass decoders}.
This capacity assignment improves translation quality and inference efficiency simultaneously because of a shorter sequence length in the first pass.
A practical capacity assignment for the MT task is studied in \citet{kasai2021deep} by trading the number of layers between the encoder and decoder.
In this work, we focus on the two-pass decoders for the S2ST task.

\section{Experimental setting}

\subsection{Data}
We use three datasets: Fisher Es$\to$En~\citep{fisher_callhome} (170 hours), CVSS-C~\citep{jia-etal-2022-cvss} (547 hours), and mutli-domain En$\leftrightarrow$Es~\citep{popuri2022enhanced} (20k hours for En$\to$Es, 14k hours for Es$\to$En) corpora.
We combine all 21 language directions to English in the CVSS-C corpus to train a single X-to-En multilingual model.
The En$\to$Es part in the multi-domain corpora consists of Europarl-ST~\citep{iranzo2020europarl}, Must-C~\citep{di-gangi-etal-2019-must}, TEDLIUM3~\citep{tedlium}, Librispeech~\citep{librispeech}, and Common Voice~\citep{ardila-etal-2020-common}.
The Es$\to$En part consists of CoVoST2~\citep{wang2021covost}, Europarl-ST, and mTEDx~\citep{elizabeth2021multilingual}, Common Voice, and multilingual Librispeech (MLS)~\citep{pratap2020mls}.
More details are described in Appendix~\ref{appendix:data}.

\vspace{-1mm}  %
\subsection{Pre-processing}
We follow the same pre-processing as ~\citep{lee2021direct,lee2021textless,popuri2022enhanced} for acoustic feature extraction, discrete unit extraction, and text normalization.
We also discarded over-generated target speech/unit by TTS/T2U models.
More details are described in Appendix~\ref{appendix:preprocessing}.

\subsection{Pre-training}
We use the same En/Es wav2vec2.0 and En-Es u-mBART models as \citet{popuri2022enhanced}.
We train a multilingual w2v-BERT~\citep{chung2021w2v} model trained on 51 languages with the same setting as ~\citet{jia2022leveraging}.
For text decoder pre-training, we use the same En-Es and 50-language t-mBART models as \citet{wang2022simple} and ~\citet{tang2020multilingual}, respectively.
We describe the training details and list model URLs in Appendix~\ref{appendix:pretraining}.

\subsection{Baseline}
We build two cascaded S2ST systems and four direct S2ST systems.
All speech encoders are based on Conformer.
When pre-training the speech encoder of direct S2ST systems with wav2vec2.0/w2v-BERT, we pre-train ASR and S2TT models in the cascaded systems with the same wav2vec2.0/w2v-BERT for a fair comparison.
We also pre-train the text decoder of the ASR and S2TT models with t-mBART in that case.

\vspace{-1mm}  %
\paragraph{Cascaded (ASR$\to$MT$\to$TTS)}
We combine a Conformer ASR, a Transformer MT, and a Transformer TTS model.
We set the reduction factor of TTS models to 4.

\vspace{-1mm}  %
\paragraph{Cascaded (S2TT$\to$TTS)}
We combine a Conformer direct S2TT model and a Transformer TTS model.

\vspace{-1mm}  %
\paragraph{\specv}
We build a direct S2ST model that predicts spectrogram with a single Transformer decoder, similar to \citet{lee2021direct} (Figure~\ref{fig:specv}).
We refer to it as {\specv} hereafter.
We set the reduction factor of the spectrogram decoder to 3.

\vspace{-1mm}  %
\paragraph{\specvvplus}
We train {\specvvplus}, an improved version of {\translatotronvv}, by enhancing the architecture and training with the proposed methods for {\unity}.
First, we replace phoneme targets with subwords in the first pass (Figure~\ref{fig:specvvplus}).
Second, we replace LSTM decoders with Transformer decoders.
Third, we introduce an additional text-to-spectrogram (T2S) encoder between text and spectrogram decoders.
The second-pass decoder attends to the T2S encoder output only.
Fourth, we use an autoregressive Transformer decoder instead of a non-attentive Tacotron (NAT)~\citep{shen2020non} for the second-pass decoder.
Last, we apply R-Drop to the first-pass decoder.
We use the same reduction factor as {\specv}.

\paragraph{\unitv}
We train a direct S2ST model that predicts discrete units with a single Transformer decoder~\citep{lee2021direct} (Figure~\ref{fig:unitv}).

\subsection{Architecture}
Let $\nlayerfirstdec$, $\nlayerseconddec$, and $\nlayerttuenc$ be the depth of the first-pass decoder, second-pass decoder, and T2U encoder of {\unity}, respectively.
We set $(\nlayerfirstdec, \nlayerseconddec, \nlayerttuenc)$ to $(4, 2, 2)$ on Fisher and CVSS-C.
On the multi-domain corpus, we use $(12, 2, 2)$ when pre-training the first-pass decoder with t-mBART.
Otherwise, we use $(6, 6, 2)$.
We describe the other configurations in Appendix~\ref{appendix:hyperparameter_architecture}.

\subsection{Training}
We apply R-Drop~\citep{wu2021rdrop} regularization to all tasks that predict discrete symbols, except the MT task.
The training objective of each model with R-Drop is defined in Appendix~\ref{appendix:training_objective}.
We implement our models based on the Fairseq toolkit~\citep{ott2019fairseq,wang2020fairseq}.
The detailed training hyperparameters are described in Appendix~\ref{appendix:hyperparameter_training}.

\subsection{Decoding}
We use a beam width of 10 for ASR, S2TT, and {\unitv} models.
For {\unity}, we set $\beamsizetext$ and $\beamsizeunit$ to 10 and 1, respectively.
We use a beam width of 10 for the first-pass decoder in {\specvvplus}.

\subsection{Vocoder}
We use a HiFi-GAN vocoder~\citep{kong2020hifi} to convert spectrograms to the waveform for TTS and direct speech-to-spectrogram models.
We use a unit-based HiFi-GAN vocoder~\citep{polyak21_interspeech} to convert discrete units to the waveform for direct speech-to-unit models.
Both the vocoders are trained separately.

\subsection{Evaluation}
Following \citet{lee2021direct}, we use a pre-trained ASR model to transcribe the generated target speech and calculate BLEU scores~\citep{papineni-etal-2002-bleu}, referred to as ASR-BLEU.
The ASR model is fine-tuned from a wav2vec2.0 with the connectionist temporal classification (CTC) objective~\citep{ctc_graves}.
We use the sacrebleu toolkit~\citep{post-2018-call} to calculate the BLEU scores.

\begin{table}[t]
    \centering
    \begingroup
    \scalebox{0.74}{
    \begin{tabular}{llllccc} \toprule
    \multirow{2}{*}{ID} & \multirow{2}{*}{Model} & \multicolumn{4}{c}{ASR-BLEU ($\uparrow$)} \\ \cmidrule(lr){3-6}
     & & Avg. & High & Mid & Low \\
     \midrule[\heavyrulewidth]
     \texttt{B0} & Synthetic target${}^{\diamondsuit}$ & 91.1 & 88.4 & 89.5 & 93.0 \\
     \midrule[\heavyrulewidth]

     \multicolumn{6}{l}{\textbf {Cascaded systems}} \\
     \texttt{B1} & S2TT $\to$ TTS${}^{\diamondsuit}$ & 10.6 & 28.8 & 15.5 & \phantom{a}2.4 \\
     \texttt{B2} & \ + ASR pre-training & 12.7 & 30.7 & 18.3 & \phantom{a}4.4 \\
     \texttt{B3} & S2TT $\to$ TTS & \phantom{a}7.8 & 18.2 & 11.9 & \phantom{a}2.6 \\  %
     \texttt{B4} & \ + w2v-BERT + t-mBART & 14.9 & 21.1 & 18.2 & 11.5 \\  %
     \midrule[\heavyrulewidth]

     \multicolumn{6}{l}{\textbf {Direct speech-to-spectrogram systems}} \\
     \texttt{B5} & {\translatotronv}${}^{\diamondsuit}$ & \phantom{a}3.4 & 11.9 & \phantom{a}3.5 & \phantom{a}0.3 \\
     \texttt{B6} & {\specv} & \phantom{a}7.6 & 21.8 & 10.6 & \phantom{a}1.5 \\
     \texttt{B7} & \ + S2TT pre-training & \phantom{a}9.6 & 23.9 & 13.8 & \phantom{a}3.2 \\
     \texttt{B8} & \ + w2v-BERT & 16.6 & 30.5 & 21.9 & \phantom{a}9.8 \\  %
     \cmidrule(lr){1-6}

     \texttt{B9} & {\translatotronvv}${}^{\diamondsuit}$ & \phantom{a}8.7 & 25.4 & 12.6 & \phantom{a}1.5 \\
     \texttt{B10} & \ + Transformer decoder${}^{\spadesuit}$ & 10.1 & 26.9 & 14.2 & \phantom{a}2.8 \\
     \texttt{B11} & \ + S2TT pre-training${}^{\diamondsuit}$ & 12.0 & 29.7 & 16.6 & \phantom{a}4.2 \\
     \texttt{B12} & \ + w2v-BERT${}^{\spadesuit}$ & 17.9 & 32.5 & 22.9 & 10.9 \\
     \texttt{B13} & \ + mSLAM${}^{\spadesuit}$ & 19.3 & 33.2 & 24.6 & 12.5 \\
     \texttt{B14} & \ \ ++ TTS augmentation${}^{\spadesuit}$ & 22.0 & 33.5 & 25.8 & 16.5 \\
     \cmidrule(lr){1-6}

     \texttt{B15} & {\specvvplus} & 11.3 & 29.1 & 16.9 & \phantom{a}3.1 \\  %
     \texttt{B16} & \ + S2TT pre-training & 13.1 & 29.8 & 18.8 & \phantom{a}5.2 \\  %
     \texttt{B17} & \ + w2v-BERT + t-mBART & 18.6 & 32.1 & 24.7 & 11.6 \\  %
     \midrule[\heavyrulewidth]

     \multicolumn{6}{l}{\textbf {Direct speech-to-unit systems}} \\
     \texttt{B18} & {\unitv} & \phantom{a}9.1 & 25.9 & 12.9 & \phantom{a}1.9 \\
     \texttt{B19} & \ + S2TT pre-training & 11.4 & 27.2 & 16.4 & \phantom{a}4.0 \\
     \texttt{B20} & \ + w2v-BERT + u-mBART & 20.8 & 31.6 & 25.4 & 15.4 \\  %
     \cmidrule(lr){1-6}

     \texttt{B21} & {\unity} & 12.0 & 29.0 & 17.8 & \phantom{a}4.0 \\
     \texttt{B22} & \ + S2TT pre-training & 13.0 & 30.4 & 18.7 & \phantom{a}4.8 \\
     \texttt{B23} & \ + w2v-BERT + t-mBART & {\bf 24.5} & {\bf 34.6} & {\bf 28.9} & {\bf 19.3} \\  %
     \bottomrule
    \end{tabular}
    }
    \endgroup
    \vspace{-1mm}
    \caption{ASR-BLEU on CVSS-C corpus. ${}^{\diamondsuit}$Results from ~\citep{jia-etal-2022-cvss}, ${}^{\spadesuit}$Results from ~\citep{jia2022leveraging}. We use the S2TT model in \texttt{B3} for S2TT pre-training. t-mBART and u-mBART stand for text-based mBART and unit-based mBART, respectively. All w2v-BERT and mSLAM encoders have 0.6B parameters.}
    \label{tab:results_cvss_c_s2st}
\end{table}

\begin{table*}[t!]
    \centering
    \begingroup
    \scalebox{0.85}{
    \resizebox{\linewidth}{!}{
    \begin{tabular}{llccccccccccccc} \toprule
        \multirow{3}{*}{ID} & \multirow{3}{*}{Model} & \multicolumn{7}{c}{ASR-BLEU ($\uparrow$)} \\
        \cmidrule(lr){3-9}

        & & \multicolumn{3}{c}{{\bf En$\to$Es}} & \multicolumn{4}{c}{{\bf Es$\to$En}} \\
        \cmidrule(lr){3-5} \cmidrule(lr){6-9}
        & & Europarl-ST & MuST-C & Avg. & CoVoST-2 & Europarl-ST & mTEDx & Avg. \\
        \midrule[\heavyrulewidth]

        \multicolumn{8}{l}{\textbf {Cascaded systems}} \\
        \texttt{C1} & ASR$\to$MT$\to$TTS${}^{\diamondsuit}$ & 28.8 & 34.2 & 31.5 & 33.8 & 29.1 & 32.4 & 31.5 \\
        \texttt{C1'} & ASR$\to$MT$\to$TTS & 36.8 & 30.8 & 33.8 & 32.9 & 34.2 & 30.3 & 32.5 \\
        \cmidrule(lr){1-9}

        \texttt{C2} & S2TT$\to$TTS${}^{\diamondsuit}$ & 32.6 & 30.1 & 31.4 & 28.4 & 23.6 & 21.5 & 24.5 \\ %
        \texttt{C2'} & S2TT$\to$TTS & 36.4 & 33.4 & 34.9 & 37.2 & 34.0 & 32.5 & 34.6 \\
        \midrule[\heavyrulewidth]

        \multicolumn{8}{l}{\textbf {Direct speech-to-spectrogram systems}} \\
        \texttt{C3} & {\specvvplus} (6L$\to$6L) & 35.6 & 33.5 & 34.6 & 37.0 & 23.4 & 31.3 & 30.6 \\
        \texttt{C4} & \ + t-mBART (12L$\to$6L) & {\bf 36.9} & {\bf 34.3} & {\bf 35.6} & {\bf 37.2} & 23.7 & 31.7 & 30.9 \\
        \midrule[\heavyrulewidth]

        \multicolumn{8}{l}{\textbf {Direct speech-to-unit systems}} \\
        \texttt{C5} & {\unitv} + u-mBART${}^{\diamondsuit}$ & 32.7 & 32.1 & 32.4 & 33.5 & 28.6 & 29.1 & 30.4 \\  %
        \texttt{C5'} & {\unitv} + u-mBART & 33.5 & 33.3 & 33.4 & 34.5 & 29.9 & 29.9 & 31.4 \\
        \cmidrule(lr){1-9}

        \texttt{C6} & {\unity} (6L$\to$6L) & 35.1 & 33.7 & 34.4 & 35.4 & 30.8 & 31.3 & 32.5 \\
        \texttt{C7} & \ + t-mBART (12L$\to$2L) & {\bf 35.3} & {\bf 34.1} & {\bf 34.7} & {\bf 36.4} & {\bf 33.1} & {\bf 32.2} & {\bf 33.9} \\ %
    \bottomrule
    \end{tabular}
    }
    }
    \endgroup
    \vspace{-1mm}
    \caption{ASR-BLEU on multi-domain En$\leftrightarrow$Es. ${}^{\diamondsuit}$Results from ~\citep{popuri2022enhanced}. The encoder in all the models is pre-trained with wav2vec2.0. t-mBART and u-mBART stand for text-based mBART and unit-based mBART, respectively. $\nlayerfirstdec$L$\to\nlayerseconddec$L stands for an $\nlayerfirstdec$-layer first-pass decoder with an $\nlayerseconddec$-layer second-pass decoder.}
    \label{tab:results_enes_s2st}
\end{table*}

\section{Experimental results}\label{sec:result}
In this section, we present the experimental results on three corpora.
We study various modeling choices from the perspective of target representation (spectrogram v.s. discrete unit) and decoder architectures (single pass v.s. two pass) in supervised and semi-supervised settings.
We also benchmark the decoding efficiency of direct S2ST models.

\subsection{CVSS-C}\label{ssec:result_cvss}
The results on CVSS-C are listed in Table~\ref{tab:results_cvss_c_s2st}.
We first compared four direct systems trained from scratch (\texttt{B6}, \texttt{B15}, \texttt{B18}, \texttt{B21}), and {\unity} (\texttt{B21}) achieved the best ASR-BLEU.
The encoder pre-training with the S2TT model in the cascaded system (\texttt{B3}) improved ASR-BLEU of all the direct S2ST models (\texttt{B7}, \texttt{B16}, \texttt{B19}, \texttt{B22}), similar to \citet{jia-etal-2022-cvss}.\footnote{Unlike \citet{jia-etal-2022-cvss}, we trained the S2TT model with an auxiliary ASR task from scratch instead of pre-training the encoder with that of an ASR model.}
In this case, {\specvvplus} (\texttt{B16}) also achieved similar translation quality to {\unity} (\texttt{B22}).
Still, {\unity} outperformed the {\unitv} model (\texttt{B19}) by 1.6 ASR-BLEU on average, indicating that the two-pass decoding was the main factor of the improvements.
{\specvvplus} (\texttt{B16}) outperformed {\translatotronvv}~\citep{jia2022translatotron} (\texttt{B11}) by 1.1 ASR-BLEU on average, from which we can confirm that parts of the proposed methods can generalize to the other S2ST architecture.\footnote{\texttt{B9} predicts phonemes while \texttt{B15} predicts subwords in the first pass.}
Compared to the best cascaded system (\texttt{B2}), the two-pass models (\texttt{B16}, \texttt{B19}) showed better translation quality.

We also pre-trained the speech encoder of all models with multilingual w2v-BERT, the first-pass text decoder of two-pass models with text-based mBART (t-mBART), and the decoder of the S2UT model with unit-based mBART (u-mBART), respectively.
Among them (\texttt{B4}, \texttt{B8}, \texttt{B12}, \texttt{B20}, \texttt{B23}), {\unity} (\texttt{B23}) showed the best ASR-BLEU.
{\unity} still outperformed {\translatotronvv} with a joint speech-text pre-training with mSLAM~\citep{bapna2022mslam} (\texttt{B13}) and TTS augmentation (\texttt{B14}) by 5.2 and 2.5 ASR-BLEU on average, respectively.
The full results in each language direction are presented in Appendix~\ref{appendix:result_cvss_detail}.

\subsection{Multi-domain En$\leftrightarrow$Es}\label{ssec:result_enes}
We present results on the multi-domain corpora~\citep{popuri2022enhanced} in Table~\ref{tab:results_enes_s2st}.
\texttt{C1'}, \texttt{C2'}, and \texttt{C5'} are our improved models of \texttt{C1}, \texttt{C2}, and \texttt{C5}, respectively.\footnote{We improved \texttt{C1} and \texttt{C2} by R-Drop and better hyperparameters. \texttt{C5} was also improved by hyperparameter tuning and checkpoint averaging.}
We observed that {\unity} with first-pass decoder pre-training with t-mBART (\texttt{C7}) improved the {\unitv} model with decoder pre-training with u-mBART (\texttt{C5'}) by 1.3 and 2.5 ASR-BLEU on average in En$\to$Es and Es$\to$En, respectively.
This confirms the effectiveness of the two-pass modeling in the high-resource scenario.
Furthermore, {\unity} without decoder pre-training (\texttt{C6}) already outperformed \texttt{C5'} and degraded from \texttt{C7} only slightly.
Comparing {\unity} and {\specvvplus}, we cannot spot a clear winner.
{\unity} outperformed {\specvvplus} in Es$\to$En on Europarl-ST and mTEDx, but {\specvvplus} performed better in En$\to$Es.
The proposed text decoder pre-training helped {\specvvplus} performance too, especially in En$\to$Es (\texttt{C4}).
Finally, we also confirmed that {\unity} approached the performance of a strong cascaded system and even outperformed it on Must-C.

\begin{figure}[t]
  \centering
  \includegraphics[width=0.99\linewidth]{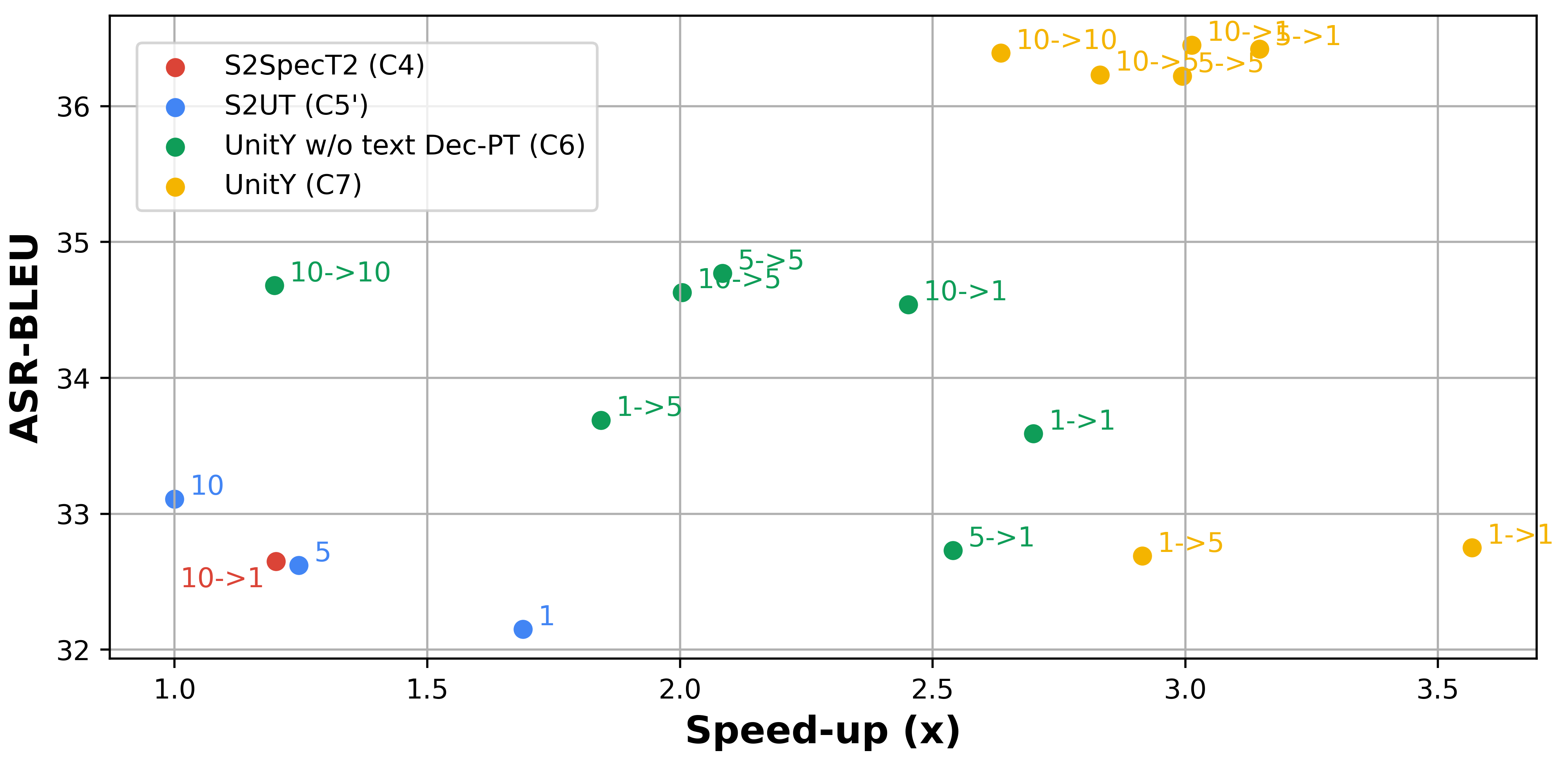}
  \vspace{-6mm}
  \caption{Runtime of direct S2ST models on multi-domain Es$\to$En corpus. X$\to$Y at each data point represents the beam width in each decoder pass.}
  \label{fig:decoding_speed}
\end{figure}

\subsection{Decoding efficiency}\label{ssec:decoding_efficiency}
We evaluated the decoding efficiency of direct S2ST models.
We measured the runtime and total number of floating point operations (FLOPs) on an Intel® Xeon® Gold 6230 CPU.
We randomly sampled 500 utterances from the multi-domain Es$\to$En dev set while keeping the ratio of the number of samples per domain.
Note that we also took the vocoder inference into account.

The results in Figure~\ref{fig:decoding_speed} showed that {\unity} achieved {\speedupoverspec}$\times$ and {\speedupoverstut}$\times$ decoding speed-ups over {\specvvplus} and {\unitv} models, respectively.
These confirms the efficiency of discrete unit prediction and two-pass decoding, thanks to reduced output sequence lengths.
Deep-shallow two-pass decoders also improved the decoding speed a lot.
We found that the translation quality of the two-pass models improved by increasing the beam width of the first-pass decoder up to 10.
On the other hand, the quality did not degrade significantly by decreasing the beam width of the second-pass decoder down to 1, {\ie} greedy decoding.
This indicates that the first pass involves more challenges in the modeling pipeline.
Therefore, we can obtain better translation quality and decoding speed by assigning more computation time to the first pass.

We also present the results of FLOPs in Appendix~\ref{appendix:result_flops}.
To summarize, {\unity} achieved {\flopsreductionoverspec}$\times$ and {\flopsreductionoverstut}$\times$ FLOPs reduction over {\specvvplus} and {\unitv} models, respectively.

\subsection{Fisher}
We also show the results on Fisher in Appendix~\ref{appendix:result_fisher}.
Although the trend was consistent with CVSS-C, a notable exception was that {\specvvplus} outperformed {\unity} when pre-training the speech encoder with wav2vec2.0.
However, {\unity} has an advantage of decoding efficiency over {\specvvplus}.

\section{Analysis}
In this section, we conduct analyses to shed light on the source of improvements in {\unity}.
We also study whether the same techniques used for {\unity} are helpful for {\specvvplus}.
We use the multi-domain Es$\to$En corpus, but pseudo-labeled ASR data is excluded for quick exploration, resulting in 196-hour source speech.
We report average dev scores over three runs with different random seeds.\footnote{We removed three long utterances from the mTEDx dev set to fit the GPU memory.}

\begin{table}[t]
    \centering
    \begingroup
    \scalebox{0.72}{
    \begin{tabular}{llcc} \toprule
     \multirow{3}{*}{ID} & \multirow{3}{*}{Model} & \multicolumn{2}{c}{(ASR-)BLEU ($\uparrow$)} \\
     \cmidrule(lr){3-4}
     & & Text & Speech \\
     \midrule
      \texttt{D1} & {\specvvplus} & {\bf 35.0}  & {\bf 30.8}\\
      \texttt{D2} & \ + w/o T2S encoder & 34.9 & 25.0 \\
      \texttt{D3} & \ + w/o R-Drop & 34.8 & 30.3 \\
     \midrule
      \texttt{D5} & {\unity} & {\bf 38.3} & {\bf 33.2} \\
      \texttt{D6} & \ + w/o T2U encoder & 38.1 & 30.7 \\
      \texttt{D7} & \ + w/o R-Drop & 37.7 & 32.1 \\
      \texttt{D8} & \ + Cross-attn to speech enc (sequential) & 38.2 & {\bf 33.2} \\
      \texttt{D9} & \ + Cross-attn to speech enc (parallel) & 38.1 & 33.1 \\
     \bottomrule
    \end{tabular}
    }
    \endgroup
    \vspace{-1mm}
    \caption{Ablation study for two-pass direct S2ST models on multi-domain Es$\to$En dev set. The first-pass decoder in all the models is pre-trained with t-mBART.}
    \label{tab:ablation_study}
\end{table}

\subsection{Ablation study}\label{ssec:result_ablation_study}
We first conducted an ablation study for two-pass direct S2ST models in Table~\ref{tab:ablation_study}.
We evaluated the translation quality of outputs from both decoders.
An additional T2U/T2S encoder was essential for bridging the gap in representations between the first-pass and second-pass decoders, especially for {\specvvplus} (\texttt{D2}, \texttt{D6}).
We attribute this to the fact that the gap in representations between text and spectrogram is larger than between text and discrete units.
R-Drop was also beneficial for boosting the translation quality of the first-pass decoder, which improved the final performance accordingly (\texttt{D3}, \texttt{D7}).
Moreover, we investigated adding another cross-attention over the speech encoder output to the unit decoder, as discussed in \cref{ssec:architecture}.
We expected that the first-pass decoder output lost useful information to generate target speech faithful to source speech.
We explored parallel (\textit{parallel}, \texttt{D8}) and sequential (\textit{sequential}, \texttt{D9}) cross-attention, similar to~\citep{zhu2019incorporating}, but neither showed any improvement.
The first-pass decoder already extracted source acoustic information well via multiple cross-attention modules.
We also show the results on Fisher in Appendix~\ref{appendix:result_ablation_study}.

\begin{table}[t]
    \centering
    \begingroup
    \scalebox{0.70}{
    \begin{tabular}{lllcccc} \toprule
     \multirow{2}{*}{ID} & \multirow{2}{*}{Model} & \multirow{2}{*}{\shortstack{Output\\unit}} & \multicolumn{2}{c}{(ASR-)BLEU ($\uparrow$)} & \multirow{2}{*}{\shortstack{Speed-up\\($\times$)}} \\
     \cmidrule(lr){4-5}
     & & & Text & Speech & \\
     \midrule
     \texttt{E1} & \multirow{3}{*}{\specvvplus} & Phoneme & -- & 29.4 & 1.00 \\
     \texttt{E2} & & Character & 31.7 & 28.9 & 0.89 \\
     \texttt{E3} & & Subword & {\bf 33.0} & {\bf 30.0} & {\bf 1.12} \\
     \midrule

     \texttt{E4} & \multirow{3}{*}{\unity} & Phoneme & -- & 27.8 & 2.31 \\
     \texttt{E5} & & Character & 33.2 & 29.6 & 2.06 \\
     \texttt{E6} & & Subword & {\bf 34.1} & {\bf 30.1} & {\bf 2.86} \\
     \bottomrule
    \end{tabular}
    }
    \endgroup
    \vspace{-1mm}
    \caption{Results of output units for the first-pass decoder in two-pass direct S2ST models on multi-domain Es$\to$En dev set. The first-pass decoder in all the models is initialized randomly.}
    \label{tab:output_unit}
    \vspace{2mm}
\end{table}

\subsection{Output unit for first-pass decoder}\label{ssec:result_output_unit}
We studied optimal granularity of the output unit for the first-pass decoder in two-pass direct S2ST models.
We explored phonemes, characters, and 2k subwords units.
The results in Table~\ref{tab:output_unit} showed that the subword unit (\texttt{E6}) was the most effective for the first-pass decoder in both {\unity} and {\specvvplus} thanks to a better translation quality.
Moreover, it gained the largest decoding speed-up.
We also show the results on Fisher in Appendix~\ref{appendix:result_output_unit}.

\begin{table}[t]
    \centering
    \begingroup
     \scalebox{0.70}{
    \begin{tabular}{lcccccc} \toprule
     \multirow{4}{*}{ID} & \multicolumn{2}{c}{Decoder depth} & \multirow{4}{*}{\shortstack{\#Params\\(Billion)}} & \multicolumn{2}{c}{(ASR-)BLEU ($\uparrow$)} & \multirow{4}{*}{\shortstack{Speed-up\\($\times$)}} \\
     \cmidrule(lr){2-3} \cmidrule(lr){5-6}
     & \multirow{3}{*}{\shortstack{First\\pass\\(text)}} & \multirow{3}{*}{\shortstack{Second\\pass\\(unit)}} & & \multirow{3}{*}{Text} & \multirow{3}{*}{Speech} & \\
     & & & & & & \\
      & & & & & & \\
    \midrule
     \texttt{G1} & \phantom{a}2\phantom{${}^{\spadesuit}$} & \phantom{a}6\phantom{${}^{\spadesuit}$} & 0.79 & 34.5 & 30.3 & 1.24 \\
     \texttt{G2} & \phantom{a}4\phantom{${}^{\spadesuit}$} & \phantom{a}6\phantom{${}^{\spadesuit}$} & 0.82 & 34.5 & 30.5 & 1.20 \\
     \cmidrule(lr){1-7}

     \texttt{G3} & \phantom{a}6\phantom{${}^{\spadesuit}$} & \phantom{a}2\phantom{${}^{\spadesuit}$} & 0.79 & 34.3 & 30.3 & 1.47 \\
     \texttt{G4} & \phantom{a}6\phantom{${}^{\spadesuit}$} & \phantom{a}4\phantom{${}^{\spadesuit}$} & 0.82 & 33.9 & 29.9 & 1.19 \\
     \texttt{G5} & \phantom{a}6\phantom{${}^{\spadesuit}$} & \phantom{a}6\phantom{${}^{\spadesuit}$} & 0.86 & {\bf 34.8} & {\bf 30.7} & 1.00 \\
     \texttt{G6} & \phantom{a}6\phantom{${}^{\spadesuit}$} & \phantom{a}8\phantom{${}^{\spadesuit}$} & 0.89 & 34.2 & 30.2 & 0.69 \\
     \texttt{G7} & \phantom{a}6\phantom{${}^{\spadesuit}$} & 12${}^{\diamondsuit}$ & 0.96 & 33.7 & 29.8 & 0.68 \\ %
     \midrule[\heavyrulewidth]

     \texttt{G8} & 12\phantom{${}^{\spadesuit}$} & \phantom{a}2\phantom{${}^{\spadesuit}$} & 0.95 & {\bf 34.9} & {\bf 30.7} & {\bf 1.44} \\
     \texttt{G9} & 12${}^{\spadesuit}$ & \phantom{a}2\phantom{${}^{\spadesuit}$} & 0.95 & {\bf 38.3} & {\bf 33.2} & {\bf1.19} \\ %
     \texttt{G10} & 12${}^{\spadesuit}$ & \phantom{a}4\phantom{${}^{\spadesuit}$} & 0.98 & 38.0 & 33.0 & 1.09 \\
     \texttt{G11} & 12${}^{\spadesuit}$ & \phantom{a}6\phantom{${}^{\spadesuit}$} & 1.00 & 38.1 & 33.1 & 0.84 \\
     \texttt{G12} & 12${}^{\spadesuit}$ & 12${}^{\diamondsuit}$ & 1.12 & 36.2 & 32.2 & 0.60 \\  %
     \bottomrule
    \end{tabular}
    }
    \endgroup
    \vspace{-1mm}
    \caption{Results of capacity assignment to two-pass decoders in {\unity} on multi-domain Es$\to$En dev set. ${}^{\spadesuit}$Pre-trained with t-mBART. ${}^{\diamondsuit}$Pre-trained with u-mBART. \texttt{G1-G8} have a 2k subword vocabulary, and \texttt{G9-G12} have a 65k subword vocabulary.}
    \label{tab:decoder_capacity}
\end{table}

\subsection{Capacity assignment to two-pass decoders}\label{ssec:result_capacity_assignment}
We sought to effectively assign the model capacity to the two decoders in {\unity} to obtain a better translation quality.
The results in Table~\ref{tab:decoder_capacity} showed that a 12-layer text decoder with a two-layer unit decoder (\texttt{G8}) was the best in translation quality and decoding speed when initializing the first-pass decoder randomly (\texttt{G1}-\texttt{G6},\texttt{G8}).
Pre-training the first-pass decoder with t-mBART (\texttt{G9}) brought a large ASR-BLEU gain with a slight speed degradation compared to \texttt{G8}.\footnote{We set the depth of the first-pass decoder to 12 because of the availability of the off-the-shelf t-mBART model.}
It was sufficient to have a two-layer unit decoder in that case (\texttt{G9}-\texttt{G11}).
We also pre-trained the second-pass decoder with u-mBART while initializing the text decoder randomly (\texttt{G7}) or with t-mBART (\texttt{G12}), but neither improved the performance further.
Therefore, it is most effective to pre-train the deep text decoder only and keep the unit decoder shallow.
Note that \texttt{G8} is faster than \texttt{G9} because of the smaller subword vocabulary size (2k v.s. 65k).

\subsection{Data scale}\label{ssec:result_data_scale}
\begin{figure}[t]
  \centering
  \includegraphics[width=0.99\linewidth]{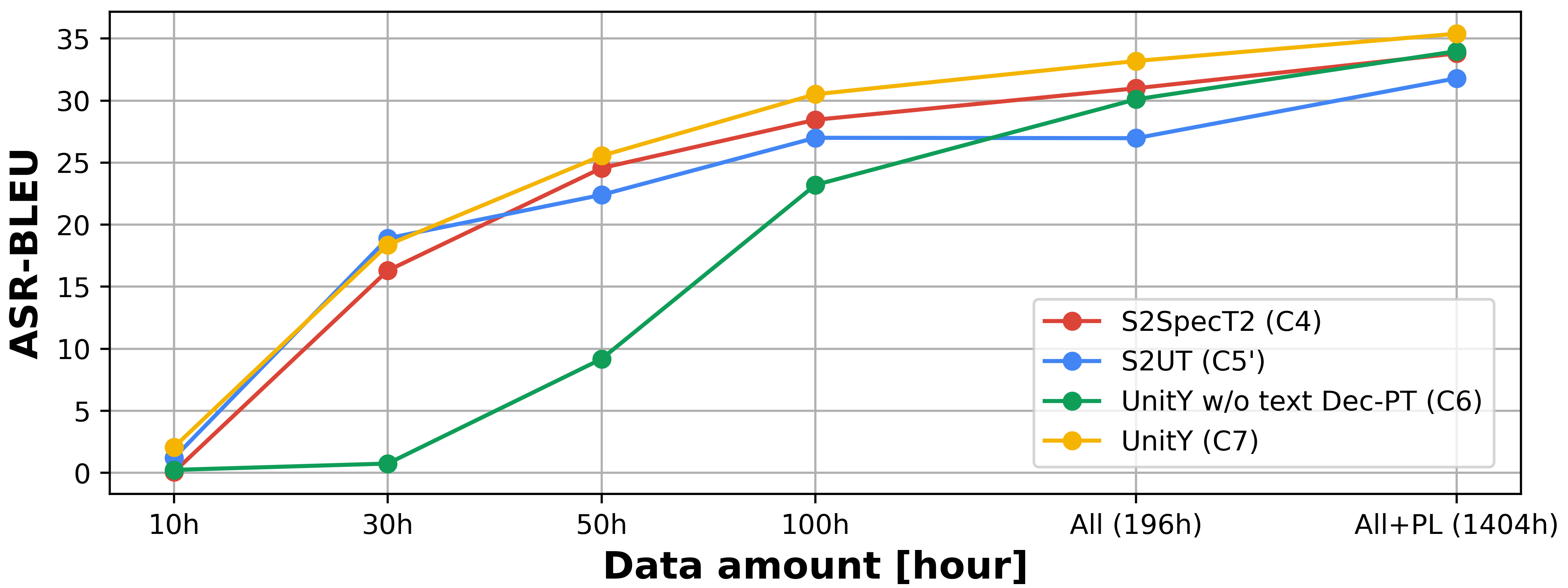}
  \vspace{-5mm}
  \caption{Dev ASR-BLEU at different data scales on the multi-domain Es$\to$En corpus. The amount of training data is measured by source speech. \textit{All} and \textit{PL} represent all supervised data and pseudo-labeled data, respectively.}
  \label{fig:data_scale}
\end{figure}

Improving the translation quality of S2ST models on low-resource data is crucial since collecting a large amount of training data is challenging.
We compared translation quality of direct S2ST models at various training data scales in Figure~\ref{fig:data_scale}.
We observed that {\unity} consistently outperformed the {\specvvplus} and {\unitv} models when the data size was no less than 50 hours.
The text decoder pre-training became less effective as the data size increased, consistent with an observation in \cref{ssec:result_enes}, where the improvement in En$\to$Es (+1.3) was smaller than Es$\to$En (+2.5).
However, pre-training the text decoder of {\unity} was essential for obtaining decent performances in the low-resource settings ($\leq$ 50 hours).

\section{Related works}

\vspace{-1mm}  %
\paragraph{Two-pass sequence generation}

Two-pass decoding has advantages of maintaining the end-to-end optimization capability while inheriting the benefits of a cascading approach.
\citet{xia2017deliberation,hu2020deliberation} incorporate an additional search process to find a better output.
\citet{dalmia-etal-2021-searchable} reranks the intermediate hypotheses using an external module such as a language model.
\citet{zhao19d_interspeech} injects specific information in the intermediate decoder to bias the output toward the desired domain.
\citet{sainath2019two} provides an intermediate output to users before generating the final output for streaming applications.
The two-pass approach makes the optimization tractable, which has advanced performance of speech translation models~\citep{anastasopoulos-chiang-2018-tied,sperber-etal-2019-attention,sung2019towards,dalmia-etal-2021-searchable,inaguma2021fast,yan-etal-2022-cmus,jia2022translatotron}.

\vspace{-1mm}  %
\paragraph{Direct speech-to-spectrogram translation}
{\translatotronv}~\citep{jia2019direct} is the first direct S2ST model but suffered from poor performance even with auxiliary ASR and S2TT tasks.
\citet{kano2021transformer} subsequently pre-trains the components with ASR and S2TT models, which is more effective for distant language pairs.
{\translatotronvv}~\citep{jia2022translatotron} significantly improves {\translatotronv} by incorporating two-pass decoding.
We showed that our methods further improved {\translatotronvv}.

\vspace{-1mm}  %
\paragraph{Direct speech-to-unit translation}
Direct speech-to-unit translation models predict discrete units rather than spectrogram.
\citet{tjandra2019speech} uses vector-quantized variational autoencoder~\citep{van2017neural} while \citet{lee2021direct} used HuBERT~\citep{hsu2021hubert} to extract target discrete units.
\citet{lee2021textless} normalizes speaker identity of real target speech using a CTC-based speech-to-unit model.
\citet{huang2022transpeech} further improves the normalization by considering rhythm, pitch, and energy.

\section{Conclusion}
We proposed {\unity}, a novel efficient two-pass direct S2ST model that subsequently generates both text and discrete unit outputs.
We improved the model performance by predicting subwords in the first pass, bridging decoder representations by an additional encoder, deep-shallow two-pass decoders, regularizing the training with R-Drop, and pre-training the first-pass decoder with text-based mBART.
Experimental evaluations demonstrated that {\unity} outperformed a single-pass {\unitv} model consistently in translation quality and inference speed.
We showed that the proposed methods improve the two-pass direct speech-to-spectrogram model as well, confirming their versatility.
Still, {\unity} achieved {\speedupoverspec}$\times$ decoding speed-up over the case.

\section{Limitation}
Since two-pass direct S2ST models require linguistic units as the target for the first-pass decoder, they cannot be used when the target language is unwritten.
Compared to cascaded S2ST systems, direct S2ST systems require more data preparation steps, including training a HuBERT model, synthesizing target speech with a TTS model, extracting discrete units with the HuBERT model, and training a unit-based vocoder, etc.
Moreover, the target audio quality of direct speech-to-unit systems relies on the quality of discrete units generated by self-supervised discrete models.
It further depends on the availability of speech data to train HuBERT models for the target languages.

Because S2ST systems could generate speech that does not necessarily represent the source speech's content, there is a potential risk of conveying wrong information.

\section*{Acknowledgement}
We would like to thank Justine Kao and Carleigh Wood for the help on human evaluation.

\bibliography{references}

\begin{thebibliography}{102}
\expandafter\ifx\csname natexlab\endcsname\relax\def\natexlab#1{#1}\fi

\bibitem[{Anastasopoulos et~al.(2021)Anastasopoulos, Bojar, Bremerman, Cattoni,
  Elbayad, Federico, Ma, Nakamura, Negri, Niehues, Pino, Salesky, St{\"u}ker,
  Sudoh, Turchi, Waibel, Wang, and Wiesner}]{anastasopoulos-etal-2021-findings}
Antonios Anastasopoulos, Ond{\v{r}}ej Bojar, Jacob Bremerman, Roldano Cattoni,
  Maha Elbayad, Marcello Federico, Xutai Ma, Satoshi Nakamura, Matteo Negri,
  Jan Niehues, Juan Pino, Elizabeth Salesky, Sebastian St{\"u}ker, Katsuhito
  Sudoh, Marco Turchi, Alexander Waibel, Changhan Wang, and Matthew Wiesner.
  2021.
\newblock {FINDINGS} {OF} {THE} {IWSLT} 2021 {EVALUATION} {CAMPAIGN}.
\newblock In \emph{Proceedings of IWSLT}, pages 1--29.

\bibitem[{Anastasopoulos and Chiang(2018)}]{anastasopoulos-chiang-2018-tied}
Antonios Anastasopoulos and David Chiang. 2018.
\newblock Tied multitask learning for neural speech translation.
\newblock In \emph{Proceedings of NAACL-HLT}, pages 82--91.

\bibitem[{Ansari et~al.(2020)Ansari, Axelrod, Bach, Bojar, Cattoni, Dalvi,
  Durrani, Federico, Federmann, Gu, Huang, Knight, Ma, Nagesh, Negri, Niehues,
  Pino, Salesky, Shi, St{\"u}ker, Turchi, Waibel, and
  Wang}]{ansari-etal-2020-findings}
Ebrahim Ansari, Amittai Axelrod, Nguyen Bach, Ond{\v{r}}ej Bojar, Roldano
  Cattoni, Fahim Dalvi, Nadir Durrani, Marcello Federico, Christian Federmann,
  Jiatao Gu, Fei Huang, Kevin Knight, Xutai Ma, Ajay Nagesh, Matteo Negri, Jan
  Niehues, Juan Pino, Elizabeth Salesky, Xing Shi, Sebastian St{\"u}ker, Marco
  Turchi, Alexander Waibel, and Changhan Wang. 2020.
\newblock {FINDINGS} {OF} {THE} {IWSLT} 2020 {EVALUATION} {CAMPAIGN}.
\newblock In \emph{Proceedings of IWSLT}, pages 1--34.

\bibitem[{Ardila et~al.(2020)Ardila, Branson, Davis, Kohler, Meyer, Henretty,
  Morais, Saunders, Tyers, and Weber}]{ardila-etal-2020-common}
Rosana Ardila, Megan Branson, Kelly Davis, Michael Kohler, Josh Meyer, Michael
  Henretty, Reuben Morais, Lindsay Saunders, Francis Tyers, and Gregor Weber.
  2020.
\newblock {Common Voice}: {A} massively-multilingual speech corpus.
\newblock In \emph{Proceedings of LREC}, pages 4218--4222.

\bibitem[{Arivazhagan et~al.(2019)Arivazhagan, Bapna, Firat, Lepikhin, Johnson,
  Krikun, Chen, Cao, Foster, Cherry et~al.}]{arivazhagan2019massively}
Naveen Arivazhagan, Ankur Bapna, Orhan Firat, Dmitry Lepikhin, Melvin Johnson,
  Maxim Krikun, Mia~Xu Chen, Yuan Cao, George Foster, Colin Cherry, et~al.
  2019.
\newblock Massively multilingual neural machine translation in the wild:
  {F}indings and challenges.
\newblock \emph{arXiv preprint arXiv:1907.05019}.

\bibitem[{Baevski et~al.(2020)Baevski, Zhou, Mohamed, and
  Auli}]{baevski2020wav2vec}
Alexei Baevski, Yuhao Zhou, Abdelrahman Mohamed, and Michael Auli. 2020.
\newblock wav2vec 2.0: {A} framework for self-supervised learning of speech
  representations.
\newblock In \emph{Proceedings of NeurIPS}, volume~33, pages 12449--12460.

\bibitem[{Bahdanau et~al.(2015)Bahdanau, Cho, and Bengio}]{bahdanau2014neural}
Dzmitry Bahdanau, Kyunghyun Cho, and Yoshua Bengio. 2015.
\newblock Neural machine translation by jointly learning to align and
  translate.
\newblock In \emph{Proceedings of ICLR}.

\bibitem[{Bahdanau et~al.(2016)Bahdanau, Chorowski, Serdyuk, Brakel, and
  Bengio}]{bahdanau2016end}
Dzmitry Bahdanau, Jan Chorowski, Dmitriy Serdyuk, Philemon Brakel, and Yoshua
  Bengio. 2016.
\newblock End-to-end attention-based large vocabulary speech recognition.
\newblock In \emph{Proceedings of ICASSP}, pages 4945--4949.

\bibitem[{Bapna et~al.(2022)Bapna, Cherry, Zhang, Jia, Johnson, Cheng, Khanuja,
  Riesa, and Conneau}]{bapna2022mslam}
Ankur Bapna, Colin Cherry, Yu~Zhang, Ye~Jia, Melvin Johnson, Yong Cheng, Simran
  Khanuja, Jason Riesa, and Alexis Conneau. 2022.
\newblock {mSLAM}: {M}assively multilingual joint pre-training for speech and
  text.
\newblock \emph{arXiv preprint arXiv:2202.01374}.

\bibitem[{B{\'e}rard et~al.(2018)B{\'e}rard, Besacier, Kocabiyikoglu, and
  Pietquin}]{berard2018end}
Alexandre B{\'e}rard, Laurent Besacier, Ali~Can Kocabiyikoglu, and Olivier
  Pietquin. 2018.
\newblock End-to-end automatic speech translation of audiobooks.
\newblock In \emph{Proceedings of ICASSP}, pages 6224--6228. IEEE.

\bibitem[{B{\'e}rard et~al.(2016)B{\'e}rard, Pietquin, Servan, and
  Besacier}]{listen_and_translate}
Alexandre B{\'e}rard, Olivier Pietquin, Christophe Servan, and Laurent
  Besacier. 2016.
\newblock Listen and translate: {A} proof of concept for end-to-end
  speech-to-text translation.
\newblock In \emph{Proceedings of NIPS 2016 End-to-end Learning for Speech and
  Audio Processing Workshop}.

\bibitem[{Chan et~al.(2021)Chan, Park, Lee, Zhang, Le, and
  Norouzi}]{chan2021speechstew}
William Chan, Daniel Park, Chris Lee, Yu~Zhang, Quoc Le, and Mohammad Norouzi.
  2021.
\newblock Speechstew: {S}imply mix all available speech recognition data to
  train one large neural network.
\newblock \emph{arXiv preprint arXiv:2104.02133}.

\bibitem[{Cherry et~al.(2018)Cherry, Foster, Bapna, Firat, and
  Macherey}]{cherry-etal-2018-revisiting}
Colin Cherry, George Foster, Ankur Bapna, Orhan Firat, and Wolfgang Macherey.
  2018.
\newblock Revisiting character-based neural machine translation with capacity
  and compression.
\newblock In \emph{Proceedings of EMNLP}, pages 4295--4305.

\bibitem[{Cho et~al.(2014)Cho, Van~Merri{\"e}nboer, Gulcehre, Bahdanau,
  Bougares, Schwenk, and Bengio}]{cho2014learning}
Kyunghyun Cho, Bart Van~Merri{\"e}nboer, Caglar Gulcehre, Dzmitry Bahdanau,
  Fethi Bougares, Holger Schwenk, and Yoshua Bengio. 2014.
\newblock Learning phrase representations using {RNN} encoder-decoder for
  statistical machine translation.
\newblock \emph{arXiv preprint arXiv:1406.1078}.

\bibitem[{Chung et~al.(2021)Chung, Zhang, Han, Chiu, Qin, Pang, and
  Wu}]{chung2021w2v}
Yu-An Chung, Yu~Zhang, Wei Han, Chung-Cheng Chiu, James Qin, Ruoming Pang, and
  Yonghui Wu. 2021.
\newblock {w2v-BERT}: {C}ombining contrastive learning and masked language
  modeling for self-supervised speech pre-training.
\newblock In \emph{Proceedings of ASRU}.

\bibitem[{Conneau et~al.(2020)Conneau, Khandelwal, Goyal, Chaudhary, Wenzek,
  Guzm{\'a}n, Grave, Ott, Zettlemoyer, and Stoyanov}]{conneau2020unsupervised}
Alexis Conneau, Kartikay Khandelwal, Naman Goyal, Vishrav Chaudhary, Guillaume
  Wenzek, Francisco Guzm{\'a}n, {\'E}douard Grave, Myle Ott, Luke Zettlemoyer,
  and Veselin Stoyanov. 2020.
\newblock Unsupervised cross-lingual representation learning at scale.
\newblock In \emph{Proceedings of ACL}, pages 8440--8451.

\bibitem[{Dalmia et~al.(2021)Dalmia, Yan, Raunak, Metze, and
  Watanabe}]{dalmia-etal-2021-searchable}
Siddharth Dalmia, Brian Yan, Vikas Raunak, Florian Metze, and Shinji Watanabe.
  2021.
\newblock Searchable hidden intermediates for end-to-end models of decomposable
  sequence tasks.
\newblock In \emph{Proceedings of NAACL-HLT}, pages 1882--1896.

\bibitem[{Di~Gangi et~al.(2019)Di~Gangi, Cattoni, Bentivogli, Negri, and
  Turchi}]{di-gangi-etal-2019-must}
Mattia~A. Di~Gangi, Roldano Cattoni, Luisa Bentivogli, Matteo Negri, and Marco
  Turchi. 2019.
\newblock {M}u{ST}-{C}: a {M}ultilingual {S}peech {T}ranslation {C}orpus.
\newblock In \emph{Proceedings of NAACL-HLT}, pages 2012--2017.

\bibitem[{Dong et~al.(2022)Dong, Yue, Ko, Wang, Bai, and
  Zhang}]{dong2022leveraging}
Qianqian Dong, Fengpeng Yue, Tom Ko, Mingxuan Wang, Qibing Bai, and Yu~Zhang.
  2022.
\newblock Leveraging pseudo-labeled data to improve direct speech-to-speech
  translation.
\newblock \emph{arXiv preprint arXiv:2205.08993}.

\bibitem[{Elizabeth et~al.(2021)Elizabeth, Matthew, Jacob, Cattoni, Negri,
  Turchi, Oard, and Matt}]{elizabeth2021multilingual}
Salesky Elizabeth, Wiesner Matthew, Bremerman Jacob, Roldano Cattoni, Matteo
  Negri, Marco Turchi, Douglas~W Oard, and Post Matt. 2021.
\newblock The multilingual {TEDx} corpus for speech recognition and
  translation.
\newblock In \emph{Proceedings of Interspeech}, pages 3655--3659.

\bibitem[{Gales et~al.(2014)Gales, Knill, Ragni, and Rath}]{gales2014speech}
Mark~JF Gales, Kate~M Knill, Anton Ragni, and Shakti~P Rath. 2014.
\newblock Speech recognition and keyword spotting for low-resource languages:
  {B}abel project research at {CUED}.
\newblock In \emph{Proceedings of SLTU}, pages 16--23.

\bibitem[{Gowda and May(2020)}]{gowda-may-2020-finding}
Thamme Gowda and Jonathan May. 2020.
\newblock Finding the optimal vocabulary size for neural machine translation.
\newblock In \emph{Findings of EMNLP}, pages 3955--3964.

\bibitem[{Graves et~al.(2006)Graves, Fern{\'a}ndez, Gomez, and
  Schmidhuber}]{ctc_graves}
Alex Graves, Santiago Fern{\'a}ndez, Faustino Gomez, and J{\"u}rgen
  Schmidhuber. 2006.
\newblock Connectionist temporal classification: {L}abelling unsegmented
  sequence data with recurrent neural networks.
\newblock In \emph{Proceedings of ICML}, pages 369--376.

\bibitem[{Gulati et~al.(2020)Gulati, Qin, Chiu, Parmar, Zhang, Yu, Han, Wang,
  Zhang, Wu, and Pang}]{gulati2020}
Anmol Gulati, James Qin, Chung-Cheng Chiu, Niki Parmar, Yu~Zhang, Jiahui Yu,
  Wei Han, Shibo Wang, Zhengdong Zhang, Yonghui Wu, and Ruoming Pang. 2020.
\newblock Conformer: {C}onvolution-augmented {Transformer} for speech
  recognition.
\newblock In \emph{Proceedings of Interspeech}, pages 5036--5040.

\bibitem[{Harper et~al.()}]{babel}
Mary Harper et~al.
\newblock {IARPA Babel Program}.
\newblock \url{https://www.iarpa.gov/research-programs/babel}.
\newblock [Online].

\bibitem[{Hochreiter and Schmidhuber(1997)}]{lstm}
Sepp Hochreiter and J{\"u}rgen Schmidhuber. 1997.
\newblock Long short-term memory.
\newblock \emph{Neural Computation}, 9(8):1735--1780.

\bibitem[{Hsu et~al.(2021)Hsu, Bolte, Tsai, Lakhotia, Salakhutdinov, and
  Mohamed}]{hsu2021hubert}
Wei-Ning Hsu, Benjamin Bolte, Yao-Hung~Hubert Tsai, Kushal Lakhotia, Ruslan
  Salakhutdinov, and Abdelrahman Mohamed. 2021.
\newblock {HuBERT}: {S}elf-supervised speech representation learning by masked
  prediction of hidden units.
\newblock \emph{IEEE/ACM Transactions on Audio, Speech, and Language
  Processing}, 29:3451--3460.

\bibitem[{Hu et~al.(2020)Hu, Sainath, Pang, and
  Prabhavalkar}]{hu2020deliberation}
Ke~Hu, Tara~N Sainath, Ruoming Pang, and Rohit Prabhavalkar. 2020.
\newblock Deliberation model based two-pass end-to-end speech recognition.
\newblock In \emph{Proceedings of ICASSP}, pages 7799--7803.

\bibitem[{Huang et~al.(2022)Huang, Zhao, Liu, Liu, Ren, Zhang, and
  He}]{huang2022transpeech}
Rongjie Huang, Zhou Zhao, Jinglin Liu, Huadai Liu, Yi~Ren, Lichao Zhang, and
  Jinzheng He. 2022.
\newblock {TranSpeech}: {S}peech-to-speech translation with bilateral
  perturbation.
\newblock \emph{arXiv preprint arXiv:2205.12523}.

\bibitem[{Inaguma et~al.(2021{\natexlab{a}})Inaguma, Dalmia, Yan, and
  Watanabe}]{inaguma2021fast}
Hirofumi Inaguma, Siddharth Dalmia, Brian Yan, and Shinji Watanabe.
  2021{\natexlab{a}}.
\newblock {Fast-MD}: {F}ast multi-decoder end-to-end speech translation with
  non-autoregressive hidden intermediates.
\newblock In \emph{Proceedings of ASRU}, pages 922--929.

\bibitem[{Inaguma et~al.(2021{\natexlab{b}})Inaguma, Kawahara, and
  Watanabe}]{inaguma-etal-2021-source}
Hirofumi Inaguma, Tatsuya Kawahara, and Shinji Watanabe. 2021{\natexlab{b}}.
\newblock Source and target bidirectional knowledge distillation for end-to-end
  speech translation.
\newblock In \emph{Proceedings of NAACL-HLT}, pages 1872--1881.

\bibitem[{Iranzo-S{\'a}nchez et~al.(2020)Iranzo-S{\'a}nchez, Silvestre-Cerda,
  Jorge, Rosell{\'o}, Gim{\'e}nez, Sanchis, Civera, and
  Juan}]{iranzo2020europarl}
Javier Iranzo-S{\'a}nchez, Joan~Albert Silvestre-Cerda, Javier Jorge, Nahuel
  Rosell{\'o}, Adria Gim{\'e}nez, Albert Sanchis, Jorge Civera, and Alfons
  Juan. 2020.
\newblock {Europarl-ST}: {A} multilingual corpus for speech translation of
  parliamentary debates.
\newblock In \emph{Proceedings of ICASSP}, pages 8229--8233.

\bibitem[{Ito and Johnson(2017)}]{ljspeech17}
Keith Ito and Linda Johnson. 2017.
\newblock The lj speech dataset.
\newblock \url{https://keithito.com/LJ-Speech-Dataset/}.

\bibitem[{Jia et~al.(2022{\natexlab{a}})Jia, Ding, Bapna, Cherry, Zhang,
  Conneau, and Morioka}]{jia2022leveraging}
Ye~Jia, Yifan Ding, Ankur Bapna, Colin Cherry, Yu~Zhang, Alexis Conneau, and
  Nobuyuki Morioka. 2022{\natexlab{a}}.
\newblock Leveraging unsupervised and weakly-supervised data to improve direct
  speech-to-speech translation.
\newblock In \emph{Proceedings of Interspeech}, pages 1721--1725.

\bibitem[{Jia et~al.(2019{\natexlab{a}})Jia, Johnson, Macherey, Weiss, Cao,
  Chiu, Ari, Laurenzo, and Wu}]{jia2019leveraging}
Ye~Jia, Melvin Johnson, Wolfgang Macherey, Ron~J Weiss, Yuan Cao, Chung-Cheng
  Chiu, Naveen Ari, Stella Laurenzo, and Yonghui Wu. 2019{\natexlab{a}}.
\newblock Leveraging weakly supervised data to improve end-to-end
  speech-to-text translation.
\newblock In \emph{Proceedings of ICASSP}, pages 7180--7184.

\bibitem[{Jia et~al.(2022{\natexlab{b}})Jia, Ramanovich, Remez, and
  Pomerantz}]{jia2022translatotron}
Ye~Jia, Michelle~Tadmor Ramanovich, Tal Remez, and Roi Pomerantz.
  2022{\natexlab{b}}.
\newblock Translatotron 2: {H}igh-quality direct speech-to-speech translation
  with voice preservation.
\newblock In \emph{Proceedings of ICML}.

\bibitem[{Jia et~al.(2022{\natexlab{c}})Jia, Tadmor~Ramanovich, Wang, and
  Zen}]{jia-etal-2022-cvss}
Ye~Jia, Michelle Tadmor~Ramanovich, Quan Wang, and Heiga Zen.
  2022{\natexlab{c}}.
\newblock {CVSS} corpus and massively multilingual speech-to-speech
  translation.
\newblock In \emph{Proceedings of LREC}, pages 6691--6703.

\bibitem[{Jia et~al.(2019{\natexlab{b}})Jia, Weiss, Biadsy, Macherey, Johnson,
  Chen, and Wu}]{jia2019direct}
Ye~Jia, Ron~J Weiss, Fadi Biadsy, Wolfgang Macherey, Melvin Johnson, Zhifeng
  Chen, and Yonghui Wu. 2019{\natexlab{b}}.
\newblock Direct speech-to-speech translation with a sequence-to-sequence
  model.
\newblock In \emph{Proceedings of Interspeech}, pages 1123--1127.

\bibitem[{Kahn et~al.(2020)Kahn, Rivi{\`e}re, Zheng, Kharitonov, Xu,
  Mazar{\'e}, Karadayi, Liptchinsky, Collobert, Fuegen et~al.}]{kahn2020libri}
Jacob Kahn, Morgane Rivi{\`e}re, Weiyi Zheng, Evgeny Kharitonov, Qiantong Xu,
  Pierre-Emmanuel Mazar{\'e}, Julien Karadayi, Vitaliy Liptchinsky, Ronan
  Collobert, Christian Fuegen, et~al. 2020.
\newblock Libri-{L}ight: {A} benchmark for asr with limited or no supervision.
\newblock In \emph{Proceedings of ICASSP}, pages 7669--7673.

\bibitem[{Kano et~al.(2021)Kano, Sakti, and Nakamura}]{kano2021transformer}
Takatomo Kano, Sakriani Sakti, and Satoshi Nakamura. 2021.
\newblock Transformer-based direct speech-to-speech translation with
  transcoder.
\newblock In \emph{Proceedings of SLT}, pages 958--965.

\bibitem[{Kasai et~al.(2021)Kasai, Pappas, Peng, Cross, and
  Smith}]{kasai2021deep}
Jungo Kasai, Nikolaos Pappas, Hao Peng, James Cross, and Noah Smith. 2021.
\newblock Deep encoder, shallow decoder: {R}eevaluating non-autoregressive
  machine translation.
\newblock In \emph{Proceedings of ICLR}.

\bibitem[{Koehn(2005)}]{koehn-2005-europarl}
Philipp Koehn. 2005.
\newblock {E}uroparl: {A} parallel corpus for statistical machine translation.
\newblock In \emph{Proceedings of Machine Translation Summit X: Papers}, pages
  79--86.

\bibitem[{Kong et~al.(2020)Kong, Kim, and Bae}]{kong2020hifi}
Jungil Kong, Jaehyeon Kim, and Jaekyoung Bae. 2020.
\newblock {HiFi-GAN}: {G}enerative adversarial networks for efficient and high
  fidelity speech synthesis.
\newblock In \emph{Proceedings of NeurIPS}, volume~33, pages 17022--17033.

\bibitem[{Kudo(2018)}]{kudo-2018-subword}
Taku Kudo. 2018.
\newblock Subword regularization: {I}mproving neural network translation models
  with multiple subword candidates.
\newblock In \emph{Proceedings of ACL}, pages 66--75.

\bibitem[{Kudo and Richardson(2018)}]{kudo-richardson-2018-sentencepiece}
Taku Kudo and John Richardson. 2018.
\newblock {S}entence{P}iece: A simple and language independent subword
  tokenizer and detokenizer for neural text processing.
\newblock In \emph{Proceedings of EMNLP: System Demonstrations}, pages 66--71.

\bibitem[{Lavie et~al.(1997)Lavie, Waibel, Levin, Finke, Gates, Gavalda,
  Zeppenfeld, and Zhan}]{lavie1997janus}
Alon Lavie, Alex Waibel, Lori Levin, Michael Finke, Donna Gates, Marsal
  Gavalda, Torsten Zeppenfeld, and Puming Zhan. 1997.
\newblock {JANUS-III}: {S}peech-to-speech translation in multiple languages.
\newblock In \emph{Proceedings of ICASSP}, pages 99--102.

\bibitem[{Lee et~al.(2022{\natexlab{a}})Lee, Chen, Wang, Gu, Ma, Polyak, Adi,
  He, Tang, Pino et~al.}]{lee2021direct}
Ann Lee, Peng-Jen Chen, Changhan Wang, Jiatao Gu, Xutai Ma, Adam Polyak, Yossi
  Adi, Qing He, Yun Tang, Juan Pino, et~al. 2022{\natexlab{a}}.
\newblock Direct speech-to-speech translation with discrete units.
\newblock In \emph{Proceedings of ACL}, pages 3327--3339.

\bibitem[{Lee et~al.(2022{\natexlab{b}})Lee, Gong, Duquenne, Schwenk, Chen,
  Wang, Popuri, Pino, Gu, and Hsu}]{lee2021textless}
Ann Lee, Hongyu Gong, Paul-Ambroise Duquenne, Holger Schwenk, Peng-Jen Chen,
  Changhan Wang, Sravya Popuri, Juan Pino, Jiatao Gu, and Wei-Ning Hsu.
  2022{\natexlab{b}}.
\newblock Textless speech-to-speech translation on real data.
\newblock In \emph{Proceedings of NAACL-HLT}, pages 860--872.

\bibitem[{Li et~al.(2019)Li, Liu, Liu, Zhao, and Liu}]{li2019neural}
Naihan Li, Shujie Liu, Yanqing Liu, Sheng Zhao, and Ming Liu. 2019.
\newblock Neural speech synthesis with {Transformer} network.
\newblock In \emph{Proceedings of AAAI}, volume~33, pages 6706--6713.

\bibitem[{Li et~al.(2021)Li, Wang, Tang, Tran, Tang, Pino, Baevski, Conneau,
  and Auli}]{li-etal-2021-multilingual}
Xian Li, Changhan Wang, Yun Tang, Chau Tran, Yuqing Tang, Juan Pino, Alexei
  Baevski, Alexis Conneau, and Michael Auli. 2021.
\newblock Multilingual speech translation from efficient finetuning of
  pretrained models.
\newblock In \emph{Proceedings of ACL}, pages 827--838.

\bibitem[{Li et~al.(2022)Li, Jia, and Chiu}]{li2022textless}
Xinjian Li, Ye~Jia, and Chung-Cheng Chiu. 2022.
\newblock Textless direct speech-to-speech translation with discrete speech
  representation.
\newblock \emph{arXiv preprint arXiv:2211.00115}.

\bibitem[{Licht et~al.(2022)Licht, Gao, Lam, Guzman, Diab, and
  Koehn}]{licht-etal-2022-consistent}
Daniel Licht, Cynthia Gao, Janice Lam, Francisco Guzman, Mona Diab, and Philipp
  Koehn. 2022.
\newblock Consistent human evaluation of machine translation across language
  pairs.
\newblock In \emph{Proceedings of AMTA}, pages 309--321.

\bibitem[{Lison et~al.(2018)Lison, Tiedemann, and
  Kouylekov}]{lison-etal-2018-opensubtitles2018}
Pierre Lison, J{\"o}rg Tiedemann, and Milen Kouylekov. 2018.
\newblock {O}pen{S}ubtitles2018: Statistical rescoring of sentence alignments
  in large, noisy parallel corpora.
\newblock In \emph{{Proceedings of LREC}}.

\bibitem[{Liu et~al.(2020)Liu, Gu, Goyal, Li, Edunov, Ghazvininejad, Lewis, and
  Zettlemoyer}]{liu2020multilingual}
Yinhan Liu, Jiatao Gu, Naman Goyal, Xian Li, Sergey Edunov, Marjan
  Ghazvininejad, Mike Lewis, and Luke Zettlemoyer. 2020.
\newblock Multilingual denoising pre-training for neural machine translation.
\newblock \emph{Transactions of the Association for Computational Linguistics},
  8:726--742.

\bibitem[{Liu et~al.(2019)Liu, Xiong, He, Zhang, Wu, Wang, and
  Zong}]{liu2019end}
Yuchen Liu, Hao Xiong, Zhongjun He, Jiajun Zhang, Hua Wu, Haifeng Wang, and
  Chengqing Zong. 2019.
\newblock End-to-end speech translation with knowledge distillation.
\newblock In \emph{Proceedings of Interspeech}, pages 1128--1132.

\bibitem[{Micikevicius et~al.(2018)Micikevicius, Narang, Alben, Diamos, Elsen,
  Garcia, Ginsburg, Houston, Kuchaiev, Venkatesh, and
  Wu}]{micikevicius2018mixed}
Paulius Micikevicius, Sharan Narang, Jonah Alben, Gregory Diamos, Erich Elsen,
  David Garcia, Boris Ginsburg, Michael Houston, Oleksii Kuchaiev, Ganesh
  Venkatesh, and Hao Wu. 2018.
\newblock Mixed precision training.
\newblock In \emph{Proceedings of ICLR}.

\bibitem[{Mohri et~al.(2002)Mohri, Pereira, and Riley}]{mohri2002weighted}
Mehryar Mohri, Fernando Pereira, and Michael Riley. 2002.
\newblock Weighted finite-state transducers in speech recognition.
\newblock \emph{Computer Speech \& Language}, 16(1):69--88.

\bibitem[{Nakamura et~al.(2006)Nakamura, Markov, Nakaiwa, Kikui, Kawai,
  Jitsuhiro, Zhang, Yamamoto, Sumita, and Yamamoto}]{nakamura2006atr}
Satoshi Nakamura, Konstantin Markov, Hiromi Nakaiwa, Gen-ichiro Kikui, Hisashi
  Kawai, Takatoshi Jitsuhiro, J-S Zhang, Hirofumi Yamamoto, Eiichiro Sumita,
  and Seiichi Yamamoto. 2006.
\newblock The {ATR} multilingual speech-to-speech translation system.
\newblock \emph{IEEE/ACM Transactions on Audio, Speech, and Language
  Processing}, 14(2):365--376.

\bibitem[{Ott et~al.(2019)Ott, Edunov, Baevski, Fan, Gross, Ng, Grangier, and
  Auli}]{ott2019fairseq}
Myle Ott, Sergey Edunov, Alexei Baevski, Angela Fan, Sam Gross, Nathan Ng,
  David Grangier, and Michael Auli. 2019.
\newblock fairseq: {A} fast, extensible toolkit for sequence modeling.
\newblock \emph{arXiv preprint arXiv:1904.01038}.

\bibitem[{Panayotov et~al.(2015)Panayotov, Chen, Povey, and
  Khudanpur}]{librispeech}
Vassil Panayotov, Guoguo Chen, Daniel Povey, and Sanjeev Khudanpur. 2015.
\newblock Librispeech: {A}n {ASR} corpus based on public domain audio books.
\newblock In \emph{Proceedings of ICASSP}, pages 5206--5210.

\bibitem[{Papineni et~al.(2002)Papineni, Roukos, Ward, and
  Zhu}]{papineni-etal-2002-bleu}
Kishore Papineni, Salim Roukos, Todd Ward, and Wei-Jing Zhu. 2002.
\newblock {B}leu: a method for automatic evaluation of machine translation.
\newblock In \emph{Proceedings of ACL}, pages 311--318.

\bibitem[{Park and Mulc(2019)}]{park2019css10}
Kyubyong Park and Thomas Mulc. 2019.
\newblock {CSS10}: {A} collection of single speaker speech datasets for 10
  languages.
\newblock In \emph{Proceedings of Interspeech}, pages 1566--1570.

\bibitem[{Pino et~al.(2020)Pino, Xu, Ma, Dousti, and Tang}]{pino2020self}
Juan Pino, Qiantong Xu, Xutai Ma, Mohammad~Javad Dousti, and Yun Tang. 2020.
\newblock Self-training for end-to-end speech translation.
\newblock In \emph{Proceedings of Interspeech}, pages 1476--1480.

\bibitem[{Polyak et~al.(2021)Polyak, Adi, Copet, Kharitonov, Lakhotia, Hsu,
  Mohamed, and Dupoux}]{polyak21_interspeech}
Adam Polyak, Yossi Adi, Jade Copet, Eugene Kharitonov, Kushal Lakhotia,
  Wei-Ning Hsu, Abdelrahman Mohamed, and Emmanuel Dupoux. 2021.
\newblock Speech resynthesis from discrete disentangled self-supervised
  representations.
\newblock In \emph{Proceedings of Interspeech}, pages 3615--3619.

\bibitem[{Popuri et~al.(2022)Popuri, Chen, Wang, Pino, Adi, Gu, Hsu, and
  Lee}]{popuri2022enhanced}
Sravya Popuri, Peng-Jen Chen, Changhan Wang, Juan Pino, Yossi Adi, Jiatao Gu,
  Wei-Ning Hsu, and Ann Lee. 2022.
\newblock Enhanced direct speech-to-speech translation using self-supervised
  pre-training and data augmentation.
\newblock In \emph{Proceedings of Interspeech}, pages 5195--5199.

\bibitem[{Post(2018)}]{post-2018-call}
Matt Post. 2018.
\newblock A call for clarity in reporting {BLEU} scores.
\newblock In \emph{Proceedings of the Third Conference on Machine Translation:
  Research Papers}, pages 186--191.

\bibitem[{Post et~al.(2013)Post, Kumar, Lopez, Karakos, Callison-Burch, and
  Khudanpur}]{fisher_callhome}
Matt Post, Gaurav Kumar, Adam Lopez, Damianos Karakos, Chris Callison-Burch,
  and Sanjeev Khudanpur. 2013.
\newblock Improved speech-to-text translation with the {Fisher} and {Callhome
  Spanish--English} speech translation corpus.
\newblock In \emph{Proceedings of IWSLT}.

\bibitem[{Pratap et~al.(2020)Pratap, Xu, Sriram, Synnaeve, and
  Collobert}]{pratap2020mls}
Vineel Pratap, Qiantong Xu, Anuroop Sriram, Gabriel Synnaeve, and Ronan
  Collobert. 2020.
\newblock {MLS}: {A} large-scale multilingual dataset for speech research.
\newblock In \emph{Proceedings of Interspeech}, pages 2757--2761.

\bibitem[{Reimers and Gurevych(2020)}]{reimers-gurevych-2020-making}
Nils Reimers and Iryna Gurevych. 2020.
\newblock Making monolingual sentence embeddings multilingual using knowledge
  distillation.
\newblock In \emph{Proceedings of EMNLP}, pages 4512--4525.

\bibitem[{Rousseau et~al.(2012)Rousseau, Del{\'e}glise, and
  Est{\`e}ve}]{tedlium}
Anthony Rousseau, Paul Del{\'e}glise, and Yannick Est{\`e}ve. 2012.
\newblock {TED}-{LIUM}: {A}n automatic speech recognition dedicated corpus.
\newblock In \emph{Proceedings of LREC}, pages 125--129.

\bibitem[{Sainath et~al.(2019)Sainath, Pang, Rybach, He, Prabhavalkar, Li,
  Visontai, Liang, Strohman, Wu et~al.}]{sainath2019two}
Tara~N Sainath, Ruoming Pang, David Rybach, Yanzhang He, Rohit Prabhavalkar,
  Wei Li, Mirk{\'o} Visontai, Qiao Liang, Trevor Strohman, Yonghui Wu, et~al.
  2019.
\newblock Two-pass end-to-end speech recognition.
\newblock In \emph{Proceedings of Interspeech}, pages 2773--2777.

\bibitem[{Salesky et~al.(2021)Salesky, M{\"a}der, and
  Klinger}]{salesky2021assessing}
Elizabeth Salesky, Julian M{\"a}der, and Severin Klinger. 2021.
\newblock Assessing evaluation metrics for speech-to-speech translation.
\newblock In \emph{Proceedings of ASRU}, pages 733--740.

\bibitem[{Schwenk et~al.(2021)Schwenk, Wenzek, Edunov, Grave, Joulin, and
  Fan}]{schwenk-etal-2021-ccmatrix}
Holger Schwenk, Guillaume Wenzek, Sergey Edunov, Edouard Grave, Armand Joulin,
  and Angela Fan. 2021.
\newblock {CCM}atrix: {M}ining billions of high-quality parallel sentences on
  the web.
\newblock In \emph{Proceedings of ACL}, pages 6490--6500.

\bibitem[{Shen et~al.(2020)Shen, Jia, Chrzanowski, Zhang, Elias, Zen, and
  Wu}]{shen2020non}
Jonathan Shen, Ye~Jia, Mike Chrzanowski, Yu~Zhang, Isaac Elias, Heiga Zen, and
  Yonghui Wu. 2020.
\newblock {Non-Attentive Tacotron}: {R}obust and controllable neural {TTS}
  synthesis including unsupervised duration modeling.
\newblock \emph{arXiv preprint arXiv:2010.04301}.

\bibitem[{Skadi{\c{n}}{\v{s}} et~al.(2014)Skadi{\c{n}}{\v{s}}, Tiedemann,
  Rozis, and Deksne}]{skadins-etal-2014-billions}
Raivis Skadi{\c{n}}{\v{s}}, J{\"o}rg Tiedemann, Roberts Rozis, and Daiga
  Deksne. 2014.
\newblock Billions of parallel words for free: Building and using the {EU}
  bookshop corpus.
\newblock In \emph{Proceedings of LREC}, pages 1850--1855.

\bibitem[{Sperber et~al.(2019)Sperber, Neubig, Niehues, and
  Waibel}]{sperber-etal-2019-attention}
Matthias Sperber, Graham Neubig, Jan Niehues, and Alex Waibel. 2019.
\newblock Attention-passing models for robust and data-efficient end-to-end
  speech translation.
\newblock \emph{Transactions of the Association for Computational Linguistics},
  7:313--325.

\bibitem[{Srivastava et~al.(2014)Srivastava, Hinton, Krizhevsky, Sutskever, and
  Salakhutdinov}]{srivastava2014dropout}
Nitish Srivastava, Geoffrey Hinton, Alex Krizhevsky, Ilya Sutskever, and Ruslan
  Salakhutdinov. 2014.
\newblock Dropout: a simple way to prevent neural networks from overfitting.
\newblock \emph{The journal of machine learning research}, 15(1):1929--1958.

\bibitem[{Sung et~al.(2019)Sung, Liu, Lee, and Lee}]{sung2019towards}
Tzu-Wei Sung, Jun-You Liu, Hung-yi Lee, and Lin-shan Lee. 2019.
\newblock Towards end-to-end speech-to-text translation with two-pass decoding.
\newblock In \emph{Proceedings of ICASSP}, pages 7175--7179.

\bibitem[{Sutskever et~al.(2014)Sutskever, Vinyals, and
  Le}]{sutskever2014sequence}
Ilya Sutskever, Oriol Vinyals, and Quoc~V Le. 2014.
\newblock Sequence to sequence learning with neural networks.
\newblock In \emph{Proceedings of NIPS}, volume~27.

\bibitem[{Szegedy et~al.(2016)Szegedy, Vanhoucke, Ioffe, Shlens, and
  Wojna}]{label_smoothing}
Christian Szegedy, Vincent Vanhoucke, Sergey Ioffe, Jon Shlens, and Zbigniew
  Wojna. 2016.
\newblock Rethinking the inception architecture for computer vision.
\newblock In \emph{Proceedings of CVPR}, pages 2818--2826.

\bibitem[{Tang et~al.(2022)Tang, Gong, Dong, Wang, Hsu, Gu, Baevski, Li,
  Mohamed, Auli, and Pino}]{tang-etal-2022-unified}
Yun Tang, Hongyu Gong, Ning Dong, Changhan Wang, Wei-Ning Hsu, Jiatao Gu,
  Alexei Baevski, Xian Li, Abdelrahman Mohamed, Michael Auli, and Juan Pino.
  2022.
\newblock Unified speech-text pre-training for speech translation and
  recognition.
\newblock In \emph{Proceedings of ACL}, pages 1488--1499.

\bibitem[{Tang et~al.(2021)Tang, Pino, Li, Wang, and
  Genzel}]{tang-etal-2021-improving}
Yun Tang, Juan Pino, Xian Li, Changhan Wang, and Dmitriy Genzel. 2021.
\newblock Improving speech translation by understanding and learning from the
  auxiliary text translation task.
\newblock In \emph{Proceedings of ACL}, pages 4252--4261.

\bibitem[{Tang et~al.(2020)Tang, Tran, Li, Chen, Goyal, Chaudhary, Gu, and
  Fan}]{tang2020multilingual}
Yuqing Tang, Chau Tran, Xian Li, Peng-Jen Chen, Naman Goyal, Vishrav Chaudhary,
  Jiatao Gu, and Angela Fan. 2020.
\newblock Multilingual translation with extensible multilingual pretraining and
  finetuning.
\newblock \emph{arXiv preprint arXiv:2008.00401}.

\bibitem[{Tjandra et~al.(2019)Tjandra, Sakti, and Nakamura}]{tjandra2019speech}
Andros Tjandra, Sakriani Sakti, and Satoshi Nakamura. 2019.
\newblock Speech-to-speech translation between untranscribed unknown languages.
\newblock In \emph{Proceedings of ASRU}, pages 593--600.

\bibitem[{Valk and Alum{\"a}e(2021)}]{valk2021voxlingua107}
J{\"o}rgen Valk and Tanel Alum{\"a}e. 2021.
\newblock {VoxLingua107}: a dataset for spoken language recognition.
\newblock In \emph{Proceedings of SLT}, pages 652--658.

\bibitem[{Van Den~Oord et~al.(2017)Van Den~Oord, Vinyals
  et~al.}]{van2017neural}
Aaron Van Den~Oord, Oriol Vinyals, et~al. 2017.
\newblock Neural discrete representation learning.
\newblock In \emph{Proceedings of NIPS}, volume~30.

\bibitem[{Vaswani et~al.(2017)Vaswani, Shazeer, Parmar, Uszkoreit, Jones,
  Gomez, Kaiser, and Polosukhin}]{vaswani2017attention}
Ashish Vaswani, Noam Shazeer, Niki Parmar, Jakob Uszkoreit, Llion Jones,
  Aidan~N Gomez, {\L}ukasz Kaiser, and Illia Polosukhin. 2017.
\newblock Attention is all you need.
\newblock In \emph{Proceedings of NIPS}, volume~30.

\bibitem[{Wahlster(2013)}]{wahlster2013verbmobil}
Wolfgang Wahlster. 2013.
\newblock \emph{Verbmobil: foundations of speech-to-speech translation}.
\newblock Springer Science \& Business Media.

\bibitem[{Wang et~al.(2022)Wang, Inaguma, Chen, Kulikov, Tang, Hsu, Auli, and
  Pino}]{wang2022simple}
Changhan Wang, Hirofumi Inaguma, Peng-Jen Chen, Ilia Kulikov, Yun Tang,
  Wei-Ning Hsu, Michael Auli, and Juan Pino. 2022.
\newblock Simple and effective unsupervised speech translation.
\newblock \emph{arXiv preprint arXiv:2210.10191}.

\bibitem[{Wang et~al.(2021{\natexlab{a}})Wang, Riviere, Lee, Wu, Talnikar,
  Haziza, Williamson, Pino, and Dupoux}]{wang2021voxpopuli}
Changhan Wang, Morgane Riviere, Ann Lee, Anne Wu, Chaitanya Talnikar, Daniel
  Haziza, Mary Williamson, Juan Pino, and Emmanuel Dupoux. 2021{\natexlab{a}}.
\newblock {VoxPopuli}: {A} large-scale multilingual speech corpus for
  representation learning, semi-supervised learning and interpretation.
\newblock In \emph{Proceedings of ACL}, pages 993--1003.

\bibitem[{Wang et~al.(2020)Wang, Tang, Ma, Wu, Okhonko, and
  Pino}]{wang2020fairseq}
Changhan Wang, Yun Tang, Xutai Ma, Anne Wu, Dmytro Okhonko, and Juan Pino.
  2020.
\newblock Fairseq {S2T}: {F}ast speech-to-text modeling with {Fairseq}.
\newblock In \emph{Proceedings of AACL: System Demonstrations}, pages 33--39.

\bibitem[{Wang et~al.(2021{\natexlab{b}})Wang, Wu, Gu, and
  Pino}]{wang2021covost}
Changhan Wang, Anne Wu, Jiatao Gu, and Juan Pino. 2021{\natexlab{b}}.
\newblock {CoVoST} 2 and massively multilingual speech translation.
\newblock In \emph{Proceedings of Interspeech}, pages 2247--2251.

\bibitem[{Wang et~al.(2021{\natexlab{c}})Wang, Wu, Pino, Baevski, Auli, and
  Conneau}]{wang21r_interspeech}
Changhan Wang, Anne Wu, Juan Pino, Alexei Baevski, Michael Auli, and Alexis
  Conneau. 2021{\natexlab{c}}.
\newblock Large-scale self- and semi-supervised learning for speech
  translation.
\newblock In \emph{Proceedings of Interspeech}, pages 2242--2246.

\bibitem[{Weiss et~al.(2017)Weiss, Chorowski, Jaitly, Wu, and
  Chen}]{weiss2017sequence}
Ron~J Weiss, Jan Chorowski, Navdeep Jaitly, Yonghui Wu, and Zhifeng Chen. 2017.
\newblock Sequence-to-sequence models can directly translate foreign speech.
\newblock In \emph{Proceedings of Interspeech}, pages 2625--2629.

\bibitem[{Wo{\l}k and Marasek(2014)}]{wolk2014building}
Krzysztof Wo{\l}k and Krzysztof Marasek. 2014.
\newblock Building subject-aligned comparable corpora and mining it for truly
  parallel sentence pairs.
\newblock \emph{Procedia Technology}, 18:126--132.

\bibitem[{Wu et~al.(2021)Wu, Li, Wang, Meng, Qin, Chen, Zhang, Liu
  et~al.}]{wu2021rdrop}
Lijun Wu, Juntao Li, Yue Wang, Qi~Meng, Tao Qin, Wei Chen, Min Zhang, Tie-Yan
  Liu, et~al. 2021.
\newblock {R-Drop}: {R}egularized dropout for neural networks.
\newblock In \emph{Proceedings off NeurIPS}, volume~34, pages 10890--10905.

\bibitem[{Xia et~al.(2017)Xia, Tian, Wu, Lin, Qin, Yu, and
  Liu}]{xia2017deliberation}
Yingce Xia, Fei Tian, Lijun Wu, Jianxin Lin, Tao Qin, Nenghai Yu, and Tie-Yan
  Liu. 2017.
\newblock Deliberation networks: {S}equence generation beyond one-pass
  decoding.
\newblock In \emph{Proceedings of NIPS}, volume~30.

\bibitem[{Yan et~al.(2022)Yan, Fernandes, Dalmia, Shi, Peng, Berrebbi, Wang,
  Neubig, and Watanabe}]{yan-etal-2022-cmus}
Brian Yan, Patrick Fernandes, Siddharth Dalmia, Jiatong Shi, Yifan Peng, Dan
  Berrebbi, Xinyi Wang, Graham Neubig, and Shinji Watanabe. 2022.
\newblock {CMU}{'}s {IWSLT} 2022 dialect speech translation system.
\newblock In \emph{Proceedings of IWSLT}, pages 298--307.

\bibitem[{Zhang et~al.(2021)Zhang, Tan, Ren, Qin, Zhang, and
  Liu}]{zhang2021uwspeech}
Chen Zhang, Xu~Tan, Yi~Ren, Tao Qin, Kejun Zhang, and Tie-Yan Liu. 2021.
\newblock Uwspeech: {S}peech to speech translation for unwritten languages.
\newblock In \emph{Proceedings of AAAI}, pages 14319--14327.

\bibitem[{Zhao et~al.(2019)Zhao, Sainath, Rybach, Rondon, Bhatia, Li, and
  Pang}]{zhao19d_interspeech}
Ding Zhao, Tara~N. Sainath, David Rybach, Pat Rondon, Deepti Bhatia, Bo~Li, and
  Ruoming Pang. 2019.
\newblock Shallow-fusion end-to-end contextual biasing.
\newblock In \emph{Proceedings of Interspeech}, pages 1418--1422.

\bibitem[{Zhu et~al.(2019)Zhu, Xia, Wu, He, Qin, Zhou, Li, and
  Liu}]{zhu2019incorporating}
Jinhua Zhu, Yingce Xia, Lijun Wu, Di~He, Tao Qin, Wengang Zhou, Houqiang Li,
  and Tieyan Liu. 2019.
\newblock Incorporating {BERT} into neural machine translation.
\newblock In \emph{Proceedings of ICLR}.

\bibitem[{Ziemski et~al.(2016)Ziemski, Junczys-Dowmunt, and
  Pouliquen}]{ziemski-etal-2016-united}
Micha{\l} Ziemski, Marcin Junczys-Dowmunt, and Bruno Pouliquen. 2016.
\newblock The {U}nited {N}ations parallel corpus v1.0.
\newblock In \emph{Proceedings of LREC}, pages 3530--3534.

\end{thebibliography}

\clearpage
\appendix
\section{Pseudo algorithm for two-pass beam search decoding}\label{appendix:decoding_algo}
Algorithm~\ref{algo:two_pass_beam} shows the two-pass beam serach decoding algorithm of {\unity} as discussed in \cref{ssec:search_algo}.
We first encode a source speech $\srcspeech$ with the speech encoder \texttt{SpeechEnc} and map it to the encoder output $\srcencout$.

The first-pass decoder takes $\srcencout$ as input and generates a text sequence.
$\textit{BeamSearch}_{1}$ is a first-pass beam search function that takes an empty hypothesis set $\omegafirst$ and returns the beam candidates.
We take the best text hypothesis $\hyptgttext$ and get the corresponding decoder output $\textdecout$ from a cache via the \textit{GetHiddenStateFromCache} function.
Next, the T2U encoder \texttt{T2UEnc} takes $\textdecout$ as input and maps it to the output $\ttuencout$.

The second-pass decoder takes $\srcencout$ and $\ttuencout$ as inputs and generates a discrete unit sequence.
$\textit{BeamSearch}_{2}$ is a second-pass beam search function that takes an empty hypothesis set $\omegasecond$ and returns the beam candidates.
We take the best discrete unit hypothesis $\hyptgtunit$.
Finally, the unit-based vocoder \texttt{UnitVocoder} converts $\hyptgtunit$ to the waveform $\hyptgtwave$.

\begin{algorithm}[t]
\caption{Two-pass beam search decoding}
\footnotesize
\begin{algorithmic}[1]
\Function{TwoPassBeamSearch}{$\srcspeech, \beamsizetext, \beamsizeunit$}
    \State $\srcencout \gets \texttt{SpeechEnc}(\srcspeech)$ \Comment{$\srcencout: (T', \dmodel)$}
    \State

    \State \texttt{// First-pass beam search}
    \State $\omegafirst \gets \{\}$
    \State $\omegafirst \gets \textit{BeamSearch}_{1}(\srcencout, \beamsizetext, \omegafirst)$
    \State $\hyptgttext \gets {\rm argmax}(\omegafirst)$ \Comment{$|\omegafirst|=\beamsizetext$}

    \State
    \State $\textdecout \gets \textit{GetHiddenStateFromCache}(\hyptgttext)$ \Comment{$\textdecout: (\textlen, \dmodel)$}
    \State $\ttuencout \gets \texttt{T2UEnc}(\textdecout)$ \Comment{$\ttuencout: (\textlen, \dmodel)$}
    \State

    \State \texttt{// Second-pass beam search}
    \State $\omegasecond \gets \{\}$
    \State $\omegasecond \gets \textit{BeamSearch}_{2}(\ttuencout, \beamsizeunit, \omegasecond)$
    \State $\hyptgtunit \gets {\rm argmax}(\omegasecond)$ \Comment{$|\omegasecond|=\beamsizeunit$}
    \State

    \State \texttt{// Convert discrete units to waveform}
    \State $\hyptgtwave \gets \texttt{UnitVocoder}(\hyptgtunit)$

    \State \Return $\hyptgtwave$
\EndFunction
\end{algorithmic}\label{algo:two_pass_beam}
\end{algorithm}

\section{Training with R-Drop}\label{appendix:rdrop}
{\unity} introduces an intermediate S2TT sub-task to make the optimization tractable while maintaining the end-to-end differentiability.
However, the easier S2TT task is more likely to overfit than the primary S2UT task.
To tackle this problem, we apply a more effective regularization based on R-Drop~\citep{wu2021rdrop} to the first-pass decoder in addition to standard regularization such as dropout~\citep{srivastava2014dropout} and label smoothing~\citep{label_smoothing}.
Theoretically, R-Drop reduces the inconsistency of model predictions between training and inference by dropout, thus improving the generalization ability.
R-Drop duplicates the network input during training and calculates two output probability distributions with different dropout masks.
Then, a constraint is introduced by minimizing the Kullback–Leibler (KL) divergence loss between the two probability distributions.
We apply R-Drop to both text and unit decoders.
The total training objective of {\unity} with R-Drop, $\losstotal$, is modified from Eq.~\eqref{eq:total_loss_unitvv} as follows:
\begin{multline}\label{eq:total_loss_unitvv_rdrop}
\losstotal = \sum_{i=1}^{2} \lossstu(\tgtunit|\srcspeechi,\tgttext) + \weightrdropstu \lossrdropstu(\srcspeechone,\srcspeechtwo) \\
+ \weightstt (\sum_{i=1}^{2} \lossstt(\tgttext|\srcspeechi) + \weightrdropstt \lossrdropstt(\srcspeechone,\srcspeechtwo)),
\end{multline}
where $\srcspeechi$ is a duplicated input from $\srcspeech$, $\lossrdropstu$ and $\lossrdropstt$ are KL losses for the unit and text decoders, $\weightstt$ is a weight for the S2TT loss, and $\weightrdropstu$ and $\weightrdropstt$ are weights for the KL losses, respectively.
We implement R-Drop by duplicating inputs instead of feeding them to the network twice.

Given a set of unique inputs ${\bf X}$, the general KL loss $\lossrdrop$ in R-Drop is formulated as follows:
\begin{multline*}
\lossrdrop({\bf X}_{1},{\bf X}_{2}) = \frac{1}{2} (\kldiv(P(\cdot|{\bf X}_{1}) || P(\cdot|{\bf X}_{2}) \\
+ \kldiv(P(\cdot|{\bf X}_{2})) || P(\cdot|{\bf X}_{1}))),
\end{multline*}
where ${\bf X}_{i}$ is a duplicated input from ${\bf X}$, $\kldiv$ is a KL divergence, and $P$ is a categorical probability distribution.

\section{Training objective}\label{appendix:training_objective}
In this section, we describe training objectives for the baseline S2ST models.
In addition to the primary S2ST/S2UT task, we introduce auxiliary S2TT and ASR tasks.
We adopted an auxiliary character-level ASR task for the direct S2ST models trained from scratch on Fisher, regardless of the choice of the output unit in the first-pass decoder.
We did not use the ASR task in the rest settings.

\paragraph{\specv}
The architecture of {\specv} is shown in Figure~\ref{fig:specv}.
Given the target spectrogram $\tgtspeech$, translation $\tgttext$, and transcription $\srctext$, corresponding to a source speech $\srcspeech$, the training objective of {\specv} is formulated as:
\begin{multline}\label{eq:total_loss_specv}
\losstotal = \losssts(\tgtspeech|\srcspeech) \\
+ \weightstt \lossstt(\tgttext|\srcspeech)
+ \weightasr \lossasr(\srctext|\srcspeech),
\end{multline}
where $\losssts$ is the primary S2ST loss, $\lossstt$ is the auxiliary S2TT loss, $\lossasr$ is the auxiliary ASR loss, $\weightstt$ is a weight for the S2TT loss, and $\weightasr$ is a weight for the ASR loss, respectively.
Note that R-Drop is not used because the output of the primary S2ST task is continuous.

We adopt the autoregressive decoder of Transformer TTS~\citep{li2019neural} as the spectrogram decoder.
Therefore, $\losssts$ is defined as a sum of the L1 loss $\losslone$, L2 loss $\lossltwo$, and end-of-sentence (EOS) prediction loss $\losseos$ as follows:
\begin{eqnarray*}
\losssts(\tgtspeech|\srcspeech) = \losslone + \lossltwo + \losseos.
\end{eqnarray*}

\paragraph{\specvvplus}
The architecture of {\specvvplus} is shown in Figure~\ref{fig:specvvplus}.
The training objective of {\specvvplus} is formulated as:
\begin{multline}\label{eq:total_loss_specvv_rdrop}
\losstotal = \sum_{i=1}^{2} \losssts(\tgtspeech|\srcspeechi,\tgttext) \\
+ \weightstt (\sum_{i=1}^{2} \lossstt(\tgttext|\srcspeechi) + \weightrdropstt \lossrdropstt(\srcspeechone, \srcspeechtwo)) \\
+ \weightasr (\sum_{i=1}^{2} \lossasr(\srctext|\srcspeechi) + \weightrdropasr \lossrdropasr(\srcspeechone,\srcspeechtwo)),
\end{multline}
where $\srcspeechi$ is a duplicated input from $\srcspeech$, $\lossrdropstt$ is the R-Drop's KL loss for the first-pass decoder, $\lossrdropstt$ is the R-Drop's KL loss for the auxiliary ASR decoder, and $\weightrdropstt$ and $\weightrdropasr$ are the corresponding weights for the R-Drop's KL losses, respectively.
Unlike Eq.~\eqref{eq:total_loss_specv}, the primary S2ST task depends on the output from the first-pass decoder.
We apply R-Drop to the S2TT and ASR tasks only.
We also investigated applying R-Drop to the second-pass spectrogram decoder by minimizing the difference of two outputs in the continuous space, but the training was unstable.

\paragraph{\unitv}
The architecture of {\unitv} is shown in Figure~\ref{fig:unitv}.
In addition to the primary S2UT loss and auxiliary S2TT and ASR losses, we use a CTC loss on top of the unit decoder following \citet{lee2021direct}.
The training objective of the {\unitv} model is formulated as:
\begin{multline}\label{eq:total_loss_unitv_rdrop}
\losstotal = \sum_{i=1}^{2} \lossstu(\tgtunit|\srcspeechi) + \weightrdropstu \lossrdropstu(\srcspeechone,\srcspeechtwo) \\
+ \weightctc \sum_{i=1}^{2} \lossctc(\tgttext|\unitdecout_{i}) \\
+ \weightstt (\sum_{i=1}^{2} \lossstt(\tgttext|\srcspeechi) + \weightrdropstt \lossrdropstt(\srcspeechone,\srcspeechtwo)) \\
+ \weightasr (\sum_{i=1}^{2} \lossasr(\srctext|\srcspeechi) + \weightrdropasr \lossrdropasr(\srcspeechone,\srcspeechtwo)),
\end{multline}
where $\lossstu$ is the primary S2UT loss, $\lossrdropstu$ is the R-Drop's KL loss for the unit decoder, $\lossctc$ is the CTC loss, $\unitdecout_{i}$ is the unit decoder output for the $i$-th forward pass, $\weightrdropstu$ is a weight for the R-Drop's KL loss, and $\weightctc$ is a weight for the CTC loss, respectively.
Unlike Eq.~\eqref{eq:total_loss_unitvv_rdrop}, there is no dependency between the primary S2UT task and auxiliary S2TT task except for sharing the same encoder.

\begin{table*}[t]
    \centering
    \begingroup
    \scalebox{0.65}{
    \begin{tabular}{lcc} \toprule
     \multirow{2}{*}{Corpus} & \multicolumn{2}{c}{Language direction} \\ \cmidrule(lr){2-3}
     & \textbf{En$\to$Es} & \textbf{Es$\to$En} \\
     \midrule[\heavyrulewidth]

     \multirow{3}{*}{\textbf{S2TT}}
       & \multirow{3}{*}{\shortstack{Europarl-ST [75.6 hours]~\citep{iranzo2020europarl}\\Must-C [495 hours]~\citep{di-gangi-etal-2019-must}}}
       & \multirow{3}{*}{\shortstack{CoVoST2 [112 hours]~\citep{wang2021covost}\\Europarl-ST [20.6 hours]\phantom{aaaaaaaaaaaaaaaaa}\\mTEDx [63.4 hours]~\citep{elizabeth2021multilingual}}} \\
     & & \\
     & & \\
     \midrule[\heavyrulewidth]

     \multirow{3}{*}{\textbf{ASR}}
       & \multirow{3}{*}{\shortstack{Librispeech [960 hours]~\citep{librispeech}\\TEDLIUM3 [452 hours]~\citep{tedlium}\\Common Voice v7 [1203 hours]~\citep{ardila-etal-2020-common}}}
       & \multirow{3}{*}{\shortstack{MLS [918 hours]~\citep{pratap2020mls}\\Common Voice v7 [290 hours]}} \\
     & & \\
     & & \\
     \midrule[\heavyrulewidth]

     \textbf{MT} \\
     \ \ Supervised MT1 & CCMatrix [86.3M sentences]~\citep{schwenk-etal-2021-ccmatrix} & -- \\
     \cmidrule(lr){1-3}

     \ \ \multirow{10}{*}{\shortstack{Supervised MT2\\(Cascaded S2ST)}}
       & \multicolumn{2}{c}{\multirow{10}{*}{\shortstack{OpenSubtitle2018 [60M sentences]~\citep{lison-etal-2018-opensubtitles2018}\\UNCorpus [21.8M sentences]~\citep{ziemski-etal-2016-united}\\EUBookshop v2 [5.2M sentences]~\citep{skadins-etal-2014-billions}\\Europarl v10 [1.9M sentences]~\citep{koehn-2005-europarl}\\Wikipedia v1.0 [1.8M sentences]~\citep{wolk2014building}\\TED2020 v1 [0.4M sentences]~\citep{reimers-gurevych-2020-making}\\Europarl-ST [32k sentences]\\Must-C [260k sentences]\\mTEDx [3.6k sentences]\\CoVosST2 [79k sentences]}} } \\
     & & \\
     & & \\
     & & \\
     & & \\
     & & \\
     & & \\
     & & \\
     & & \\
     & & \\
     \midrule[\heavyrulewidth]

     \textbf{T2U/TTS} & CSS100 [23.8 hours]~\citep{park2019css10} & LJSpeech [24 hours]~\citep{ljspeech17} \\
     \midrule[\heavyrulewidth]

     \multicolumn{1}{l}{\textbf{Unlabeled text}} \\
     \ \ t-mBART
       & \multicolumn{2}{c}{CC100 [5.6B tokens]~\citep{conneau2020unsupervised}} \\
     \midrule[\heavyrulewidth]

     \multicolumn{1}{l}{\textbf{Unlabeled speech}} \\
      \ \ wav2vec2.0
       & Libri-Light [60k hours]~\citep{kahn2020libri}
       & VoxPopuli Es [16k hours]~\citep{wang2021voxpopuli} \\
     \midrule

     \ \ \multirow{3}{*}{u-mBART}
       & \multicolumn{2}{c}{\multirow{3}{*}{\shortstack{VoxPopuli En [14k hours]\\VoxPopuli Es [16k hours]\\Libri-Light [60k hours]}}} \\
     & & \\
     & & \\
     \midrule

     \ \ \multirow{3}{*}{mHuBERT}
    & \multicolumn{2}{c}{\multirow{3}{*}{\shortstack{VoxPopuli En [4.5k hours]\\VoxPopuli Es [4.5k hours]\\VoxPopuli Fr [4.5k hours]}}} \\
     & & \\
     & & \\
     \bottomrule
    \end{tabular}
    }
    \endgroup
    \vspace{-1mm}
    \caption{Statistics for the multi-domain En$\leftrightarrow$Es corpora}
    \label{tab:data_statistics}
\end{table*}

\paragraph{S2TT, ASR}
We also apply R-Drop to S2TT and ASR tasks.
The training objective of the S2TT model is formulated as:
\begin{multline}\label{eq:total_loss_sttt}
\losstotal = \sum_{i=1}^{2} \lossstt(\tgttext|\srcspeechi) + \weightrdropstt \lossrdropstt(\srcspeechone,\srcspeechtwo).
\end{multline}
Similarly, the training objective of the ASR model is formulated as:
\begin{multline}\label{eq:total_loss_asr}
\losstotal = \sum_{i=1}^{2} \lossasr(\srctext|\srcspeechi) + \weightrdropasr \lossrdropasr(\srcspeechone,\srcspeechtwo).
\end{multline}

\begin{table*}[t]
    \centering
    \begingroup
    \scalebox{0.75}{
    \begin{tabular}{ll} \toprule
     Model & URL \\
    \midrule[\heavyrulewidth]

     En wav2vec2.0 & \tiny{\url{https://github.com/facebookresearch/fairseq/blob/main/examples/speech_to_speech/docs/enhanced_direct_s2st_discrete_units.md\#wav2vec-20}} \\
     Es wav2vec2.0 & \tiny{\url{https://github.com/facebookresearch/fairseq/blob/main/examples/speech_to_speech/docs/enhanced_direct_s2st_discrete_units.md\#wav2vec-20}} \\

     En HuBERT & \scriptsize{\url{https://github.com/facebookresearch/fairseq/blob/main/examples/speech_to_speech/docs/direct_s2st_discrete_units.md}} \\
     mHuBERT & \scriptsize{\url{https://github.com/facebookresearch/fairseq/blob/main/examples/speech_to_speech/docs/textless_s2st_real_data.md}} \\
     \midrule[\heavyrulewidth]

     En-Es u-mBART & \scriptsize{\url{https://dl.fbaipublicfiles.com/fairseq/speech_to_speech/s2st_finetuning/unit_mBART/checkpoint.pt}} \\
     \midrule[\heavyrulewidth]

     En Transformer TTS & \scriptsize{\url{https://huggingface.co/facebook/tts_transformer-en-ljspeech}} \\
     Es Transformer TTS & \scriptsize{\url{https://huggingface.co/facebook/tts_transformer-es-css10}} \\
    \bottomrule
    \end{tabular}
    }
    \vspace{-1mm}
    \caption{Links to pre-trained self-supervised models and TTS models}
    \label{tab:results_model_link}
    \endgroup
\end{table*}

\section{Data}\label{appendix:data}

\paragraph{Fisher Es$\to$En~\citep{fisher_callhome}}
This corpus contains 170-hour Spanish conversational telephone speech with the corresponding transcriptions as well as the English translations.
The target speech is synthesized by a high-quality in-house TTS model trained with a single female speaker~\citep{lee2021direct}.

\paragraph{CVSS-C~\citep{jia-etal-2022-cvss}}
CVSS is a public multilingual S2ST corpus based on CoVoST2~\citep{wang2021covost}.
It covers 21 language directions to English.
We use the CVSS-C part of the CVSS corpus, in which a single-speaker female TTS synthesizes the target speech.
We combine all language directions to train a single X-to-En multilingual model.

\paragraph{Multi-domain En$\leftrightarrow$Es~\citep{popuri2022enhanced}}
Following \citet{popuri2022enhanced}, we use all samples from multiple public S2TT corpora in each direction to improve the robustness of model training~\citep{jia2022translatotron,chan2021speechstew}.
We also use all samples from validation sets in all domains for checkpoint selection.
We further augment the S2ST training data by pseudo-labeling ASR corpora with MT and T2U/TTS models.
We use the TTS model in the cascaded system to synthesize the target speech for direct speech-to-spectrogram models.
For direct speech-to-unit models, we use a T2U model~\citep{lee2021textless} to generate discrete units on the ASR corpora and the TTS+HuBERT pipeline for the S2T corpora.
Both T2U and TTS models are based on Transformer.
We train En and Es T2U/TTS models on the LJSpeech~\citep{ljspeech17} and CSS10~\citep{park2019css10} corpora, respectively.

For \textbf{En$\to$Es}, we use all samples from Europarl-ST~\citep{iranzo2020europarl} and Must-C~\citep{di-gangi-etal-2019-must} and augment the training data by TEDLIUM3~\citep{tedlium}, Librispeech~\citep{librispeech}, and Common Voice~\citep{ardila-etal-2020-common}, resulting in 3180-hour source speech.
We removed samples overlapped with mtedx dev/test sets from TEDLIUM3.
For \textbf{Es$\to$En}, we use all samples from CoVoST2, Europarl-ST, and mTEDx~\citep{elizabeth2021multilingual}, and augment the training data by Common Voice and multilingual Librispeech (MLS)~\citep{pratap2020mls}, resulting in 1404-hour source speech.
In Table~\ref{tab:data_statistics}, we list all the datasets used in each task.

\section{Pre-processing}\label{appendix:preprocessing}
\paragraph{Speech}
We convert source audio to 16kHz and generate target speech with 22kHz.
When extracting discrete units, we downsample the target speech to 16kHz.
For filterbank features, we extract 80-dimensional coefficients on both the source and target sides.
We apply utterance-level cepstral mean-variance normalization to both inputs.

\paragraph{Discrete units}
We extract discrete units with an English HuBERT trained on Librispeech after performing k-means clustering with 100 clusters on Fisher~\citep{lee2021direct}.
For the rest corpora, we extract discrete units with a multilingual HuBERT (mHuBERT)~\citep{popuri2022enhanced} trained on En, En, and Fr parts of VoxPopuli~\citep{wang2021voxpopuli} with the number of k-means clusters of 1000.

\paragraph{Text}
We lowercase text data and remove all punctuation marks except for apostrophes.
When initializing the text decoder in two-pass direct S2ST models randomly, we build vocabularies of 1k, 6k, and 2k unigram subword units~\citep{kudo-2018-subword} with the SentencePiece toolkit~\citep{kudo-richardson-2018-sentencepiece} for the Fisher, CVSS-C, and multi-domain corpora, respectively.
When pre-training the text decoder with t-mBART, we use the same vocabulary as t-mBART.
The reference target translation to calculate ASR-BLEU is normalized with lowercasing, removal of punctuation marks, conversion of digits to spoken forms, and removal of non-verbal words in parentheses like “(Applause)” or “(Music).”

\paragraph{Data filtering}
For discrete unit generation with a T2U model, we found that target discrete units were over-generated in long-form samples.
We filtered out such samples by thresholding with a ratio of the sequence length of the discrete units over the number of corresponding source speech frames.
We used a threshold of 0.7 for the multi-domain En$\to$Es corpus while using $\infty$ for the rest.
We used the same number of samples for all direct S2ST models for a fair comparison.

\begin{table*}[t]
    \centering
    \begingroup
    \scalebox{0.65}{
    \begin{tabular}{lccccccccccccc} \toprule
     \multirow{2}{*}{ID} & \multirow{2}{*}{\#GPU} & \multirow{2}{*}{\shortstack{\# of frames $\times$\\gradient accumulation}} & \multirow{2}{*}{\shortstack{Learning\\rate}} &\multirow{2}{*}{Warmup} & \multirow{2}{*}{Dropout} & \multirow{2}{*}{\shortstack{Label\\smoothing}} & \multicolumn{3}{c}{Loss weight} & \multicolumn{3}{c}{R-Drop} &  \\ \cmidrule(lr){8-10} \cmidrule(lr){11-13}
     & & & & & & & $\weightasr$ & $\weightstt$ & $\weightctc$ & $\weightrdropasr$ & $\weightrdropstt$ & $\weightrdropstu$ \\
    \midrule[\heavyrulewidth]

     \texttt{A6} & \phantom{a}4 & 40k$\times$1 & 1.3e-3 & \multirow{10}{*}{10k} & 0.2 & \multirow{10}{*}{0.2} & -- & -- & -- & -- & \phantom{a}8.6 & -- \\
     \texttt{A7} & 16 & \phantom{a}2k$\times$4 & 1.0e-3 & \phantom{a} & 0.1 & \phantom{a} & -- & -- & -- & -- & \phantom{a}8.6 & -- \\  %
    \texttt{A11} & 16 & 20k$\times$1 & 1.0e-3 & \phantom{a} & 0.3 & & 0.1 & 0.1 & -- & 0.0 & \phantom{a}0.0 & -- \\  %
     \texttt{A12} & 16 & \phantom{a}4k$\times$2 & 1.0e-3 & \phantom{a} & 0.1 & \phantom{a} & -- & -- & -- & -- & \phantom{a}-- & -- \\  %
     \texttt{A15} & 16 & 20k$\times$1 & 1.5e-3 & \phantom{a} & 0.3 & \phantom{a} & 0.1 & 0.1 & -- & 3.0 & \phantom{a}3.0 & -- \\  %
     \texttt{A16} & 16 & \phantom{a}4k$\times$2 & 1.0e-3 & \phantom{a} & 0.1 & \phantom{a} & -- & 0.1 & -- & -- & \phantom{a}3.0 & -- \\  %
     \texttt{A18} & \phantom{a}4 & 20k$\times$1 & 8.6e-4 & \phantom{a} & 0.3 & \phantom{a} & 8.0 & 8.0 & 1.6 & 1.0 & \phantom{a}1.0 & 1.0 \\  %
     \texttt{A19} & 16 & \phantom{a}2k$\times$4 & 1.0e-3 & \phantom{a} & 0.1 & \phantom{a} & -- & -- & -- & -- & \phantom{a}-- & 1.0 \\  %
     \texttt{A20} & \phantom{a}4 & 20k$\times$1 & 6.0e-4 & \phantom{a} & 0.3 & \phantom{a} & 8.0 & 8.0 & -- & 3.0 & \phantom{a}3.0 & 1.0 \\  %
     \texttt{A21} & 16 & \phantom{a}2k$\times$4 & 1.0e-3 & \phantom{a} & 0.1 & \phantom{a} & -- & 8.0 & -- & -- & \phantom{a}3.0 & 1.0 \\  %
     \midrule[\heavyrulewidth]

     \texttt{B3} & \phantom{a}8 & 35k$\times$4 & 2.1e-3 & \multirow{14}{*}{10k} & 0.1 & \todo{0.1} & 0.6 & -- & -- & 4.6 & \phantom{a}4.6 & -- \\  %
     \texttt{B4} & 32 & \phantom{a}2k$\times$24 & 1.0e-3 & \phantom{a} & 0.1 & 0.2 & 0.0 & -- & -- & 5.0 & \phantom{a}5.0 & -- \\  %

     \texttt{B6} & 32 & 40k$\times$1 & 1.0e-3 & \phantom{a} & 0.1 & 0.2 & -- & 0.1 & -- & -- & \phantom{a}0.0 & -- \\  %
     \texttt{B7} & 32 & 40k$\times$1 & 1.0e-3 & \phantom{a} & 0.1 & 0.2 & -- & 0.1 & -- & -- & \phantom{a}0.0 & -- \\  %
     \texttt{B8} & 32 & \phantom{a}2k$\times$24 & 1.0e-3 & \phantom{a} & 0.1 & 0.2 & -- & -- & -- & -- & \phantom{a}-- & -- \\  %

     \texttt{B15} & 32 & 40k$\times$1 & 1.1e-3 & \phantom{a} & 0.1 & 0.2 & -- & 0.1 & -- & -- & 10.0 & -- \\  %
     \texttt{B16} & 32 & 40k$\times$1 & 1.0e-3 & \phantom{a} & 0.1 & 0.2 & -- & 0.1 & -- & -- & 10.0 & -- \\  %
     \texttt{B17} & 32 & \phantom{a}2k$\times$24 & 1.0e-3 & \phantom{a} & 0.1 & 0.2 & -- & 0.1 & -- & -- & \phantom{a}5.0 & -- \\  %

     \texttt{B18} & 32 & 20k$\times$2 & 8.6e-4 & \phantom{a} & 0.3 & 0.2 & -- & 8.0 & 1.6 & -- & \phantom{a}0.5 & 0.5 \\  %
     \texttt{B19} & 32 & 20k$\times$2 & 7.0e-4 & \phantom{a} & 0.3 & 0.2 & -- & 8.0 & 1.6 & -- & \phantom{a}0.5 & 0.5 \\  %
     \texttt{B20} & 32 & \phantom{a}2k$\times$24 & 1.0e-3 & \phantom{a} & 0.1 & 0.2 & -- & -- & -- & -- & \phantom{a}-- & 0.5 \\  %

     \texttt{B21} & 32 & 20k$\times$2 & 1.5e-3 & \phantom{a} & 0.3 & 0.2 & -- & 8.0 & -- & -- & \phantom{a}1.5 & 1.5 \\  %
     \texttt{B22} & 32 & 20k$\times$2 & 7.0e-4 & \phantom{a} & 0.3 & 0.2 & -- & 8.0 & -- & -- & \phantom{a}5.0 & 1.5 \\  %
     \texttt{B23} & 32 & \phantom{a}2k$\times$24 & 1.0e-3 & \phantom{a} & 0.1 & 0.2 & -- & 8.0 & -- & -- & \phantom{a}5.0 & 1.5 \\  %
     \midrule[\heavyrulewidth]

     \texttt{C1'} & \multirow{7}{*}{32} & \multirow{7}{*}{\phantom{a}2k$\times$30} & \multirow{7}{*}{5.0e-4} & 1k & \multirow{7}{*}{0.1} & 0.1 & -- & -- & -- & -- & 10.0 & --  \\
     \texttt{C2'} & \phantom{a} & \phantom{a} & \phantom{a} & 1k & \phantom{a} & 0.2 & -- & -- & -- & -- & 10.0 & --  \\
     \texttt{C3} & \phantom{a} & \phantom{a} & \phantom{a} & 5k & \phantom{a} & 0.2 & -- & 8.0 & -- & -- & 10.0 & -- \\  %
     \texttt{C4} & \phantom{a} & \phantom{a} & \phantom{a} & 5k & \phantom{a} & 0.2 & -- & 8.0 & -- & -- & 10.0 & -- \\  %
     \texttt{C5'} & \phantom{a} & & & 1k & \phantom{a} & 0.2 & -- & -- & -- & -- & \phantom{a}-- & 0.0 \\  %
     \texttt{C6} & \phantom{a} & \phantom{a} & \phantom{a} & 1k & \phantom{a} & 0.2 & -- & 8.0 & -- & -- & 10.0 & 0.0 \\  %
     \texttt{C7} & \phantom{a} & \phantom{a} & \phantom{a} & 1k & \phantom{a} & 0.2 & -- & 8.0 & -- & -- & 10.0 & 0.0 \\  %

    \bottomrule
    \end{tabular}
    }
    \vspace{-1mm}
    \caption{Training hyperparameters}
    \label{tab:training_hyperparameter}
    \endgroup
\end{table*}

\section{Pre-training}\label{appendix:pretraining}
In Table~\ref{tab:results_model_link}, we list all the pre-trained self-supervised models and TTS models used in \cref{sec:result}.

\paragraph{wav2vec2.0}
We use 24-layer Conformer wav2vec2.0~\citep{baevski2020wav2vec} models trained on Libri-Light~\citep{kahn2020libri} for En and VoxPopuli for Es, respectively.

\paragraph{w2v-BERT}
Same as~\citet{jia2019leveraging}, we pre-train the w2v-BERT~\citep{chung2021w2v} on approximately~430k hours of unlabeled speech data in~51 languages spanning from VoxPopuli, Common Voice, MLS, BABEL~\citep{babel,gales2014speech}, and VoxLingua107~\citep{valk2021voxlingua107}.
The w2v-BERT was composed of~24 Conformer layers with~0.6 billions of parameters.

\paragraph{Text-based mBART (t-mBART)}
We train a t-mBART model with En and Es unlabeled text on CC100~\citep{conneau2020unsupervised}.
We use of a 65k unigram subword unit for the vocabulary.
For multilingual experiments on CVSS-C, we use mBART50~\citep{tang2020multilingual} with multilingual fine-tuning to En.
The vocabulary size is a 250k subword unit.

\paragraph{Unit-based mBART (u-mBART)}
We use a u-mBART model trained with En and Es unlabeled speech on VoxPopuli.
The unit vocabulary is the same as that of the mHuBERT model.

\begin{table*}[t]
    \centering
    \begingroup
    \scalebox{0.70}{
    \begin{tabular}{lllccc} \toprule
    \multirow{2}{*}{ID} & \multirow{2}{*}{Model} & \multirow{2}{*}{Encoder}  & \multicolumn{3}{c}{ASR-BLEU ($\uparrow$)} \\
    \cmidrule(lr){4-6}
     & & & dev & dev2 & test \\
     \midrule[\heavyrulewidth]

     \texttt{A0} & \multicolumn{2}{l}{Synthetic target~\citep{lee2021direct}} & 88.5 &  89.4 & 90.5 \\ %
     \midrule[\heavyrulewidth]

     \multicolumn{6}{l}{\textbf {Cascaded systems}} \\
     \texttt{A1} & \multirow{1}{*}{ASR $\to$ MT $\to$ TTS} & LSTM~\citep{lee2021direct} & 42.1 & 43.5 & 43.9 \\  %
     \cmidrule(lr){1-6}

     \texttt{A2} & \multirow{6}{*}{S2TT $\to$ TTS}
     & LSTM~\citep{jia2019direct} & 39.4 & 41.2 & 41.4 \\
     \texttt{A3} & & LSTM~\citep{jia2022translatotron} & -- & -- & 43.3 \\ %
     \texttt{A4} & & LSTM~\citep{lee2021direct} & 38.5 & 39.9 & 40.2 \\ %
     \texttt{A5} & & Transformer~\citep{dong2022leveraging} & 44.3 & 45.4 & 45.1 \\
     \texttt{A6} & & Conformer & 47.8 & 48.9 & 48.3 \\  %
     \texttt{A7} & & Conformer wav2vec2.0 & 51.0 & 52.2 & 52.1 \\  %
     \midrule[\heavyrulewidth]

     \multicolumn{6}{l}{\textbf {Direct speech-to-spectrogram systems}} \\
     \texttt{A8} & \multirow{5}{*}{\specv}
     & Transformer~\citep{jia2019direct} & 30.1 & 31.5 & 31.1 \\
     \texttt{A9} & & Transformer~\citep{lee2021direct} & -- & -- & 33.2 \\
     \texttt{A10} & & Transformer~\citep{dong2022leveraging} & 42.4 & 43.3 & 43.6 \\
     \texttt{A11} & & Conformer & 43.9 & 44.4 & 43.8 \\  %
     \texttt{A12} & & Conformer wav2vec2.0 & 45.5 & 47.6 & 46.3 \\  %
     \cmidrule(lr){1-6}

     \texttt{A13} & \multirow{2}{*}{\translatotronvv}
     & Conformer~\citep{jia2022translatotron} & -- & -- & 42.4 \\ %
     \texttt{A14} & & Conformer w2v-BERT~\citep{li2022textless} & -- & -- & 52.2 \\
     \cmidrule(lr){1-6}

     \texttt{A15} & \multirow{2}{*}{\specvvplus} & Conformer & 50.4 & 51.1 & 50.8 \\  %
     \texttt{A16} & & Conformer wav2vec2.0 & {\bf 58.4} & {\bf 59.5} & {\bf 58.6} \\  %
     \midrule[\heavyrulewidth]

     \multicolumn{6}{l}{\textbf {Direct speech-to-unit systems}} \\
     \texttt{A17}& \multirow{3}{*}{\unitv}
     & Transformer~\citep{lee2021direct} & -- & -- & 39.9 \\ %
     \texttt{A18} & & Conformer & 46.2 & 47.6 & 47.4 \\  %
     \texttt{A19} & & Conformer wav2vec2.0 & 53.4 & 53.9 & 53.7 \\  %
     \cmidrule(lr){1-6}

     \texttt{A20} & \multirow{2}{*}{\unity}
     & Conformer & 50.5 & 51.6 & 51.4 \\  %
     \texttt{A21} & & Conformer wav2vec2.0 & {\bf 55.1} & {\bf 56.5} & {\bf 55.9} \\  %
     \bottomrule
    \end{tabular}
    }
    \endgroup
    \vspace{-1mm}
    \caption{ASR-BLEU on Fisher Es$\to$En corpus. The decoder in all the models is initialized randomly. {\specvvplus} is our improved version of {\translatotronvv}. Note that \texttt{A10} uses pseudo labeled external resources with a cascaded S2ST system, and \texttt{A13} uses data augmentation by concatenating multiple utterances.}
    \label{tab:results_fisher_s2st}
\end{table*}

\section{Architecture details}\label{appendix:hyperparameter_architecture}
Let $\dmodel$ be a model dimension of Transformer, $\dff$ be an inner dimension of the FFN layers, and $\nheads$ be the number of attention heads.

\paragraph{Speech encoder}
We used a 16-layer Conformer encoder stacked on 2-dimensional convolution blocks when training models from scratch.
The convolution blocks reduced the input sequence length by a factor of 4.
We set $(\dmodel, \dff, \nheads)$ to $(256, 2048, 4)$.
We set the kernel size of the depthwise convolution in the convolution module of each Conformer block to 31.
When pre-training the encoder with wav2vec2.0 and w2v-BERT, we used a 24-layer Conformer encoder and stacked a one-layer length adaptor~\citep{li-etal-2021-multilingual} on it.
Because an output frame of wav2vec2.0 and w2v-BERT corresponds to 20ms and the length adaptor halved the sequence length, the frame rate of every final encoder output corresponds to 40ms in both cases.
In this case, we set $(\dmodel, \dff, \nheads)$ to $(1024, 4096, 16)$.

\paragraph{\specv}
We used a six-layer Transformer spectrogram decoder.
We set $(\dmodel, \dff, \nheads)$ to $(512, 2048, 8)$.
When pre-training the speech encoder with wav2vec2.0 or w2v-BERT, we doubled these three values.
We set the pre-net dimension and reduction factor of the spectrogram decoder to 32 and 3, respectively.

\paragraph{\specvvplus}
Let $\nlayerttsenc$ be the depth of the T2S encoder.
We set $(\nlayerfirstdec, \nlayerseconddec, \nlayerttsenc)$ to $(4, 6, 2)$ on Fisher and CVSS-C.
On the multi-domain corpus, we set $(\nlayerfirstdec, \nlayerseconddec, \nlayerttsenc)$ to $(12, 6, 2)$ when pre-training the first-pass decoder with t-mBART.
Otherwise, we set $(\nlayerfirstdec, \nlayerseconddec, \nlayerttsenc)$ to $(6, 6, 2)$.
We used the same $\dmodel$, $\dff$, and $\nheads$ as {\specv} in all the settings.

\paragraph{\unitv}
We used a six-layer Transformer unit decoder.
When training models from scratch on Fisher, we set $(\dmodel, \dff, \nheads)$ to $(256, 2048, 4)$.
We set $(\dmodel, \dff, \nheads)$ to $(512, 2048, 8)$ on CVSS-C.
When pre-training the speech encoder with wav2vec2.0 or w2v-BERT, we set $(\dmodel, \dff, \nheads)$ to $(1024, 4096, 16)$.

\paragraph{\unity}
We used the same first-pass decoder as {\specvvplus} in all the settings.
We set $(\nlayerseconddec, \nlayerttuenc)$ to $(2, 2)$.
We used the same $\dmodel$, $\dff$, and $\nheads$ as the {\unitv} model in all the settings.

\paragraph{S2TT}
We used a six-layer Transformer decoder.
When initializing it with t-mBART, we set the depth to 12.

\paragraph{ASR}
We used the same architecture as the S2TT model except for the vocabulary in all the settings.

\section{Training details}\label{appendix:hyperparameter_training}
We optimized all models with the mixed precision training using 32GB V100 GPUs~\citep{micikevicius2018mixed}.
When fine-tuning the speech encoder from wav2vec2.0 and w2v-BERT, we updated all parameters in the speech encoder.
For multilingual training with speech encoder pre-training with w2v-BERT on CVSS-C, we over-sampled training data of low-resource directions with an inverse temperature of 0.6, following \citep{arivazhagan2019massively}.
We list the training hyperparameters in Table~\ref{tab:training_hyperparameter}.
The training of \texttt{A*}, \texttt{B*}, and \texttt{C*} models converged within approximately 1, 3, and 5 days, respectively.

\begin{figure}[t]
  \centering
  \includegraphics[width=0.95\linewidth]{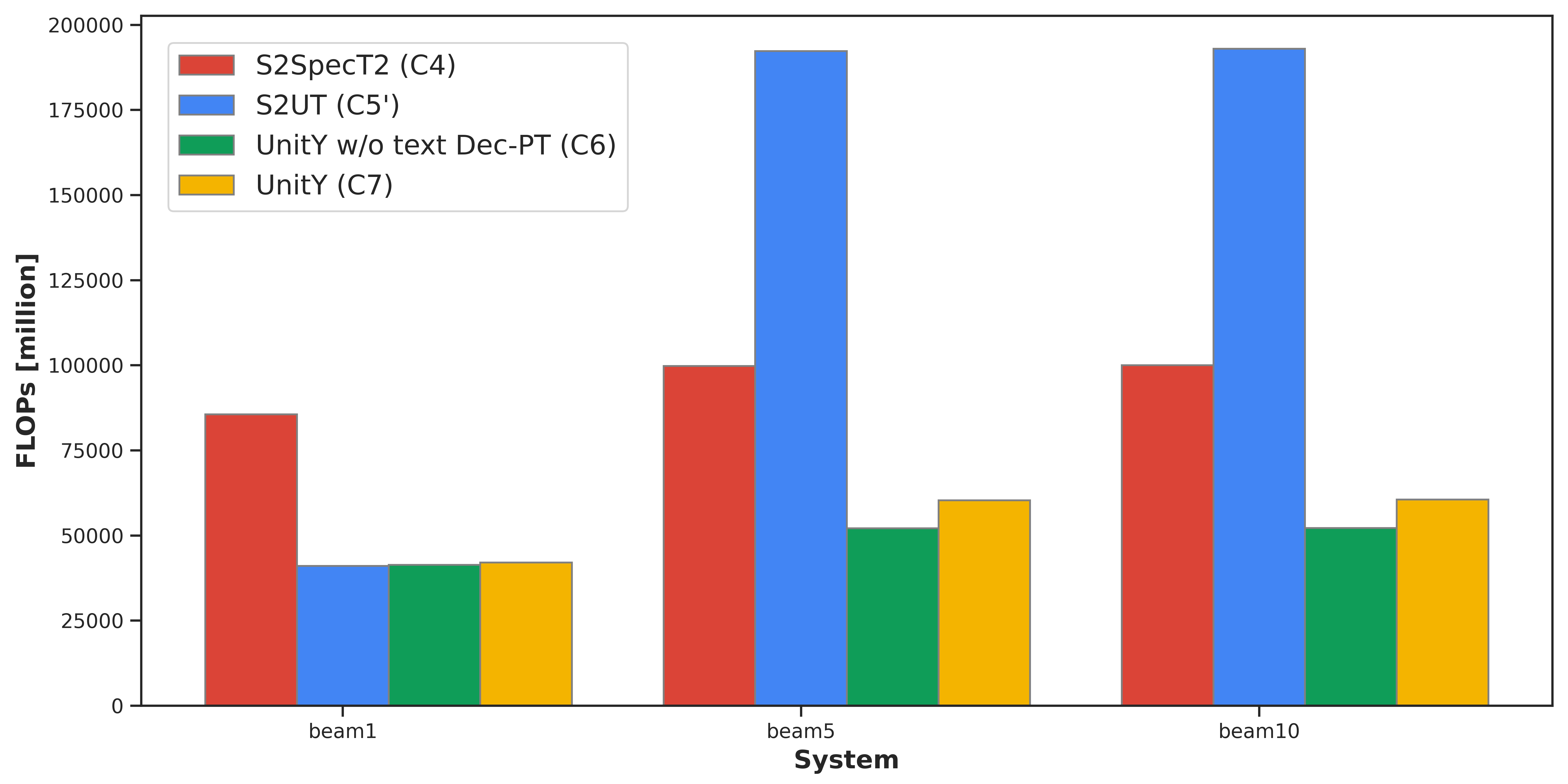}
  \vspace{-2mm}
  \caption{FLOPs of direct S2ST models on multi-domain Es$\to$En corpus. The beam width of two-pass models corresponds to the first-pass decoder.}
  \label{fig:flops}
\end{figure}

\begin{table*}[t]
    \centering
    \begingroup
    \scalebox{0.53}{
    \begin{tabular}{llllccccccccccccccccccccccc} \toprule
    \multirow{3}{*}{ID} & \multirow{3}{*}{Model} & \multicolumn{22}{c}{ASR-BLEU ($\uparrow$)} \\ \cmidrule(lr){3-24}
     & & \multirow{2}{*}{Avg.} & \multicolumn{4}{c}{High} & \multicolumn{5}{c}{Mid} & \multicolumn{12}{c}{Low} \\ \cmidrule(lr){4-7} \cmidrule(lr){8-12} \cmidrule(lr){13-24}
     & & & fr & de & ca & es & fa & it & ru & zh & pt & nl & tr & et & mn & ar & lv & sl & sv & cy & ta & ja & id \\
     \midrule[\heavyrulewidth]
     \texttt{B0} & Synthetic target${}^{\diamondsuit}$ & 91.1 & 84.6 & 88.4 & 92.0 & 88.6 & 91.7 & 89.5 & 94.0 & 77.8 & 93.1 & 90.6 & 92.7 & 89.3 & 92.4 & 94.2 & 94.8 & 94.9 & 94.1 & 92.0 & 90.6 & 95.3 & 92.6 \\
     \midrule[\heavyrulewidth]

     \multicolumn{6}{l}{\textbf {Cascaded systems}} \\
     \texttt{B1} & S2TT $\to$ TTS${}^{\diamondsuit}$ & 10.6 & 31.2 & 23.9 & 26.8 & 33.3 & \phantom{a}3.4 & 28.1 & 24.4 & \phantom{a}6.8 & 14.8 & \phantom{a}9.8 & \phantom{a}5.1 & \phantom{a}1.7 & \phantom{a}0.3 & \phantom{a}4.1 & \phantom{a}2.3 & \phantom{a}0.6 & \phantom{a}1.4 & \phantom{a}2.1 & \phantom{a}0.2 & \phantom{a}0.7 & \phantom{a}0.9 \\
     \texttt{B2} & \ + ASR pre-training & 12.7 & 32.9 & 26.2 & 28.6 & 34.9 & \phantom{a}5.6 & 30.2 & 27.1 & \phantom{a}8.7 & 19.8 & 14.4 & 10.7 & \phantom{a}3.2 & \phantom{a}0.6 & \phantom{a}7.8 & \phantom{a}2.8 & \phantom{a}2.0 & \phantom{a}3.4 & \phantom{a}5.0 & \phantom{a}0.2 & \phantom{a}0.9 & \phantom{a}1.6 \\
     \texttt{B3} & S2TT $\to$ TTS & \phantom{a}7.8 & 18.3 & 16.1 & 18.5 & 19.9 & \phantom{a}4.2 & 18.1 & 17.6 & \phantom{a}3.7 & 15.8 & 11.5 & \phantom{a}6.5 & \phantom{a}2.1 & \phantom{a}0.2 & \phantom{a}2.2 & \phantom{a}1.3 & \phantom{a}2.3 & \phantom{a}1.0 & \phantom{a}2.9 & \phantom{a}0.2 & \phantom{a}0.3 & \phantom{a}0.3 \\  %
     \texttt{B4} & \ + w2v-BERT + t-mBART & 14.9 & 20.5 & 20.0 & 21.6 & 22.1 & \phantom{a}8.5 & 21.8 & 27.6 & \phantom{a}5.5 & 27.6 & 21.6 & 13.6 & 13.2 & \phantom{a}1.7 & 12.7 & 10.6 & 17.4 & 18.5 & 11.5 & \phantom{a}1.3 & \phantom{a}3.7 & 12.0 \\
     \midrule[\heavyrulewidth]

     \multicolumn{6}{l}{\textbf {Direct speech-to-spectrogram systems}} \\
     \texttt{B5} & {\translatotronv}${}^{\diamondsuit}$ & \phantom{a}3.4 & 15.5 & \phantom{a}6.9 & 11.0 & 14.1 & \phantom{a}1.4 & \phantom{a}9.3 & \phantom{a}4.3 & \phantom{a}1.5 & \phantom{a}2.2 & \phantom{a}2.1 & \phantom{a}1.2 & \phantom{a}0.1 & \phantom{a}0.1 & \phantom{a}0.1 & \phantom{a}0.2 & \phantom{a}0.3 & \phantom{a}0.4 & \phantom{a}0.3 & \phantom{a}0.1 & \phantom{a}0.2 & \phantom{a}0.1 \\
     \texttt{B6} & {\specv} & \phantom{a}7.6 & 24.1 & 17.8 & 20.3 & 25.1 & \phantom{a}1.8 & 20.3 & 18.7 & \phantom{a}2.5 & \phantom{a}9.8 & \phantom{a}9.0 & \phantom{a}3.8 & \phantom{a}0.5 & \phantom{a}0.1 & \phantom{a}0.5 & \phantom{a}1.2 & \phantom{a}0.8 & \phantom{a}1.3 & \phantom{a}0.7 & \phantom{a}0.1 & \phantom{a}0.2 & \phantom{a}0.2 \\
     \texttt{B7} & \ + S2TT pre-training & \phantom{a}9.6 & 26.3 & 20.1 & 21.8 & 27.5 & \phantom{a}6.2 & 22.3 & 21.9 & \phantom{a}5.7 & 12.6 & 11.4 & \phantom{a}9.1 & \phantom{a}2.7 & \phantom{a}0.3 & \phantom{a}4.3 & \phantom{a}1.3 & \phantom{a}1.8 & \phantom{a}1.5 & \phantom{a}4.0 & \phantom{a}0.3 & \phantom{a}0.6 & \phantom{a}0.7 \\
     \texttt{B8} & \ + w2v-BERT & 16.6 & 31.8 & 27.3 & 28.4 & 34.4 & \phantom{a}8.9 & 30.0 & 34.1 & \phantom{a}5.0 & 31.6 & 23.3 & 11.5 & 10.0 & \phantom{a}0.3 & 10.8 & 14.4 & 14.5 & 22.4 & \phantom{a}4.8 & \phantom{a}0.1 & \phantom{a}0.6 & \phantom{a}5.3 \\  %
     \cmidrule(lr){1-24}

     \texttt{B9} & {\translatotronvv}${}^{\diamondsuit}$ & \phantom{a}8.7 & 28.3 & 19.7 & 23.5 & 30.1 & \phantom{a}2.4 & 24.1 & 19.6 & \phantom{a}4.5 & 12.5 & \phantom{a}6.5 & \phantom{a}3.8 & \phantom{a}0.6 & \phantom{a}0.2 & \phantom{a}1.7 & \phantom{a}1.5 & \phantom{a}0.4 & \phantom{a}1.3 & \phantom{a}0.9 & \phantom{a}0.1 & \phantom{a}0.5 & \phantom{a}0.4 \\
     \texttt{B10} & \ + Transformer decoder${}^{\spadesuit}$ & 10.1 & 29.5 & 22.3 & 25.0 & 30.8 & \phantom{a}3.4 & 26.0 & 21.7 & \phantom{a}5.5 & 14.3 & 10.5 & \phantom{a}6.6 & \phantom{a}1.1 & \phantom{a}0.2 & \phantom{a}3.8 & \phantom{a}3.0 & \phantom{a}2.3 & \phantom{a}2.8 & \phantom{a}1.6 & \phantom{a}0.1 & \phantom{a}0.5 & \phantom{a}0.8 \\
     \texttt{B11} & \ + S2TT pre-training${}^{\diamondsuit}$ & 12.0 & 32.4 & 24.8 & 28.2 & 33.4 & \phantom{a}6.3 & 28.6 & 23.2 & \phantom{a}6.3 & 18.3 & 15.8 & 10.6 & \phantom{a}2.5 & \phantom{a}0.4 & \phantom{a}5.4 & \phantom{a}2.3 & \phantom{a}3.1 & \phantom{a}3.2 & \phantom{a}4.5 & \phantom{a}0.1 & \phantom{a}1.0 & \phantom{a}1.0 \\
     \texttt{B12} & \ + w2v-BERT${}^{\spadesuit}$ & 17.9 & 33.6 & 30.6 & 30.1 & 35.9 & \phantom{a}6.0 & 32.5 & 38.9 & \phantom{a}5.2 & 31.9 & 29.3 & \phantom{a}9.2 & 16.0 & \phantom{a}0.2 & 10.4 & 15.6 & 17.8 & 25.9 & \phantom{a}4.2 & \phantom{a}0.3 & \phantom{a}0.9 & \phantom{a}1.5 \\
     \texttt{B13} & \ + mSLAM${}^{\spadesuit}$ & 19.3 & 33.9 & 31.5 & 30.6 & 36.8 & \phantom{a}7.2 & 33.7 & 41.6 & \phantom{a}6.4 & 34.1 & 31.1 & 16.1 & 17.1 & \phantom{a}0.3 & 10.0 & 14.4 & 22.9 & 28.4 & \phantom{a}5.4 & \phantom{a}0.2 & \phantom{a}1.3 & \phantom{a}2.5 \\
     \texttt{B14} & \ \ ++ TTS augmentation${}^{\spadesuit}$ & 22.0 & 34.5 & 32.0 & 30.7 & 37.1 & \phantom{a}8.2 & 33.8 & {\bf 42.6} & 10.6 & 34.0 & 31.8 & {\bf 23.9} & 17.2 & \phantom{a}1.1 & 22.4 & 15.6 & 23.3 & 31.1 & \phantom{a}7.6 & \phantom{a}0.6 & \phantom{a}5.5 & 18.5 \\
     \cmidrule(lr){1-24}

     \texttt{B15} & {\specvvplus} & 11.3 & 31.7 & 25.9 & 27.4 & 32.8 & \phantom{a}4.6 & 28.4 & 27.5 & \phantom{a}7.0 & 18.0 & 15.4 & \phantom{a}9.2 & \phantom{a}1.7 & \phantom{a}0.3 & \phantom{a}1.7 & \phantom{a}2.5 & \phantom{a}1.3 & \phantom{a}1.8 & \phantom{a}1.9 & \phantom{a}0.2 & \phantom{a}0.7 & \phantom{a}1.0 \\  %
     \texttt{B16} & \ + S2TT pre-training & 13.1 & 31.9 & 26.1 & 28.0 & 33.3 & \phantom{a}7.9 & 28.8 & 28.6 & \phantom{a}8.5 & 20.3 & 17.8 & 13.9 & \phantom{a}4.6 & \phantom{a}0.6 & \phantom{a}6.4 & \phantom{a}2.6 & \phantom{a}4.8 & \phantom{a}2.4 & \phantom{a}7.4 & \phantom{a}0.4 & \phantom{a}0.6 & \phantom{a}1.2 \\  %
     \texttt{B17} & \ + w2v-BERT + t-mBART & 18.6 & 32.5 & 30.9 & 31.0 & 34.1 & 13.9 & 30.7 & 36.9 & 10.6 & 31.2 & 26.1 & 18.4 & 11.6 & \phantom{a}1.9 & 14.7 & 10.4 & 15.1 & 16.2 & 10.6 & \phantom{a}1.1 & \phantom{a}3.9 & \phantom{a}9.7 \\  %
     \midrule[\heavyrulewidth]

     \multicolumn{6}{l}{\textbf {Direct speech-to-unit systems}} \\
     \texttt{B18} & {\unitv} & \phantom{a}9.1 & 28.3 & 21.7 & 24.6 & 29.0 & \phantom{a}2.5 & 25.2 & 21.7 & \phantom{a}4.0 & 11.1 & 10.2 & \phantom{a}4.9 & \phantom{a}0.8 & \phantom{a}0.1 & \phantom{a}0.9 & \phantom{a}1.8 & \phantom{a}1.4 & \phantom{a}1.2 & \phantom{a}0.5 & \phantom{a}0.1 & \phantom{a}0.4 & \phantom{a}0.7 \\
     \texttt{B19} & \ + S2TT pre-training & 11.4 & 29.4 & 23.3 & 25.7 & 30.5 & \phantom{a}7.4 & 26.5 & 24.6 & \phantom{a}6.9 & 16.7 & 15.6 & 10.6 & \phantom{a}3.3 & \phantom{a}0.5 & \phantom{a}4.6 & \phantom{a}2.2 & \phantom{a}2.6 & \phantom{a}1.4 & \phantom{a}4.7 & \phantom{a}0.3 & \phantom{a}0.9 & \phantom{a}1.0 \\
     \texttt{B20} & \ + w2v-BERT + u-mBART & 20.8 & 32.7 & 28.5 & 30.6 & 34.8 & 12.8 & 31.7 & 37.5 & \phantom{a}7.6 & 37.2 & 27.2 & 18.2 & 15.0 & \phantom{a}1.8 & 18.6 & 18.5 & 20.5 & 29.8 & 13.1 & \phantom{a}1.3 & \phantom{a}4.0 & 16.2 \\  %
     \cmidrule(lr){1-24}

     \texttt{B21} & {\unity} & 12.0 & 30.9 & 25.5 & 27.2 & 32.3 & \phantom{a}5.1 & 28.2 & 28.2 & \phantom{a}7.2 & 20.3 & 17.1 & \phantom{a}9.1 & \phantom{a}2.5 & \phantom{a}0.4 & \phantom{a}2.2 & \phantom{a}3.7 & \phantom{a}6.1 & \phantom{a}1.8 & \phantom{a}2.3 & \phantom{a}0.1 & \phantom{a}1.2 & \phantom{a}1.0 \\
     \texttt{B22} & \ + S2TT pre-training & 13.0 & 32.1 & 26.8 & 29.1 & 33.4 & \phantom{a}8.3 & 29.4 & 27.6 & \phantom{a}7.9 & 20.3 & 19.7 & 12.1 & \phantom{a}3.5 & \phantom{a}0.6 & \phantom{a}4.6 & \phantom{a}2.5 & \phantom{a}4.9 & \phantom{a}1.9 & \phantom{a}5.8 & \phantom{a}0.3 & \phantom{a}1.0 & \phantom{a}1.0 \\
     \texttt{B23} & \ + w2v-BERT + t-mBART & {\bf 24.5} & {\bf 35.2} & {\bf 32.6} & {\bf 33.3} & {\bf 37.2} & {\bf 14.9} & {\bf 35.0} & {\bf 42.3} & {\bf 10.8} & {\bf 41.7} & {\bf 32.5} & {\bf 22.2} & {\bf 18.7} & {\bf \phantom{a}2.7} & {\bf 24.6} & {\bf 21.3} & {\bf 26.6} & {\bf 34.1} & {\bf 16.5} & {\bf \phantom{a}1.8} & {\bf \phantom{a}8.0} & {\bf 22.9}  \\  %
     \bottomrule
    \end{tabular}
    }
    \endgroup
    \vspace{-1mm}
    \caption{Full results of ASR-BLEU on CVSS-C corpus. ${}^{\diamondsuit}$Results from ~\citep{jia-etal-2022-cvss}, ${}^{\spadesuit}$Results from ~\citep{jia2022leveraging}. We use the S2TT model in \texttt{B3} for S2TT pre-training. t-mBART and u-mBART stand for text-based mBART and unit-based mBART, respectively. All w2v-BERT and mSLAM encoders have 0.6B parameters. }
    \label{tab:results_cvss_c_s2st_detail}
\end{table*}

\section{Additional experimental results}
In this section, we present additional experimental results in \cref{sec:result}.

\paragraph{FLOPs}\label{appendix:result_flops}
In Figure~\ref{fig:flops}, we show the results of FLOPs measured with a subset of the multi-domain Es$\to$En dev set, as discussed in \cref{ssec:decoding_efficiency}.
{\unity} achieved {\flopsreductionoverspec}$\times$ and {\flopsreductionoverstut}$\times$ FLOPs reduction over {\specvvplus} and {\unitv} models, respectively.

\paragraph{Fisher Es$\to$En}\label{appendix:result_fisher}
The results on Fisher are shown in Table~\ref{tab:results_fisher_s2st}.
We report average scores over three runs with different random seeds.
Among our four direct systems trained from scratch (\texttt{A11}, \texttt{A15}, \texttt{A18}, \texttt{A20}), {\unity} (\texttt{A20}) achieved the best ASR-BLEU.
Our {\unitv} (\texttt{A18}) and {\specvvplus} (\texttt{A15}) outperformed the previous studies (\texttt{A13}, \texttt{A17}) by a large margin.\footnote{\texttt{A15} predicts phonemes while \texttt{A16} predicts subwords in the first pass.}
Because {\specvvplus} outperformed {\unitv}, the two-pass decoding was the main factor of the improvements although it was complementary to targeting discrete units.
Moreover, the two-pass direct models (\texttt{A15}, \texttt{A20}) outperformed a cascaded system (\texttt{A6}).

Next, we pre-trained the speech encoder with wav2vec2.0 (\texttt{A12}, \texttt{A16}, \texttt{A19}, \texttt{A21}).\footnote{We did not pre-train the text decoder with t-mBART because it was not helpful on this corpus. This is because Fisher is a conversational domain, which is very different from text data used for t-mBART pre-training. We could make the text decoder pre-training effective by including conversational data during t-mBART pre-training, which we leave future work.}
We confirmed that all the models benefited from the pre-training, but the gain was small for {\specv}.
Unlike when training the models from scratch, {\specvvplus} gained the most and achieved the best test ASR-BLEU, 58.3.
To the best of our knowledge, this is the new state-of-the-art S2ST result on this corpus.
However, {\unity} has an advantage of decoding efficiency over {\specvvplus} as discussed in \cref{ssec:decoding_efficiency}.
All direct models (\texttt{A16}, \texttt{A19}, \texttt{A21}) except for {\specv} outperformed the corresponding cascaded system (\texttt{A7}).

\paragraph{CVSS-C}\label{appendix:result_cvss_detail}
We show the full results of each language direction on CVSS-C in Table~\ref{tab:results_cvss_c_s2st_detail}.

\begin{table}[t]
    \centering
    \begingroup
    \scalebox{0.85}{
    \begin{tabular}{lllcc} \toprule
    \multirow{2}{*}{ID} & \multirow{2}{*}{Model} & \multirow{2}{*}{\shortstack{Initialization of\\first-pass decoder}} & \multicolumn{2}{c}{(ASR-)BLEU ($\uparrow$)} \\ \cmidrule(lr){4-5}
     & & & Text & Speech \\
    \midrule
     \texttt{F1} & \multirow{7}{*}{\unity} & Random & 34.8 & 30.7 \\
     \texttt{F2} & & t-mBART & {\bf 38.3} & {\bf 33.2} \\
     \texttt{F3} & & Unsupervised MT & 38.2 & {\bf 33.2} \\
     \texttt{F4} & & Supervised MT1 & 36.6 & 33.0 \\  %
     \texttt{F5} & & Supervised MT2 & 37.5 & {\bf 33.3} \\  %
     \texttt{F6} & & S2TT (\texttt{F7}) & 37.8 & 32.5 \\
     \midrule

     \texttt{F7} & S2TT & t-mBART & 38.0 & -- \\
     \bottomrule
    \end{tabular}
    }
    \endgroup
    \vspace{-1mm}
    \caption{Results of pre-training strategies for the first-pass decoder in {\unity} on multi-domain Es$\to$En dev set}
    \label{tab:pretraining_first_pass_decoder}
\end{table}

\paragraph{Pre-training first-pass decoder}\label{ssec:result_text_decoder_pretraining}
We explored a better pre-training strategy for the first-pass text decoder in {\unity}.
We investigated pre-training it with an MT model trained with bitext data from scratch (\textit{Supervised MT1}, \textit{Supervised MT2}).
Supervised MT1 used CCMatrix~\citep{schwenk-etal-2021-ccmatrix} while Supervised MT2 is the MT model in the cascaded system\footnote{We used OpenSubtitle2018, UNCorpus, EUBookshop v2, Europarl v10, Wikipedia v1.0, and TED2020 v1 for training.}.
Moreover, we fine-tuned the t-mBART model to the MT task in an unsupervised MT way via online back translation~\citep{liu2020multilingual} on CC100 (\textit{unsupervised MT}).
Furthermore, we studied initializing the speech encoder and the text decoder with a separate direct S2TT model.
The S2TT model was fine-tuned from wav2vec2.0 and t-mBART models on the same corpus.
After the initialization, we fine-tuned the whole parameters of {\unity} except FFN layers in the first-pass text decoder (\textit{S2TT}).

The results in Table~\ref{tab:pretraining_first_pass_decoder} showed that pre-training the first-pass decoder with the vanilla t-mBART (\texttt{F2}) or the unsupervised MT model (\texttt{F3}) was the most effective.
Pre-training with supervised MT models (\texttt{F4}, \texttt{F5}) did not improve performance, even for the first pass.
This is consistent with a finding in \citet{jia2022leveraging} although they pre-train the first-pass phoneme decoder of {\translatotronvv} with a phoneme-based supervised MT model.
Therefore, leveraging a separate MT system is effective for generating weak supervisions~\citep{popuri2022enhanced} rather than parameter initialization.
Pre-training a part of {\unity} with an independent S2TT model (\texttt{F7}) was not helpful either.
Surprisingly, the BLEU score from the text decoder in {\unity} was better than that of \texttt{F7}.
Therefore, training signals from the unit decoder never affect the text decoder.

\begin{table*}[t]
    \centering
    \begingroup
    \scalebox{0.81}{
    \begin{tabular}{llcccc} \toprule
     \multirow{4}{*}{ID} & \multirow{4}{*}{Model} & \multicolumn{4}{c}{(ASR-)BLEU ($\uparrow$)} \\
     \cmidrule(lr){3-6}
     & & \multicolumn{2}{c}{\multirow{2}{*}{Fisher}} & \multicolumn{2}{c}{\multirow{2}{*}{\shortstack{Multi-domain\\Es$\to$En}}} \\
     & &  &  &  &  \\ \cmidrule(lr){3-4} \cmidrule(lr){5-6}
     & & Text & Speech & Text & Speech \\
     \midrule
      \texttt{D1} & {\specvvplus} & {\bf 54.4} & {\bf 49.2} & {\bf 35.0} & {\bf 30.8} \\ %
      \texttt{D2} & \ + w/o T2S encoder & {\bf 54.3} & 17.4 & 34.9 & 25.0 \\
      \texttt{D3} & \ + w/o R-Drop & 51.6 & 45.9 & 34.8 & 30.3 \\
      \midrule

      \texttt{D5} & {\unity} & {\bf 55.4} & {\bf 50.5} & {\bf 38.3} & {\bf 33.2} \\
      \texttt{D6} & \ + w/o T2U encoder & 55.0 & 49.1 & 38.1 & 30.7 \\
      \texttt{D7} & \ + w/o R-Drop & 53.2 & 48.2 & 37.7 & 32.1 \\
      \texttt{D8} & \ + Cross-attention to speech encoder (sequential) & {\bf 55.4} & 50.3 & 38.2 & {\bf 33.2} \\
      \texttt{D9} & \ + Cross-attention to speech encoder (parallel) & 55.3 & 50.4 & 38.1 & 33.1 \\
      \texttt{D10} & \ + CTC on unit decoder & 55.3 & 50.2 & \todo{n/a} & \todo{n/a} \\  %
     \bottomrule
    \end{tabular}
    }
    \endgroup
    \vspace{-1mm}
    \caption{Ablation study for two-pass direct S2ST models on Fisher Es$\to$En and multi-domain Es$\to$En dev sets. The first-pass decoder in all the models on Fisher is initialized randomly while it is pre-trained with t-mBART on multi-domain corpora.}
    \label{tab:ablation_study_all}
\end{table*}

\begin{table*}[t]
    \centering
    \begingroup
    \scalebox{0.86}{
    \begin{tabular}{lcllcccc} \toprule
     \multirow{4}{*}{ID} & \multirow{4}{*}{\shortstack{Encoder\\pre-training}} & \multirow{4}{*}{Model} & \multirow{4}{*}{\shortstack{Output\\unit}} & \multicolumn{4}{c}{(ASR-)BLEU ($\uparrow$)} \\
     \cmidrule(lr){5-8}
     & & & & \multicolumn{2}{c}{\multirow{2}{*}{Fisher}} & \multicolumn{2}{c}{\multirow{2}{*}{\shortstack{Multi-domain\\Es$\to$En}}} \\
     & & & &  &  &  &  \\ \cmidrule(lr){5-6} \cmidrule(lr){7-8}
     & & & & Text & Speech & Text & Speech \\
     \midrule
     \texttt{E1} & & \multirow{3}{*}{\specvvplus} & Phoneme & -- & {\bf 50.4} & -- & -- \\
     \texttt{E2} & & & Character & 54.0 & 50.2 & -- & -- \\
     \texttt{E3} & & & Subword & {\bf 54.4} & 49.2 & -- & -- \\
      \cmidrule(lr){1-8}

     \texttt{E1'} & \multirow{3}{*}{$\checkmark$} & \multirow{3}{*}{\specvvplus} & Phoneme & -- & 58.1 & -- & 29.4 \\
     \texttt{E2'} & & & Character & 61.5 & 58.1 & 31.7 & 28.9 \\
     \texttt{E3'} & & & Subword & {\bf 62.0} & {\bf 58.4} & {\bf 33.0} & {\bf 30.0} \\
     \midrule[\heavyrulewidth]

     \texttt{E4} & & \multirow{3}{*}{\unity} & Phoneme & -- & 49.8 & -- & -- \\
     \texttt{E5} & & & Character & 53.7 & 48.9 & -- & -- \\
     \texttt{E6} & & & Subword & {\bf 55.4} & {\bf 50.5} & -- & -- \\
     \cmidrule(lr){1-8}

     \texttt{E4'} & \multirow{3}{*}{$\checkmark$} & \multirow{3}{*}{\unity} & Phoneme & -- & 54.7 & -- & 27.8  \\
     \texttt{E5'} & & & Character & 60.9 & 55.0 & 33.2 & 29.6 \\
     \texttt{E6'} & & & Subword & {\bf 61.2} & {\bf 55.1} & {\bf 34.1} & {\bf 30.1} \\
     \bottomrule
    \end{tabular}
    }
    \endgroup
    \vspace{-1mm}
    \caption{Results of output units for the first-pass decoder in two-pass direct S2ST models on Fisher Es$\to$En and multi-domain Es$\to$En dev sets. We use 1k and 2k units for the subword vocabulary on Fisher and multi-domain Es$\to$En corpora, respectively. The first-pass decoder in all the models is initialized randomly.}
    \label{tab:output_unit_all}
\end{table*}

\paragraph{Ablation study}\label{appendix:result_ablation_study}
In Table~\ref{tab:ablation_study_all}, we show full results of the ablation study presented in \cref{ssec:result_ablation_study}.
An auxiliary CTC objective for the unit decoder, as used for the S2UT model~\citep{lee2021direct}, was not helpful for {\unity} (\texttt{D10}).
This was because the introduction of the first-pass decoder already eased for the second-pass decoder to learn monotonic alignments.

\paragraph{Output unit for first-pass decoder}\label{appendix:result_output_unit}
In Table~\ref{tab:output_unit_all}, we show full results of the comparison of the output units for the first-pass decoder in two-pass direct S2ST models presented in \cref{ssec:result_output_unit}.
The results showed that the subword unit was the best for {\unity} regardless of pre-training the speech encoder with wav2vec2.0.
In contrast, in the case of {\specvvplus}, the best output unit differed according to whether we pre-trained the speech encoder or not.
The phoneme unit was best when training the model from scratch (\texttt{E1}) while the subword unit was best when pre-training the encoder (\texttt{E3'}).
However, predicting subwords in the first pass led to the best BLEU score for the text output in all the settings.

\paragraph{ASR-chrF}
Following a finding that ASR-chrF is a more robust evaluation metric than ASR-BLEU in \citet{salesky2021assessing}, we also calculated ASR-chrF on Fisher, CVSS-C, and multi-domain corpora in Table~\ref{tab:results_fisher_s2st_chrf}, Table~\ref{tab:results_cvss_c_s2st_chrf}, and Table~\ref{tab:results_enes_s2st_chrf}, respectively.
Overall, we confirmed the similar trends to ASR-BLEU.

\subsection{Human evaluation}\label{appendix:result_human_eval}
Finally, we conducted an audio-only human evaluation to assess the translation quality while removing the necessity of ASR systems.
We adopted cross-lingual semantic textual similarity (XSTS)~\citep{licht-etal-2022-consistent} and percent acceptable translations.

\paragraph{Mean translation score}
We used XSTS, which emphasizes adequacy rather than fluency, as the most appropriate human evaluation protocol.
Annotators judged the semantic similarity between the source and the translated sentence.
As a result, whether a translation conveys the original meaning is more important than whether it has perfect syntax, wording, and grammar.
Annotators assigned each item a score from one to five.
A score of no less than three means the meaning is at least “mostly equivalent.”
We treat a translation that received a score of no less than three as having “acceptable” quality.
Annotators need to be bilingual, as they compare the source and translated sentences directly.
Since XSTS is an audio-only evaluation metric, it also considers the audio quality.

For each system, we computed the average XSTS score across items.
We set a target of over four average XSTS for systems where we expect or desire high-quality translations.
We set a target of over three average XSTS for systems where we expect a medium level of quality.

\paragraph{Percent acceptable translations}
For each system, we also computed the percentage of items that received an XSTS score of three or above.
We refer to this as the percent acceptable translations.
This metric helps us understand what percentage of translations produced by the system can preserve meaning adequately and what percentage has very low and unacceptable quality.
This metric tends to be more stable and less sensitive to annotator agreement than the average XSTS score.

\paragraph{Evaluation setting}
We used the mTEDx test set (989 samples) and generated the target audio from the S2ST systems.
Moreover, we randomly sampled 495 samples and generated the target audio from the reference translation followed by TTS.
The reference translations serve as a reference point and a ceiling against which to compare our systems.
Three bilingual annotators evaluated each item and assigned it a score from one to five.
The median score was taken per item.

\begin{figure}[t]
  \centering
  \includegraphics[width=0.92\linewidth]{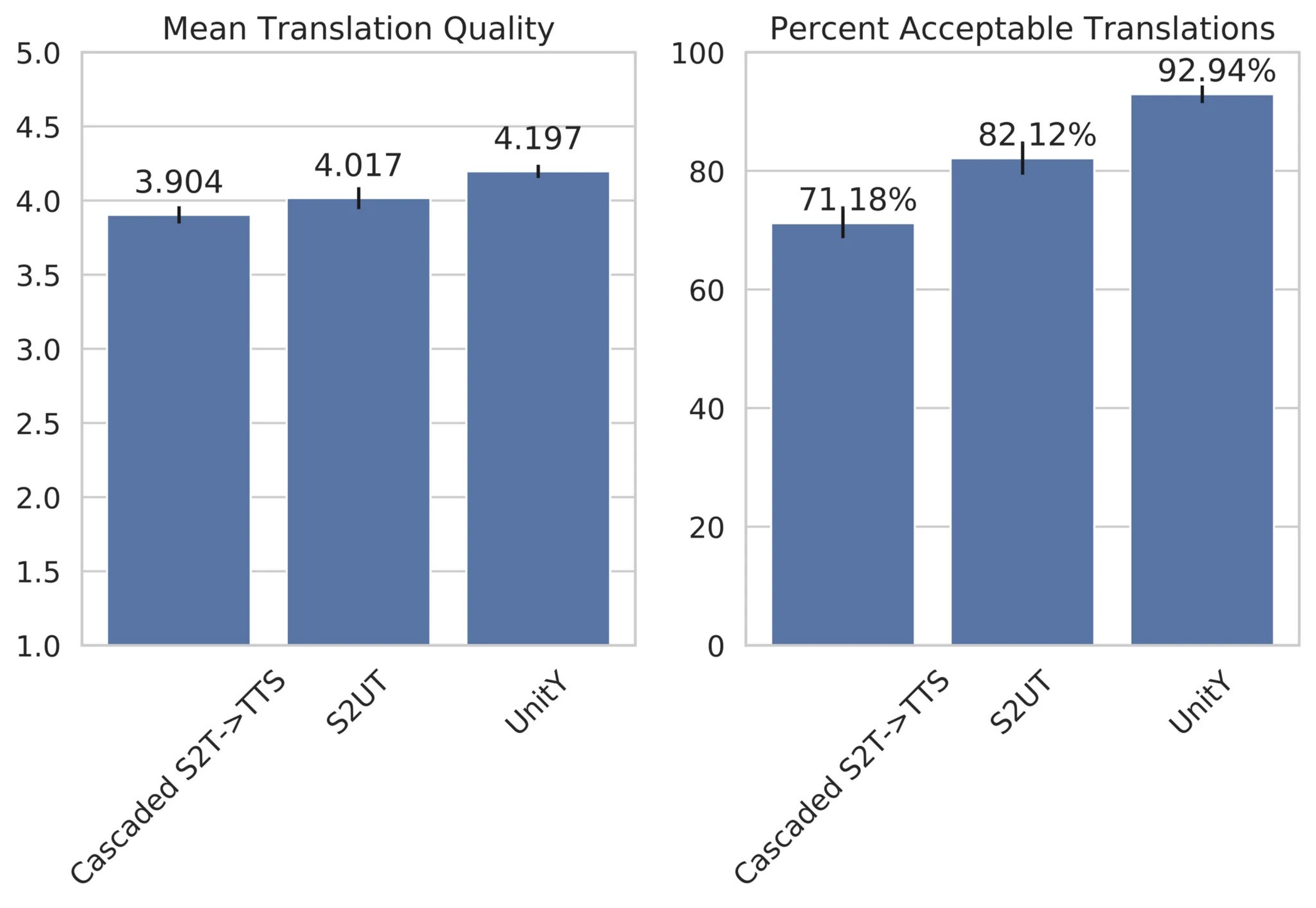}
  \vspace{-3mm}
  \caption{Results of human evaluation on multi-domain Es$\to$En corpus}
  \label{fig:human_eval}
  \vspace{-2mm}
\end{figure}

\paragraph{Results}
The results are presented in Figure~\ref{fig:human_eval}.\footnote{The models used here are early versions and slightly different from the models in Table~\ref{tab:results_enes_s2st}.}
We confirmed that {\unity} consistently outperformed the cascaded and {\unitv} models in both metrics.

\clearpage

\begin{table*}[t]
    \centering
    \begingroup
    \scalebox{0.72}{
    \begin{tabular}{lllccc} \toprule
    \multirow{2}{*}{ID} & \multirow{2}{*}{Model} & \multirow{2}{*}{Encoder} & \multicolumn{3}{c}{ASR-chrF ($\uparrow$)} \\
    \cmidrule(lr){4-6}
     & & & dev & dev2 & test \\
     \midrule[\heavyrulewidth]

     \multicolumn{6}{l}{\textbf {Cascaded systems}} \\
     \texttt{A6} & \multirow{2}{*}{S2TT $\to$ TTS} & Conformer & 0.642 & 0.652 & 0.649 \\  %
     \texttt{A7} & & Conformer wav2vec2.0 & 0.671 & 0.684 & 0.680 \\  %
     \midrule[\heavyrulewidth]

     \multicolumn{6}{l}{\textbf {Direct speech-to-spectrogram systems}} \\
     \texttt{A11} & \multirow{2}{*}{\specv} & Conformer & 0.612 & 0.621 & 0.618 \\  %
     \texttt{A12} & & Conformer wav2vec2.0 & 0.638 & 0.655 & 0.649 \\  %
     \cmidrule(lr){1-6}

     \texttt{A15} & \multirow{2}{*}{\specvvplus} & Conformer & 0.649 & 0.661 & 0.657 \\  %
     \texttt{A16} & & Conformer wav2vec2.0 & {\bf 0.695} & {\bf 0.708} & {\bf 0.702} \\  %
     \midrule[\heavyrulewidth]

     \multicolumn{6}{l}{\textbf {Direct speech-to-unit systems}} \\
     \texttt{A18} & \multirow{2}{*}{\unitv} & Conformer & 0.626 & 0.642 & 0.643 \\  %
     \texttt{A19} & & Conformer wav2vec2.0 & 0.677 & 0.688 & 0.685 \\  %
     \cmidrule(lr){1-6}

     \texttt{A20} & \multirow{2}{*}{\unity}
     & Conformer & 0.646 & 0.658 & 0.658 \\  %
     \texttt{A21} & & Conformer wav2vec2.0 & {\bf 0.678} & {\bf 0.692} & {\bf 0.687} \\  %
     \bottomrule
    \end{tabular}
    }
    \endgroup
    \vspace{-1mm}
    \caption{ASR-chrF on Fisher Es$\to$En corpus. The decoder in all the models is initialized randomly. {\specvvplus} is our improved version of {\translatotronvv}.}
    \label{tab:results_fisher_s2st_chrf}
\end{table*}

\begin{table*}[t]
    \centering
    \begingroup
    \scalebox{0.72}{
    \begin{tabular}{llllccc} \toprule
    \multirow{2}{*}{ID} & \multirow{2}{*}{Model} & \multicolumn{4}{c}{ASR-chrF ($\uparrow$)} \\ \cmidrule(lr){3-6}
     & & Avg. & High & Mid & Low \\
     \midrule[\heavyrulewidth]
     \multicolumn{6}{l}{\textbf {Cascaded systems}} \\
     \texttt{B3} & S2TT $\to$ TTS & 0.304 & 0.504 & 0.384 & 0.204 \\
     \texttt{B4} & \ + w2v-BERT + t-mBART & 0.420 & 0.533 & 0.463 & 0.365 \\
     \midrule[\heavyrulewidth]

     \multicolumn{6}{l}{\textbf {Direct speech-to-spectrogram systems}} \\
     \texttt{B6} & {\specv} & 0.273 & 0.498 & 0.328 & 0.175 \\
     \texttt{B7} & \ + S2TT pre-training & 0.311 & 0.521 & 0.377 & 0.213 \\
     \texttt{B8} & \ + w2v-BERT & 0.395 & 0.582 & 0.461 & 0.306 \\
     \cmidrule(lr){1-6}

     \texttt{B15} & {\specvvplus} &  0.306 & 0.560 & 0.389 & 0.187 \\  %
     \texttt{B16} & \ + S2TT pre-training & 0.336 & 0.566 & 0.417 & 0.226 \\  %
     \texttt{B17} & \ + w2v-BERT + t-mBART & 0.419 & 0.592 & 0.492 & 0.331 \\
     \midrule[\heavyrulewidth]

     \multicolumn{6}{l}{\textbf {Direct speech-to-unit systems}} \\
     \texttt{B18} & {\unitv} & 0.294 & 0.536 & 0.356 & 0.188 \\
     \texttt{B19} & \ + S2TT pre-training & 0.329 & 0.550 & 0.405 & 0.224 \\
     \texttt{B20} & \ + w2v-BERT + u-mBART & 0.445 & 0.588 & 0.495 & 0.377 \\
     \cmidrule(lr){1-6}

     \texttt{B21} & {\unity} & 0.312 & 0.564 & 0.396 & 0.192 \\
     \texttt{B22} & \ + S2TT pre-training & 0.333 & 0.572 &0.415 & 0.220 \\
     \texttt{B23} & \ + w2v-BERT + t-mBART & {\bf 0.474} & {\bf 0.607} & {\bf 0.521} & {\bf 0.410} \\
     \bottomrule
    \end{tabular}
    }
    \endgroup
    \vspace{-1mm}
    \caption{ASR-chrF on CVSS-C corpus. We use the S2TT model in \texttt{B3} for S2TT pre-training. t-mBART and u-mBART stand for text-based mBART and unit-based mBART, respectively. All w2v-BERT encoders have 0.6B parameters.}
    \label{tab:results_cvss_c_s2st_chrf}
\end{table*}

\begin{table*}[t!]
    \centering
    \begingroup
    \scalebox{0.82}{
    \resizebox{\linewidth}{!}{
    \begin{tabular}{llccccccccccccc} \toprule
        \multirow{3}{*}{ID} & \multirow{3}{*}{Model} & \multicolumn{7}{c}{ASR-chrF ($\uparrow$)} \\
        \cmidrule(lr){3-9}

        & & \multicolumn{3}{c}{{\bf En$\to$Es}} & \multicolumn{4}{c}{{\bf Es$\to$En}} \\ \cmidrule(lr){3-5} \cmidrule(lr){6-9}
        & & Europarl-ST & MuST-C & Avg. & CoVoST-2 & Europarl-ST & mTEDx & Avg. \\
        \midrule[\heavyrulewidth]

        \multicolumn{8}{l}{\textbf {Cascaded systems}} \\
        \texttt{C1'} & ASR$\to$MT$\to$TTS & 0.634 & 0.587 & 0.611 & 0.611 & 0.618 & 0.569 & 0.599 \\
        \texttt{C2'} & S2TT$\to$TTS & 0.639 & 0.613 & 0.626 & 0.642 & 0.620 & 0.588 & 0.620 \\
        \midrule[\heavyrulewidth]

        \multicolumn{8}{l}{\textbf {Direct speech-to-spectrogram systems}} \\
        \texttt{C3} & {\specvvplus} (6L$\to$6L) & 0.634 & 0.606 & 0.620 & {\bf 0.642} & 0.484 & 0.578 & 0.568 \\
        \texttt{C4} & \ + t-mBART (12L$\to$6L) & {\bf 0.642} & {\bf 0.611} & {\bf 0.627} & {\bf 0.642} & 0.485 & {\bf 0.583} & 0.570 \\
        \midrule[\heavyrulewidth]

        \multicolumn{8}{l}{\textbf {Direct speech-to-unit systems}} \\
        \texttt{C5'} & {\unitv} + u-mBART & 0.610 & 0.615 & 0.613 & 0.621 & 0.587 & 0.568 & 0.592 \\
        \cmidrule(lr){1-9}

        \texttt{C6} & {\unity} (6L$\to$6L) & {\bf 0.643} & 0.618 & 0.631 & 0.628 & 0.591 & 0.575 & 0.598 \\
        \texttt{C7} & \ + t-mBART (12L$\to$2L) & 0.641 & {\bf 0.622} & {\bf 0.632} & {\bf 0.633} & {\bf 0.606} & {\bf 0.583} & {\bf 0.607} \\ %
    \bottomrule
    \end{tabular}
    }
    }
    \endgroup
    \vspace{-1mm}
    \caption{ASR-chrF on multi-domain En$\leftrightarrow$Es. The encoder in all the models is pre-trained with wav2vec2.0. t-mBART and u-mBART stand for text-based mBART and unit-based mBART, respectively. $\nlayerfirstdec$L$\to\nlayerseconddec$L stands for an $\nlayerfirstdec$-layer first-pass decoder with an $\nlayerseconddec$-layer second-pass decoder.}
    \label{tab:results_enes_s2st_chrf}
\end{table*}

\end{document}